\documentclass[letterpaper]{article} 
\usepackage{aaai2026}  
\usepackage{times}  
\usepackage{helvet}  
\usepackage{courier}  
\usepackage[hyphens]{url}  
\usepackage{graphicx} 
\usepackage{natbib}  
\usepackage{caption} 
\usepackage{algorithm}
\usepackage{algorithmic}
\usepackage{amsmath}
\usepackage{amsthm}

\newtheorem{theorem}{Theorem}
\usepackage{newfloat}
\usepackage{listings}

\usepackage{amsfonts}
\usepackage{amsmath}
\usepackage{multirow}
\usepackage{booktabs}
\usepackage{rotating}
\usepackage[tight,footnotesize]{subfigure}

\DeclareCaptionStyle{ruled}{labelfont=normalfont,labelsep=colon,strut=off} 
\floatstyle{ruled}
\newfloat{listing}{tb}{lst}{}
\floatname{listing}{Listing}
\title{M$^2$FMoE: Multi-Resolution Multi-View Frequency Mixture-of-Experts for Extreme-Adaptive Time Series Forecasting}
\author{
    Yaohui Huang, Runmin Zou, Yun Wang\thanks{Corresponding author.}, Laeeq Aslam, Ruipeng Dong,
}
\affiliations{
    School of Automation, Central South University, Changsha, China\\

    \{yaohuihuang, rmzou, wangyun19, laeeq\_aslam, darol22\}@csu.edu.cn
}

\usepackage{bibentry}

\begin{document}

\maketitle

\begin{abstract}

Forecasting time series with extreme events is critical yet challenging due to their high variance, irregular dynamics, and sparse but high-impact nature. While existing methods excel in modeling dominant regular patterns, their performance degrades significantly during extreme events, constituting the primary source of forecasting errors in real-world applications. Although some approaches incorporate auxiliary signals to improve performance, they still fail to capture extreme events' complex temporal dynamics. To address these limitations, we propose M$^2$FMoE, an extreme-adaptive forecasting model that learns both regular and extreme patterns through multi-resolution and multi-view frequency modeling. It comprises three modules: (1) a multi-view frequency mixture-of-experts module assigns experts to distinct spectral bands in Fourier and Wavelet domains, with cross-view shared band splitter aligning frequency partitions and enabling inter-expert collaboration to capture both dominant and rare fluctuations; (2) a multi-resolution adaptive fusion module that hierarchically aggregates frequency features from coarse to fine resolutions, enhancing sensitivity to both short-term variations and sudden changes; (3) a temporal gating integration module that dynamically balances long-term trends and short-term frequency-aware features, improving adaptability to both regular and extreme temporal patterns. Experiments on real-world hydrological datasets with extreme patterns demonstrate that M$^2$FMoE outperforms state-of-the-art baselines without requiring extreme-event labels. Code are available at: https://github.com/Yaohui-Huang/M2FMoE .

\end{abstract}


\section{Introduction}
\label{sec:intro}

\begin{figure}
\centering
\includegraphics[width=1.0\linewidth]{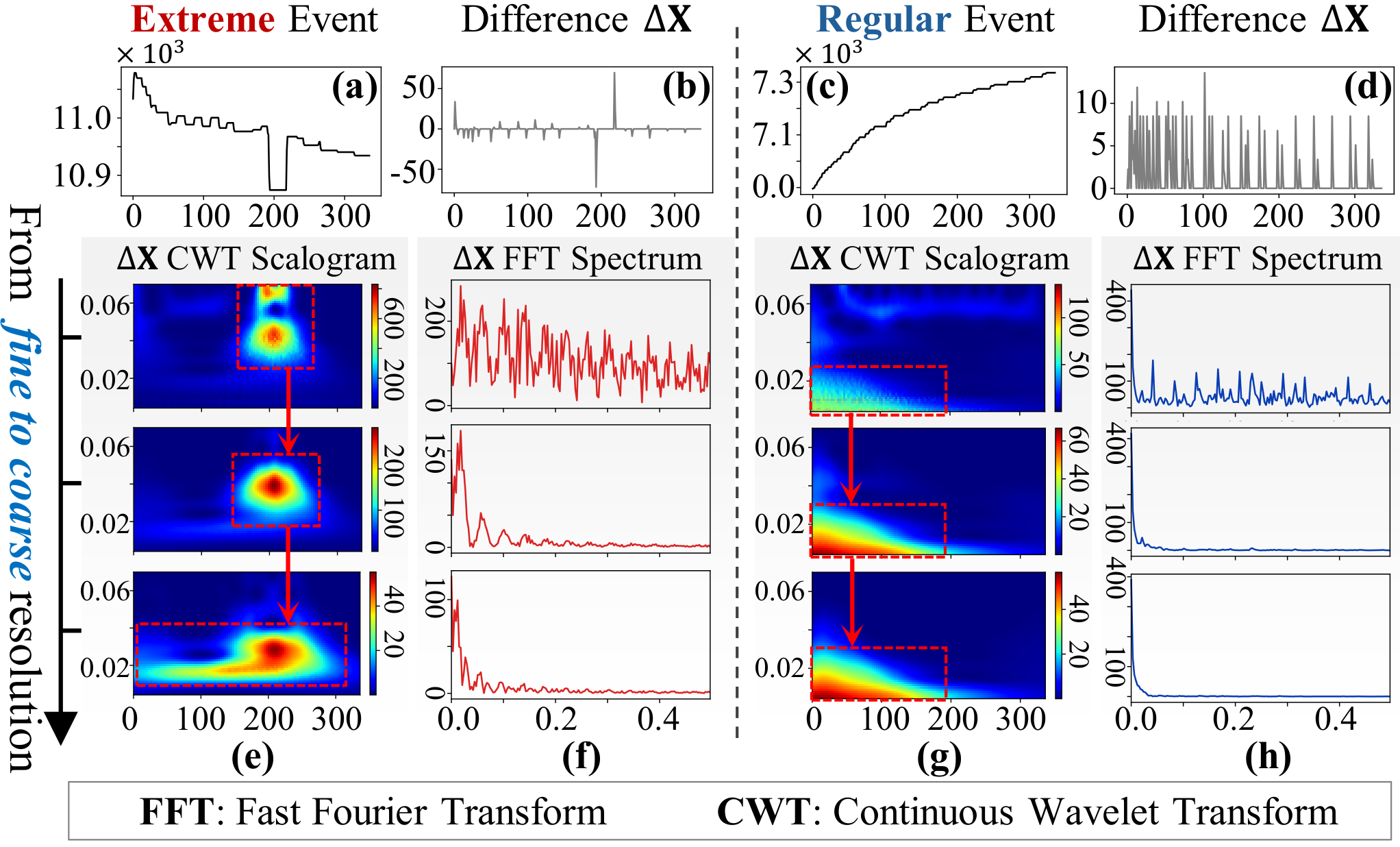}
\caption{Comparison of frequency spectra between regular and extreme events. }
\label{fig:intro}
\end{figure}

Time series forecasting is vital for decision-making across various real-world systems, including energy, transportation, and environmental monitoring \cite{journal/tpami2024/jin2, wang2024deep}. 
Among these, hydrological forecasting is particularly difficult due to extreme events like flash floods, heavy rainfall, and sudden water level rises \cite{lavers2014extending}. 
These events are rare, abrupt, and high variance, often causing significant deviations from regular temporal patterns \cite{camps2025artificial}. 
Despite their importance for risk management, forecasting such extremes remains one of the most challenging problems in time series modeling \cite{conference/aaai2024/li27768}.

Classical statistical models often fail under extreme or non-stationary conditions \cite{journal/neurocom2023/zhang}.
Recent deep learning advancements offer enhanced flexibility in modeling intricate temporal dependencies \cite{conference/ijcai/WenZZCMY023, wang2024deep}.
However, these models typically emphasize capturing dominant patterns such as periodic trends, smooth transitions, and local correlations, resulting in the inadequate representation of rare, high-impact extreme events. Consequently, forecasting models tend to perform well under regular conditions but struggle to accurately represent these infrequent but critical dynamics.
This limitation is especially pronounced in hydrological forecasting, where systems are highly sensitive to abrupt shifts, such as sudden heavy rainfall or rapid runoff \cite{journal/tpami2025/li6888}. Inaccurate predictions in such scenarios may lead to delayed warnings and severe consequences like widespread flooding.
These challenges highlight the urgent need for forecasting models that can accurately capture both regular trends and extreme deviations within a unified framework.

Frequency-domain representations provide a promising way to decompose temporal dynamics into spectral components, facilitating models to separate high-frequency fluctuations from low-frequency trends \cite{conference/aaai2024/ma, conference/aistats2025/liu25i}.
The spectral characteristics of extreme and regular events are shown in \textbf{Fig.\ref{fig:intro}}. As illustrated in \textbf{Fig.\ref{fig:intro}(a)–(d)}, the differenced sequences $\Delta \mathbf{X}$ reveal clear contrasts between the two types of events.
These differences become more pronounced in the wavelet domain (\textbf{Fig.~\ref{fig:intro}(e),~\ref{fig:intro}(g)}),  where extreme events produce sharp, localized energy at fine resolutions.
As the resolution becomes coarser, energy gradually shifts toward lower frequencies with reduced intensity, while the main structure of the event remains consistent across resolutions.
In contrast, regular events exhibit smooth low-frequency dynamics, resulting in diffuse and uniform energy distributions at all resolutions. Similar patterns are observed in the Fourier domain (\textbf{Fig.~\ref{fig:intro}(f),~\ref{fig:intro}(h)}). Extreme sequences exhibit broad-spectrum, multi-peaked energy with slow spectral decay, whereas regular sequences concentrate energy within narrow low-frequency bands.
These observations highlight the need for frequency-aware modeling to capture the varied spectral properties of temporal patterns. In particular, the results reveal \textbf{\textit{frequency heterogeneity}}, where different frequency bands contribute unequally to regular and extreme events. Accurately modeling such variation requires adaptive focus on informative frequencies, which is challenging within a single spectral domain. Fourier transforms provide accurate global frequency information but lack temporal resolution. In contrast, wavelet transforms offer time-frequency localization but suffer from reduced resolution at lower frequencies \cite{journal/aaai2025/Fei11645}.  Combining both views yields a more complete spectral representation that supports the modeling of both abrupt variations and long-term dependencies. Nevertheless, this dual-view strategy also introduces \textbf{\textit{cross-view spectral misalignment}}. Differences in basis functions and resolution cause the same signal to localize inconsistently across Fourier and Wavelet domains, resulting in cross-view incompatibility that undermines shared modeling.

To address this, we propose a Multi-resolution Multi-view Frequency Mixture-of-Experts (\textbf{M$^2$FMoE}) to model frequency-aware temporal dynamics under both regular and extreme conditions. Specifically, M$^2$FMoE first proposes a multi-view frequency mixture-of-experts (MFMoE) module to assigns specialized spectral experts to distinct frequency bands across both Fourier and Wavelet domains, thereby enabling selective specialization to handle diverse frequency characteristics. To ensure semantic coherence among experts and alleviate spectral misalignment, a cross-view shared band splitter (CSS) is integrated within MFMoE, aligning spectral boundaries across views. Furthermore, to capture temporal patterns at multiple resolutions, we introduce the multi-resolution adaptive fusion (MAF) module, which hierarchically aggregates features from coarse to fine frequency scales. Finally, a temporal gating integration (TGI) module adaptively fuses recent dynamics with long-range historical context via a learnable gating mechanism. 
Experiments on five hydrological datasets with extreme events demonstrate that M$^2$FMoE outperforms state-of-the-art methods without using auxiliary event labels.

\section{Related Work}
\label{sec:related}

Time series forecasting has evolved from classical models like ARIMA, which are interpretable but limited by linearity and stationarity \cite{journal/neurocom2023/zhang,wang2024deep}, to deep learning methods that offer greater flexibility. Early deep learning approaches, including RNNs \cite{conference/nips2024/jia, conference/aaai2025/kong} and CNNs \cite{conference/iclr2023/wu, conference/ecmlpkdd2023/chen}, focused on local dependencies. Transformers \cite{conference/iclr2025/liu, conference/iclr2024/liu, conference/neurips2024/kim} then introduced self-attention for long-range structure, while recent MLP-based models \cite{conference/icml2025/lin, conference/nips2024/106315lin, conference/iclr2025/liu2} offer efficient alternatives. Further advancements include GNNs \cite{journal/tpami2024/jin2, journal/tpami2025/jin} and Mixture-of-Experts (MoE) models \cite{conference/aistats2025/liu25i, conference/iclr2025/Shi} for nonlinear modeling, alongside multi-scale and multi-resolution representations \cite{conference/iclr2025/Wang, conference/iclr2024/wangtmixer} for capturing varied temporal granularities. Frequency-based approaches have also emerged, utilizing Fourier transforms for global periodic structures and wavelet transforms for localized time-frequency representations \cite{conference/aaai2024/ma, journal/aaai2025/Fei11645}. However, most existing models primarily target regular patterns, struggling with the irregular variations crucial for extreme events.

Extreme-adaptive forecasting targets time series with rare, abrupt, and high-impact changes such as floods or sudden surges in water levels, which require effective handling of rarity and volatility. Recent efforts span architectural designs and loss functions. Architecturally, models such as NEC+ \cite{conference/aaai2023/li26045}, VIE \cite{conference/aaai2021/Xiu10469}, SADI \cite{conference/icassp2023/Liu1}, and SEED \cite{conference/bigdata/Li728} use multi-phase learning for non-stationary dynamics. Others, like MCANN \cite{journal/tpami2025/li6888} and DAN \cite{conference/aaai2024/li27768}, integrate priors or clustering for enhanced robustness. On the loss side, specialized objectives like EPL \cite{conference/ijcai2024/Wang5135}, EVL \cite{conference/sigkdd2019/Ding1114}, and GEVL \cite{journal/tkde2021/Zhang2021} leverage Extreme Value Theory or biases to emphasize tail behavior. Despite these advancements, current extreme-adaptive methods often neglect the joint modeling of frequency-aware patterns and resolution-specific dynamics, limiting their generalization across diverse temporal variations.

\section{Preliminaries}
\label{sec:preliminaries}

\subsubsection{Problem Statement}

Let $\mathbf{X} = \{X_1, X_2, \ldots, X_{T_{in}}\}$ denote a multivariate time series, where each $X_t \in \mathbb{R}^C$ represents a $C$-dimensional observation at time step $t$, and $T_{in}$ is the input sequence length. The objective is to learn a forecasting model $\mathbb{F}(\cdot)$ that predicts the subsequent $T_p$ future values, denoted as
$\hat{\mathbf{X}} = \{\hat{X}_{T_{in}+1}, \ldots, \hat{X}_{T_{in}+T_p}\}$.

\subsubsection{Discrete Fourier Transform (DFT)}
The discrete Fourier transform (DFT) decomposes a sequence into global sinusoidal components. For $\mathbf{X}$ of length $T_{in}$, its $n$-th frequency coefficient is:
\begin{equation}
\mathcal{F}_n = \sum_{t=1}^{T_{in}} X_t \cdot e^{-j 2\pi n t / T_{in}}, \quad n \in \{0, 1, \ldots, T_{in}-1\},
\end{equation}
where $\mathcal{F}_n$ encodes periodicity but lacks temporal localization, limiting its applicability for non-stationary signals. 
FFT is implemented to efficiently compute $\mathcal{F}_n$.

\subsubsection{Continuous Wavelet Transform (CWT)}

The CWT enables localized time-frequency analysis. For a signal $X(t)$, the wavelet coefficient at scale $a$ and position $b$ is defined as:
\begin{equation}
\mathcal{W}(a, b) = \frac{1}{\sqrt{|a|}} \int_{-\infty}^{\infty} X(t) \, \psi^*\left( \frac{t - b}{a} \right) dt,
\end{equation}
where $\psi^*$ is the complex conjugate of the mother wavelet $\psi$.

\begin{figure*}[t]
\centering
\includegraphics[width=1.0\textwidth]{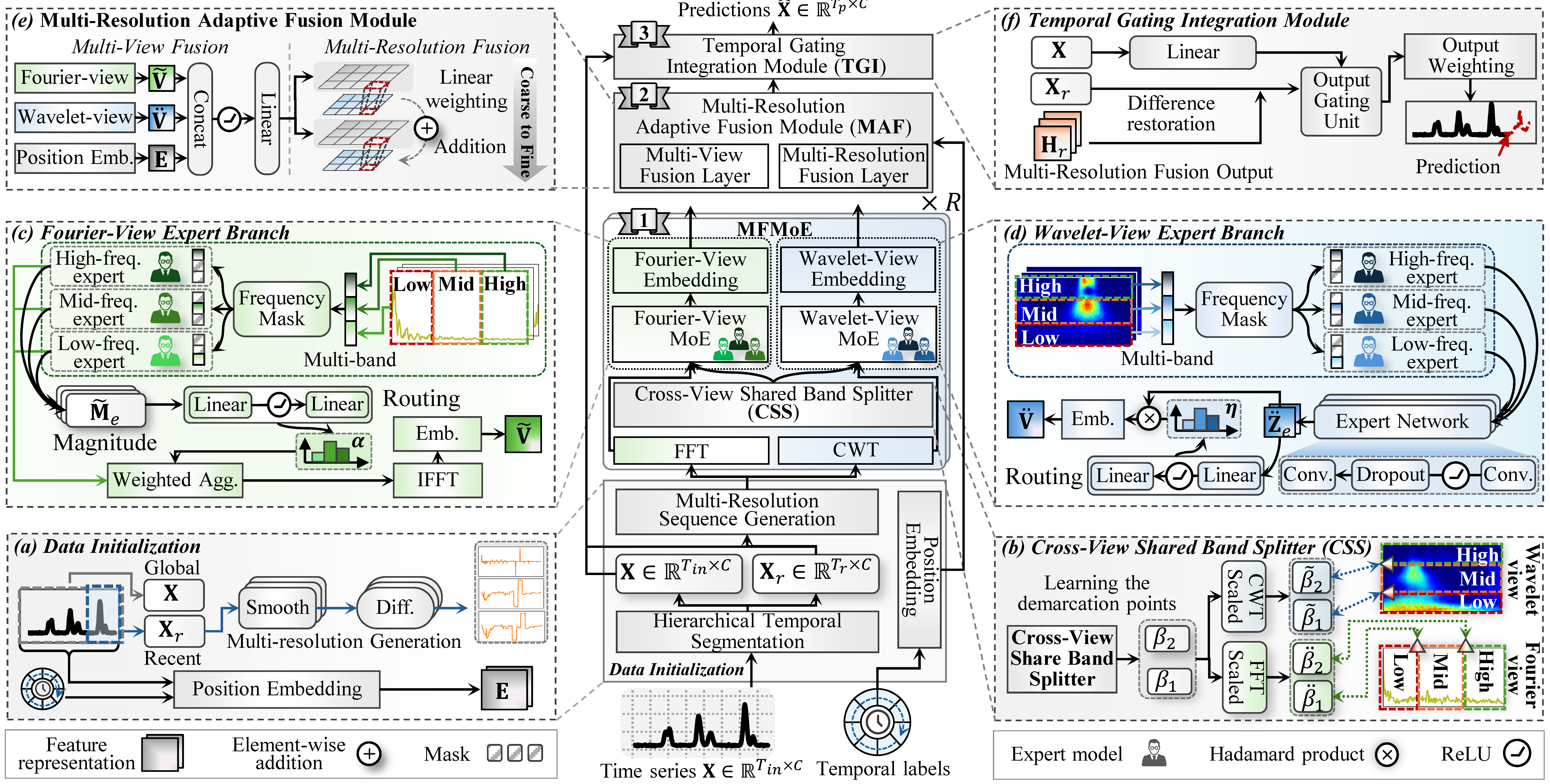}
\caption{The proposed M$^2$FMoE with three experts per branch for capturing high-, mid-, and low-frequency patterns.}
\label{fig:overview}
\end{figure*}

\section{Methodology}
\label{sec:method}

As illustrated in \textbf{Fig.~\ref{fig:overview}}, the proposed M$^2$FMoE comprises three modules: (1) an MFMoE module, (2) an MAF module, and (3) a TGI module. 
Each component is detailed below.

\subsubsection{Hierarchical Temporal Segmentation}

As shown in \textbf{Fig.~\ref{fig:overview}(a)}, the hierarchical temporal segmentation module extracts a recent segment $\mathbf{X}_r = \{X_{T_{in}-T_r+1}, \ldots, X_{T_{in}}\}$ to capture short-term dynamics, while the entire input sequence $\mathbf{X} = \{X_1, \ldots, X_{T_{in}}\}$ serves as the historical context to model long-term temporal patterns.

\subsubsection{Multi-Resolution Sequence Generation}

To capture temporal dynamics at varying granularities, the recent segment $\mathbf{X}_r \in \mathbb{R}^{T_r \times C}$ is decomposed into a multi-resolution set via 1D smoothing convolutions:
\begin{equation}
\mathcal{S} = \left\{ \tilde{\mathbf{X}}_r^{(k)} = \text{SmoothConv}(\mathbf{X}_r, k) \mid k \in \mathcal{K} \right\}.
\end{equation}

For each resolution $k$ (with $k_1 = 1$ retaining the original sequence), we compute the first-order difference of the transformed recent sequence to highlight local variations, i.e., $\Delta \mathbf{X}_r^{(k)} = \tilde{\mathbf{X}}_r^{(k)}[1:T_r] - \tilde{\mathbf{X}}_r^{(k)}[0:T_r-1]$.
The resulting multi-resolution differences ${\Delta \mathbf{X}_r^{(k)}}$ disentangle coarse and fine dynamics, forming a diverse input set for subsequent frequency-aware modeling in the MFMoE module.

\subsubsection{Temporal Embedding}

To encode temporal order, a fixed sinusoidal positional embedding is added as an auxiliary feature \cite{conference/iclr2024/liu}. Each time step is mapped to a combination of sine and cosine functions at varying frequencies:
\begin{equation}
\text{PE}(t, 2i) = \sin(t \cdot \omega_i), \quad
\text{PE}(t, 2i+1) = \cos(t \cdot \omega_i),
\end{equation}
where $t$ is the time index and $d$ the embedding dimension, $i$ is the index of the embedding dimension, and $\omega_i = 1 / 10000^{2i/d}$. These embeddings are added to input features to provide position-aware inductive bias across time steps.

\subsection{Multi-View Frequency Mixture-of-Experts Module}

The recent segment of extreme time series is challenging due to the sparsity and volatility of extreme events. To address this, we shift the learning paradigm to the frequency domain via the MFMoE module, which comprises two expert branches: a Fourier-view and a Wavelet-view branch. A CSS is introduced to align frequency bands across both domains.

\subsubsection{Cross-View Shared Band Splitter}

To capture multi-resolution temporal dynamics, we construct dual-view spectral representations using both Fourier and Wavelet transforms. However, a key challenge arises from the inherent differences in these views. The Fourier transform organizes spectral components on a uniform frequency axis, while the CWT uses scales that correspond nonlinearly to frequency. Consequently, aligning expert assignments between the two views is difficult, as the same frequency content can appear at different positions in each representation.

To ensure consistent expert specialization across both spectral views, we propose the \textit{Cross-View Shared Band Splitter}. \textbf{Theorem \ref{thm:band_alignment}} provides the theoretical basis for this module by formalizing the correspondence between frequency and wavelet scale.
As shown in \textbf{Fig.~\ref{fig:overview}(b)}, the splitter learns shared frequency boundaries $\{\beta_1, \beta_2, \ldots, \beta_{E-1}\}$ to divide the frequency range $[0, 1]$ into $E$ bands. For the FFT view, these boundaries are directly scaled into frequency indices $\{\tilde{\beta}_1, \ldots, \tilde{\beta}_{E-1}\}$. For the CWT view, they are nonlinearly mapped into wavelet scales $\{\ddot{\beta}_1, \ldots, \ddot{\beta}_{E-1}\}$ using the inverse relationship from \textbf{Theorem \ref{thm:band_alignment}}. This shared segmentation allows both views to decompose the input into semantically aligned sub-bands, ensuring experts operate on consistent spectral content.

\begin{theorem}[Spectral Boundary Correspondence]
\label{thm:band_alignment}
Let $f$ denote the normalized frequency, $a$ is the scale in the CWT, and $\gamma = f_0 / f_{\mathrm{nyq}}$ is a wavelet-dependent constant. The mapping $a = \gamma / f$ establishes a one-to-one correspondence between frequency and scale boundaries \cite{mallat2002theory}, such that $f_{\max} \mapsto a_{\min} = \gamma / f_{\max}$ and $f_{\min} \mapsto a_{\max} = \gamma / f_{\min}$. Under this mapping, signal energy is conserved, satisfying $\int_{f_{\min}}^{f_{\max}} |\mathcal{F}(f)|^2 df \propto \iint_{a \in [a_{\min}, a_{\max}]} |\mathcal{W}(a, b)|^2 \frac{da}{a^2} db$.
\text{Detailed proof is provided in the \textbf{Appendix A.}}
\end{theorem}

\subsubsection{Fourier-View Expert Branch}

As illustrated in \textbf{Fig.~\ref{fig:overview}(c)}, the Fourier-view expert branch is designed to extract frequency-aware representations by assigning specialized experts to distinct frequency bands. Using the shared boundaries $\{\tilde{\beta}_1, \ldots, \tilde{\beta}_{E-1}\}$, we divide the full frequency range into $E$ non-overlapping intervals (i.e., ${[0, \tilde{\beta}_1), [\tilde{\beta}_1, \tilde{\beta}_2), \ldots, [\tilde{\beta}_{E-1}, F]}$), where $F = T_r/2 + 1$ is the number of frequency bins after real-valued FFT. Each expert $e$ is responsible for modeling the frequency components within its assigned band, such as high-frequency, mid-frequency, and low-frequency patterns.

Given an input sequence $\Delta \mathbf{X}_r^{(k)} \in \mathbb{R}^{T_r \times C}$, we first perform per-channel standardization and apply the real FFT. Then, to isolate expert-specific frequency components, we define a set of binary masks ${\tilde{\mathbb{I}}_e} \in \mathbb{R}^{F \times C}$, each indicating the active sub-band for expert $e$. The masked spectrum for expert $e$ is:
\begin{equation}
    \mathcal{F}_{e} = \tilde{\mathbb{I}}_e \odot \mathcal{F}, \quad
    \tilde{\mathbb{I}}_e = \begin{cases}
        1, & \text{if } f \in [\tilde{\beta}_{e-1}, \tilde{\beta}_e) \\
        0, & \text{otherwise}
    \end{cases},
\end{equation}
where $\mathcal{F}$ represents the full spectrum generated by applying FFT to $\Delta \mathbf{X}_r^{(k)}$. To adaptively determine expert contributions, inspired by \cite{conference/icml2025/jin}, we employ a lightweight routing network. Specifically, we first compute the magnitude spectrum and average across channels, and then the summary vector is passed through a routing network $\tilde{\mathcal{G}}(\cdot)$ consisting of two linear layers with ReLU activation and softmax output to produce the expert routing weights:
\begin{equation}
\tilde{\mathbf{M}} = \frac{1}{C} \sum\nolimits_{c=1}^{C} |\mathcal{F}_{e}[c]|, \quad
\boldsymbol{\alpha} = \text{Softmax}\big(\tilde{\mathcal{G}}(\tilde{\mathbf{M}})\big),
\end{equation}
where $\boldsymbol{\alpha} = [\alpha_1, \ldots, \alpha_E]$ are the routing weights for each expert. $\tilde{\mathbf{M}} \in \mathbb{R}^{F}$ is the magnitude spectrum averaged across channels.
The final output of the Fourier-view expert branch is obtained by aggregating expert-specific frequency components using routing weights $\alpha_e$, followed by inverse FFT and a linear projection:
\begin{equation}
    \tilde{\mathbf{V}} = \text{Linear} \big( \text{IFFT} ( \sum\nolimits_{e=1}^{E} \alpha_e \cdot \mathcal{F}_{e} ) \big),
\end{equation}
where $\tilde{\mathbf{V}} \in \mathbb{R}^{T_p \times C}$ is the final output of the Fourier-view expert branch. This design enables dynamic selection of frequency bands and temporal adaptation based on input spectral statistics.

\subsubsection{Wavelet-View Expert Branch}

As illustrated in \textbf{Fig.~\ref{fig:overview}(d)}, the Wavelet-view expert branch captures temporally localized dynamics by operating on the CWT power spectrogram $\mathcal{P} = |\mathcal{W}(a, b)|^2 \in \mathbb{R}^{C \times S \times T_r}$, where $S$ is the number of wavelet scales. The CWT is computed using the complex Gaussian wavelet, ensuring balanced localization in both time and frequency domains.

The shared frequency boundaries are first converted to scale indices $\{\ddot{\beta}_1, \dots, \ddot{\beta}_{E-1}\}$ via the inverse mapping defined in \textbf{Theorem~\ref{thm:band_alignment}}. Each expert $e$ is assigned a binary scale mask $\mathbb{I}_e \in \{0,1\}^{S}$, and its corresponding component is computed as:
\begin{equation}
    \mathcal{P}_{e} = \ddot{\mathbb{I}}_e \odot \mathcal{P}, \quad
    \ddot{\mathbb{I}}_e =
    \begin{cases}
    1, & \text{if } s \in [\ddot{\beta}_{e-1}, \ddot{\beta}_e), \\
    0, & \text{otherwise}.
    \end{cases}
\end{equation}

Each expert network processes its masked input $\mathcal{P}_{e}$ using a convolutional block:
\begin{equation}
\ddot{\mathbf{Z}}_{e} = \mathbf{W}_{e, 2} * \bigg( \mathcal{D} \big( \text{ReLU}\left( \mathbf{W}_{e, 1} * \mathcal{P}_{e} \right) \big) \bigg),
\end{equation}
where $*$ denotes the convolution operation, $\mathbf{W}_{e, 1}, \mathbf{W}_{e, 2}$ are convolution kernels, and the dropout layer $\mathcal{D}(\cdot)$ is applied after the activation to prevent overfitting.

To adaptively assign expert contributions, the power spectrogram $\mathcal{P}$ is first averaged over the channel dimension to obtain a global summary $\ddot{\mathbf{M}} \in \mathbb{R}^{S \times T_r}$. This matrix is then flattened and passed through a lightweight routing network $\ddot{\mathcal{G}}(\cdot)$, which consists of two linear layers with ReLU activation, followed by a softmax function to produce the expert weighting vector:
\begin{equation}
    \ddot{\mathbf{M}} = \frac{1}{C} \sum\nolimits_{c=1}^{C} \mathcal{P}[c], ~~~
    \boldsymbol{\eta} = \text{Softmax} \left( \ddot{\mathcal{G}} \left(\text{Flatten}(\ddot{\mathbf{M}})\right) \right),
\end{equation}
where $\boldsymbol{\eta}$ represents the soft assignment weights over the $E$ experts for a given input. 
The final output is obtained via gated aggregation of expert outputs, followed by two linear layers to project the result to the target shape:
\begin{equation}
    \ddot{\mathbf{V}} = \mathbf{W}_{o, 1} \left(\mathbf{W}_{o, 2} \left(\text{Flatten}\left( \sum\nolimits_{e=1}^{E}  \eta_{e} \cdot \ddot{\mathbf{Z}}_{e} \right)\right)^{\top} \right),
\end{equation}
where $\mathbf{W}_{o, 1}, \mathbf{W}_{o, 2}$ are learnable weighting matrices, and $\ddot{\mathbf{V}} \in \mathbb{R}^{T_p \times C}$ is the final output of the Wavelet-view expert branch. $\eta_{e} \in \mathbb{R}$ is the gating weight for expert $e$.

\subsection{Multi-Resolution Adaptive Fusion Module}

The MAF module consists of two key phases: (1) a multi-view fusion phase and (2) a multi-resolution fusion phase, as illustrated in \textbf{Fig.~\ref{fig:overview}(e)}.

In the \textbf{multi-view fusion} phase, the temporal outputs from the Fourier and Wavelet expert branches, denoted as $\{\tilde{\mathbf{V}}, \ddot{\mathbf{V}} \} \in \mathbb{R}^{T_p \times C}$, are concatenated along the channel axis with temporal encoding $\mathbf{E} \in \mathbb{R}^{T_p \times 2}$. Then, the fused representation is processed by a stacked projection block with two linear layers and batch normalization:
\begin{equation}
\mathbf{H}_{u}^{(i)} = \mathbf{W}_{u, 2} \cdot \mathcal{D}\left(\text{ReLU}\left(\text{BN}\left(\mathbf{W}_{u, 1} [\tilde{\mathbf{V}}^{(i)}; \ddot{\mathbf{V}}^{(i)}; \mathbf{E}]^\top\right)\right)\right),
\end{equation}
where $\mathbf{W}_{u,1} \in \mathbb{R}^{H' \times (2C + 2)}$ and $\mathbf{W}_{u,2} \in \mathbb{R}^{C \times H'}$ are learnable weights; $H'$ is the hidden dimension; 
$\text{BN}(\cdot)$ denotes batch normalization;
and $[\cdot;\cdot]$ indicates channel-wise concatenation. The output $\mathbf{H}_u^{(i)} \in \mathbb{R}^{T_p \times C}$ represents the unified feature at the $i$-th resolution, with $i \in \{1, 2, \ldots, R\}$, where $R$ is the total number of resolutions.

In the \textbf{multi-resolution fusion} phase, representations from different resolutions are projected into a shared space and combined via additive accumulation:
\begin{equation}
    \mathbf{H}_{r} = \sum\nolimits_{i=1}^{R} \text{Linear}_i(\mathbf{H}_u^{(i)}) \in \mathbb{R}^{T_p \times C},
\end{equation}
where $\text{Linear}_i(\cdot)$ is a resolution-specific linear transformation. Since all frequency-view representations are learned from differenced sequences, the final fused output is shifted by adding back the last observed input slice to restore the original value space. This process enables coarse-to-fine feature refinement and enhances the integration of multi-scale temporal dynamics.

\subsection{Temporal Gating Integration Module}
The TGI module adaptively combines the recent prediction and historical scene representation to produce the final output, as illustrated in \textbf{Fig.~\ref{fig:overview}(f)}.
To adaptively integrate the recent prediction and the historical scene representation, a gating mechanism is applied. Let $\mathbf{H}_{r} \in \mathbb{R}^{T_p \times C}$ denote the output from the multi-resolution fusion module, and $\mathbf{H}_{h} \in \mathbb{R}^{T_p \times C}$ be the transformed embedding of the historical input, obtained via a linear projection $\mathbf{H}_{h} \leftarrow \mathbf{W}_g \mathbf{X}$ with $\mathbf{W}_g \in \mathbb{R}^{T_{in} \times T_p}$.
The gating coefficient is computed as:
\begin{align}
\mathbf{G} &= \sigma\left(\text{Linear}\left([\mathbf{H}_{r}; \mathbf{H}_{h}]\right)\right), \\
\hat{\mathbf{X}} &= \mathbf{G} \odot \mathbf{H}_{r} + (1 - \mathbf{G}) \odot \mathbf{H}_{h},
\end{align}
where $\sigma(\cdot)$ denotes the sigmoid activation. $\hat{\mathbf{X}} \in \mathbb{R}^{T_p \times C}$ is the final output of the model.

\subsection{Optimization Objective}

The overall objective of M$^2$FMoE consists of three components. The primary term is the forecasting loss $\mathcal{L}_{\text{pred}}$, measured by Mean Squared Error (MSE). To promote diverse specialization within each branch and ensure consistency across branches, we introduce a regularization term comprising the expert diversity loss $\mathcal{L}_{\text{div}}$ and expert consistency loss $\mathcal{L}_{\text{cons}}$:
\begin{align}
    &\mathcal{L}_{\text{div}} = \sqrt{ \frac{1}{E} \sum\nolimits_{e=1}^{E} \left( \|\mathbf{Z}_e\|_2 - \frac{1}{E} \sum\nolimits_{j=1}^{E} \|\mathbf{Z}_j\|_2 \right)^2 }, \\
    &\mathcal{L}_{\text{cons}} = \frac{1}{E} \sum\nolimits_{e=1}^{E} \bigg( 1 - {cossim}\left( \tilde{\mathbf{Z}}_e,\; \ddot{\mathbf{Z}}_e \right) \bigg),
\end{align}
where $\mathbf{Z}_e \in \{ \tilde{\mathbf{Z}}_e, \ddot{\mathbf{Z}}_e \}$ is the output of the $e$-th expert in either the Fourier-view or Wavelet-view expert branch, $\|\cdot\|_2$ denotes the $\ell_2$ norm, and ${cossim}(\cdot, \cdot)$ denotes the cosine similarity function.
Here, $\tilde{\mathbf{Z}}_e$ indicates the inverse FFT of the masked frequency component $\mathcal{F}_e$ for the $e$-th expert, and $\ddot{\mathbf{Z}}_e$ is the output of the $e$-th expert in the Wavelet-view expert branch. 
The expert diversity loss $\mathcal{L}_{\text{div}}$ encourages the outputs of different experts to be diverse, while the expert consistency loss $\mathcal{L}_{\text{cons}}$ encourages the outputs of the same expert in different branches to be consistent.
Finally, the overall optimization objective is defined as:
\begin{equation}
    \mathcal{L}_{\text{total}} = \mathcal{L}_{\text{pred}} + \lambda \mathcal{L}_{\text{div}} + \mu \mathcal{L}_{\text{cons}},
\end{equation}
where $\lambda$ and $\mu$ are hyperparameters that control the trade-off between the forecasting loss and the regularization terms.

\begin{table*}[t!]
\centering
\small
\setlength{\tabcolsep}{1mm}
    \begin{tabular}{ccc|cccccccccc}
    \toprule
    \multirow{2}[4]{*}{\textbf{Data}} & \multirow{2}[4]{*}{\textbf{Metrics}} & \multirow{2}[4]{*}{\textbf{Horizon}} & \multicolumn{8}{c|}{\textit{ without extreme lables}}         & \multicolumn{2}{c}{\textit{ with extreme lables}} \\
\cmidrule{4-13}          &       &       & \multicolumn{1}{c|}{\textbf{M$^2$FMoE}} & \textbf{CATS} & \textbf{CycleNet} & \textbf{FreqMoE} & \textbf{iTrans.} & \textbf{KAN} & \textbf{TQNet} & \multicolumn{1}{c|}{\textbf{Umixer}} & \textbf{DAN} & \textbf{MCANN} \\
    \midrule
    \multirow{4}[4]{*}{\begin{sideways}\textbf{Almaden}\end{sideways}} & RMSE  & \multirow{2}[2]{*}{8} & \multicolumn{1}{c|}{\textbf{7.990 }} & 16.087  & 17.754  & 14.729  & 32.127  & 18.934  & 18.023  & \multicolumn{1}{c|}{18.658 } & 37.857  & \underline{8.447}  \\
          & MAPE  &       & \multicolumn{1}{c|}{\textbf{0.002 }} & \underline{0.006}  & 0.007  & 0.005  & 0.017  & 0.009  & 0.010  & \multicolumn{1}{c|}{0.007 } & 0.021  & \textbf{0.002 } \\
\cmidrule{2-13}          & RMSE  & \multirow{2}[2]{*}{72} & \multicolumn{1}{c|}{\textbf{54.120 }} & 57.916  & 61.379  & 63.038  & 65.325  & 70.181  & 59.427  & \multicolumn{1}{c|}{64.816 } & 66.597  & \underline{56.840}  \\
          & MAPE  &       & \multicolumn{1}{c|}{\textbf{0.015 }} & \textbf{0.015 } & 0.019  & \underline{0.017}  & 0.025  & 0.033  & 0.018  & \multicolumn{1}{c|}{0.018 } & 0.025  & \textbf{0.015 } \\
    \midrule
    \multirow{4}[4]{*}{\begin{sideways}\textbf{Coyote}\end{sideways}} & RMSE  & \multirow{2}[2]{*}{8} & \multicolumn{1}{c|}{\textbf{48.797 }} & 110.849  & 113.706  & 593.141  & 372.523  & 116.398  & 103.521  & \multicolumn{1}{c|}{174.892 } & 505.941  & \underline{86.829}  \\
          & MAPE  &       & \multicolumn{1}{c|}{\textbf{0.002 }} & 0.004  & \underline{0.003}  & 0.018  & 0.022  & 0.005  & \underline{0.003}  & \multicolumn{1}{c|}{0.005 } & 0.025  & \textbf{0.002 } \\
\cmidrule{2-13}          & RMSE  & \multirow{2}[2]{*}{72} & \multicolumn{1}{c|}{\textbf{449.944 }} & 509.077  & 528.962  & 855.096  & 673.853  & 587.132  & \underline{504.606}  & \multicolumn{1}{c|}{566.429 } & 829.623  & 559.747  \\
          & MAPE  &       & \multicolumn{1}{c|}{\textbf{0.012 }} & \textbf{0.012 } & \textbf{0.012 } & 0.025  & 0.029  & 0.021  & \textbf{0.012 } & \multicolumn{1}{c|}{\underline{0.013} } & 0.042  & \textbf{0.012 } \\
    \midrule
    \multirow{4}[4]{*}{\begin{sideways}\textbf{Lexington}\end{sideways}} & RMSE  & \multirow{2}[2]{*}{8} & \multicolumn{1}{c|}{\textbf{251.957 }} & 618.991  & 463.293  & 386.995  & 690.426  & 429.054  & 400.991  & \multicolumn{1}{c|}{466.669 } & 476.936  & \underline{252.965}  \\
          & MAPE  &       & \multicolumn{1}{c|}{\underline{0.004}}  & 0.011  & 0.011  & 0.006  & 0.041  & 0.008  & 0.013  & \multicolumn{1}{c|}{0.008 } & 0.015  & \textbf{0.003 } \\
\cmidrule{2-13}          & RMSE  & \multirow{2}[2]{*}{72} & \multicolumn{1}{c|}{\textbf{772.836 }} & 906.531  & 865.092  & 1003.818  & 960.652  & 956.134  & 860.456  & \multicolumn{1}{c|}{829.541 } & 908.308  & \underline{778.023}  \\
          & MAPE  &       & \multicolumn{1}{c|}{\textbf{0.014 }} & 0.020  & 0.021  & 0.018  & 0.048  & 0.020  & 0.025  & \multicolumn{1}{c|}{0.018 } & 0.024  & \underline{0.015}  \\
    \midrule
    \multicolumn{1}{c}{\multirow{4}[4]{*}{\begin{sideways}\textbf{\shortstack{Stevens \\ Creek}}\end{sideways}}} & RMSE  & \multirow{2}[2]{*}{8} & \multicolumn{1}{c|}{\textbf{10.559 }} & 18.500  & 28.400  & 80.937  & 48.876  & 25.672  & 24.475  & \multicolumn{1}{c|}{37.654 } & 24.319  & \underline{12.130}  \\
          & MAPE  &       & \multicolumn{1}{c|}{\textbf{0.002 }} & \underline{0.004}  & 0.005  & 0.017  & 0.010  & 0.005  & 0.006  & \multicolumn{1}{c|}{0.007 } & 0.011  & \textbf{0.002 } \\
\cmidrule{2-13}          & RMSE  & \multirow{2}[2]{*}{72} & \multicolumn{1}{c|}{\textbf{76.939 }} & 82.739  & 94.578  & 117.282  & 106.606  & 94.034  & 89.265  & \multicolumn{1}{c|}{141.505 } & 82.794  & \underline{81.084}  \\
          & MAPE  &       & \multicolumn{1}{c|}{0.014}  & \textbf{0.011 } & 0.014  & 0.025  & 0.017  & 0.015  & \underline{0.012}  & \multicolumn{1}{c|}{0.017 } & 0.020  & \textbf{0.011 } \\
    \midrule
    \multirow{4}[4]{*}{\begin{sideways}\textbf{Vasona}\end{sideways}} & RMSE  & \multirow{2}[2]{*}{8} & \multicolumn{1}{c|}{\textbf{5.129 }} & 6.913  & 7.903  & 14.318  & 12.179  & 11.308  & 7.741  & \multicolumn{1}{c|}{9.299 } & 9.562  & \underline{5.353}  \\
          & MAPE  &       & \multicolumn{1}{c|}{\textbf{0.004 }} & \underline{0.007}  & \underline{0.007}  & 0.020  & 0.013  & 0.019  & \underline{0.007}  & \multicolumn{1}{c|}{0.009 } & 0.012  & \textbf{0.004 } \\
\cmidrule{2-13}          & RMSE  & \multirow{2}[2]{*}{72} & \multicolumn{1}{c|}{\underline{19.571}}  & 20.381  & 20.713  & 20.740  & 21.534  & 21.605  & 20.173  & \multicolumn{1}{c|}{23.718 } & 20.542  & \textbf{18.634 } \\
          & MAPE  &       & \multicolumn{1}{c|}{0.021}  & 0.021  & 0.021  & 0.027  & 0.023  & 0.027  & \underline{0.020}  & \multicolumn{1}{c|}{0.023 } & 0.023  & \textbf{0.019 } \\
    \midrule
    \multicolumn{3}{c|}{Average Rank ~/~ Significance} & \multicolumn{1}{c|}{\textbf{1.4 }} & 3.7~/~$\star$   & 4.9~/~$\ast$   & 7.3~/~$\ast$   & 8.5~/~$\ast$   & 7.0~/~$\ast$   & 4.4~/~$\ast$   & \multicolumn{1}{c|}{6.3~/~$\ast$}   & 7.9~/~$\ast$   & \underline{1.7} ~/~$\star$ \\
    \bottomrule
    \end{tabular}
  \caption{Performance comparison on five reservoirs with predicted length as \{8, 72\} hours. 
  $\ast$: both metrics are statistically significant ($p<0.05$, Wilcoxon signed-rank test); $\star$: indicates significance in RMSE. Best results are \textbf{bold}, second-best \underline{underlined}.}
  \label{tab:main_results}
\end{table*}

\section{Experiments}
\label{sec:experiments}

\subsection{Experimental Settings}

\subsubsection{Datasets} 
The experiment uses five public datasets containing hourly water level records from reservoirs in Santa Clara County, California. The datasets include Almaden, Coyote, Lexington, Stevens Creek, and Vasona, spanning the period from 1991 to 2019. 
Following the experimental protocol of \cite{journal/tpami2025/li6888}, the training and validation sets are randomly sampled from data between January 1991 and June 2018. The forecasting task targets the period from July 2018 to June 2019.
To alleviate the data imbalance, we employed the same oversampling strategy as described in \cite{journal/tpami2025/li6888}. 
The detailed description of the datasets is summarized in \textbf{Appendix B}.

\subsubsection{Benchmarks}
To ensure a comprehensive evaluation, nine representative state-of-the-art baselines are selected: attention-based models (CATS \cite{conference/neurips2024/kim}, TQNet \cite{conference/icml2025/lin}, iTransformer (iTrans.) \cite{conference/iclr2024/liu}), frequency-domain models (FreqMoE \cite{conference/aistats2025/liu25i}, Umixer \cite{conference/aaai2024/ma}), linear-based models (KAN \cite{conference/iclr2025/liu2}, CycleNet \cite{conference/nips2024/106315lin}), and two extreme-enhanced methods that leverage event labels (DAN \cite{conference/aaai2024/li27768}, MCANN \cite{journal/tpami2025/li6888}).

\subsubsection{Implementation Details}

For fair evaluation, the optimal configurations from the official implementations of MCANN and DAN are adopted. Following the experimental protocol in \cite{journal/tpami2025/li6888}, the prediction horizons are set to 8 and 72 hours, with a look-back window of 360 hours (i.e., 15 days). For other methods, standard baseline practices are followed: all datasets are normalized using z-score normalization, and denormalization is applied during evaluation to ensure predictions are in the original scale.
The models are trained using the Adam \cite{conference/iclr2015/Kingma} optimizer with a batch size of 48. 
Following the official protocol, evaluation is performed using Root Mean Squared Error (RMSE) and Mean Absolute Percentage Error (MAPE). 
The detailed settings are provided in \textbf{Appendix C}.

\begin{figure}[t]
    \centering
    \includegraphics[width=0.5\columnwidth]{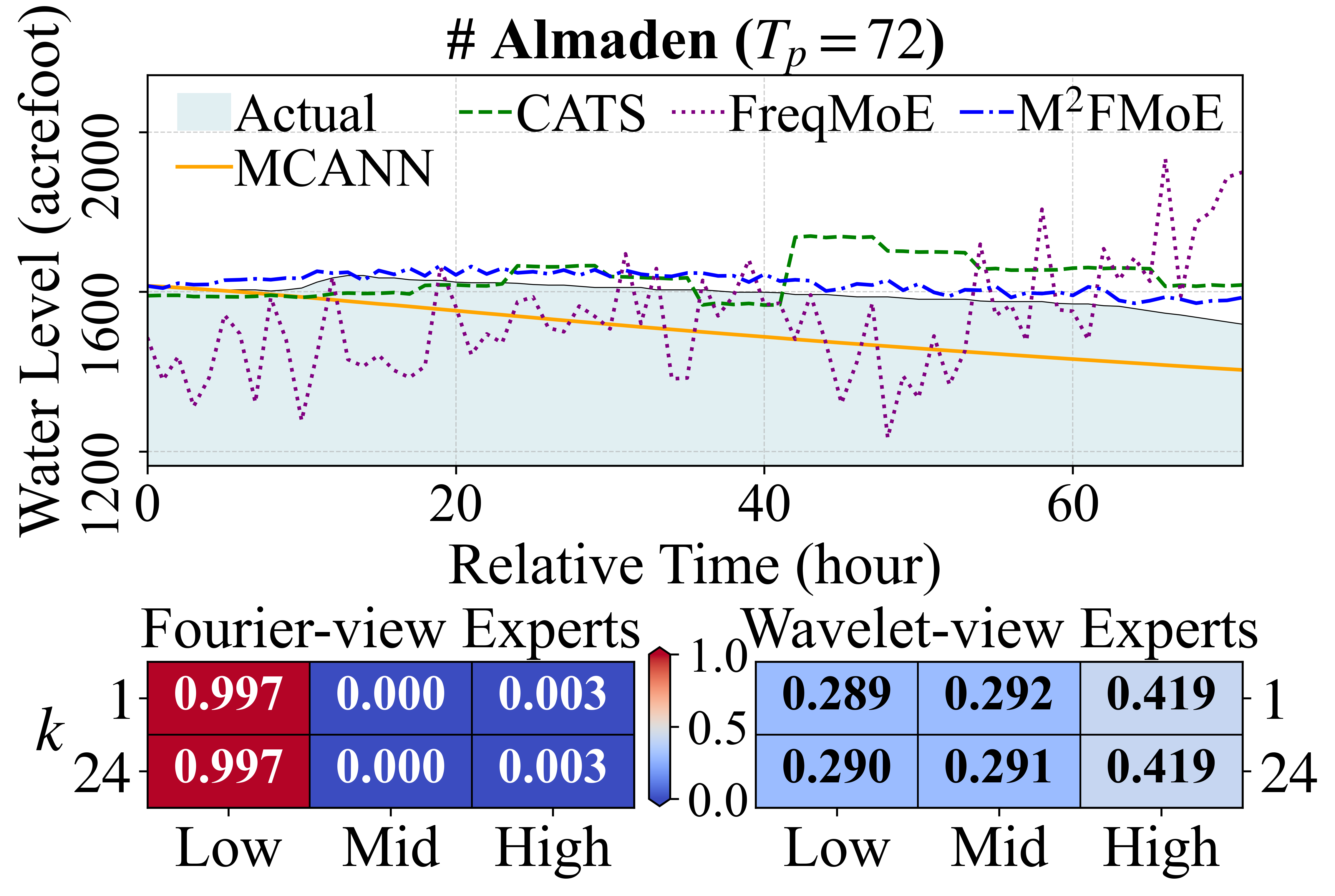}~
    \includegraphics[width=0.5\columnwidth]{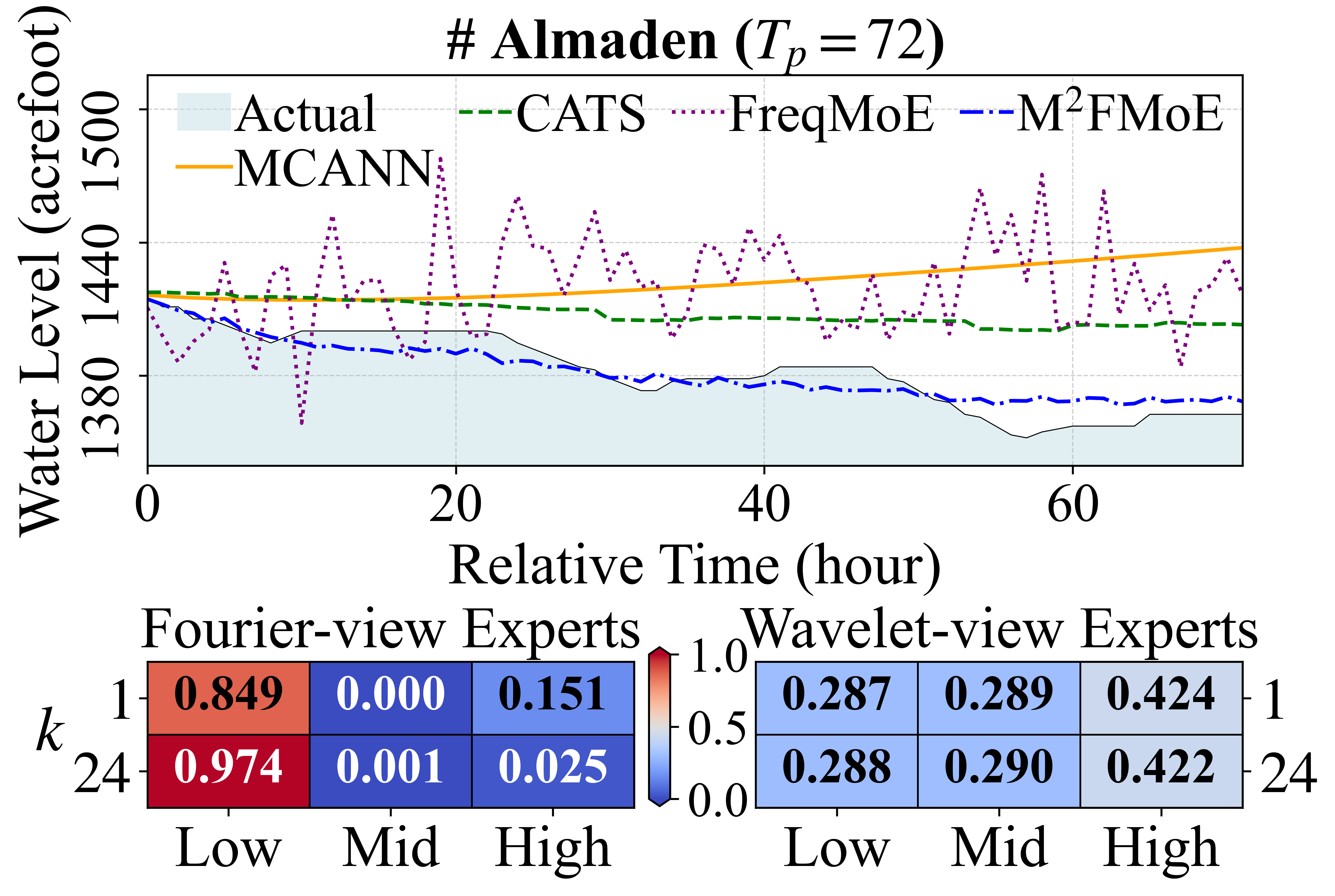}
    \includegraphics[width=0.5\columnwidth]{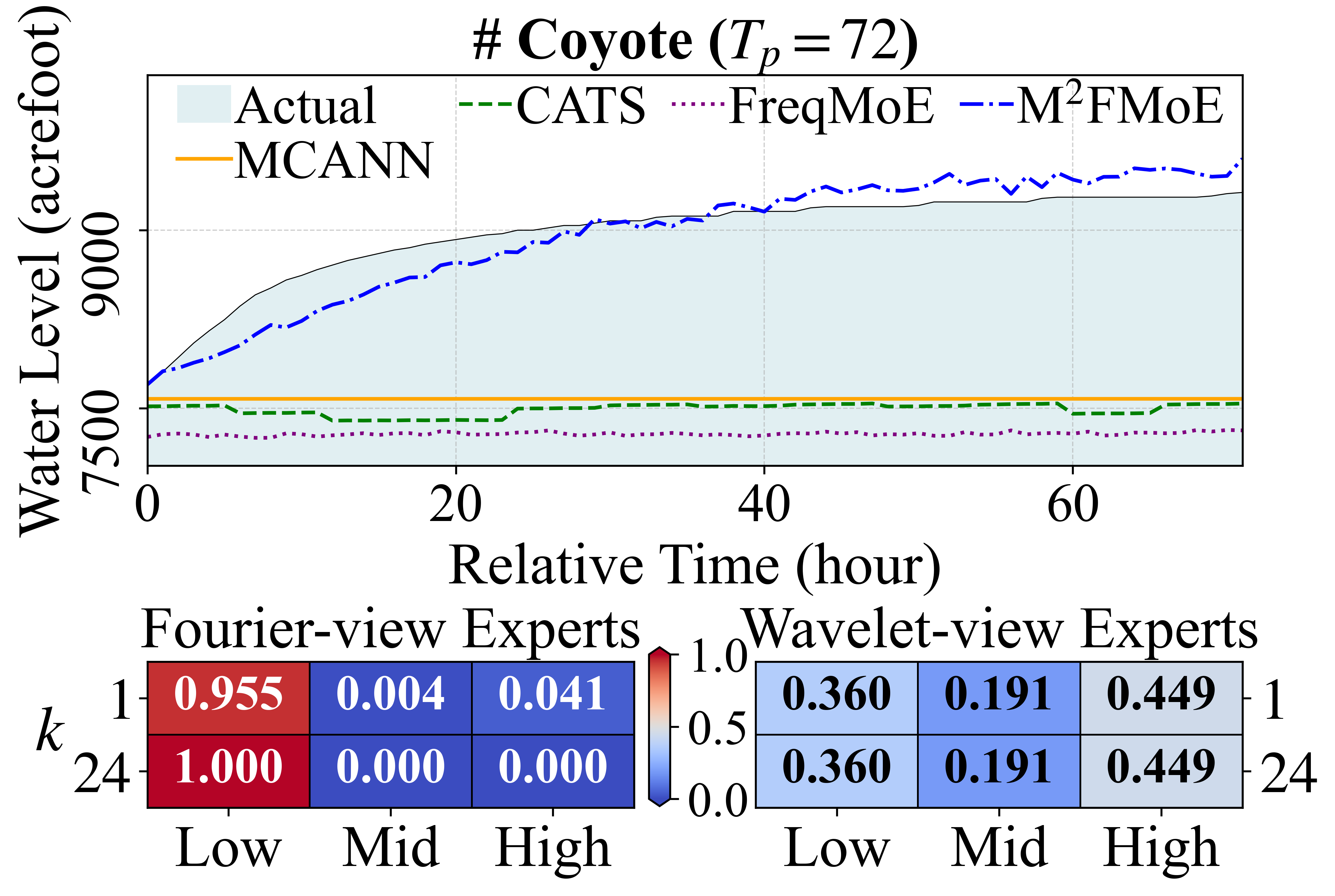}~
    \includegraphics[width=0.5\columnwidth]{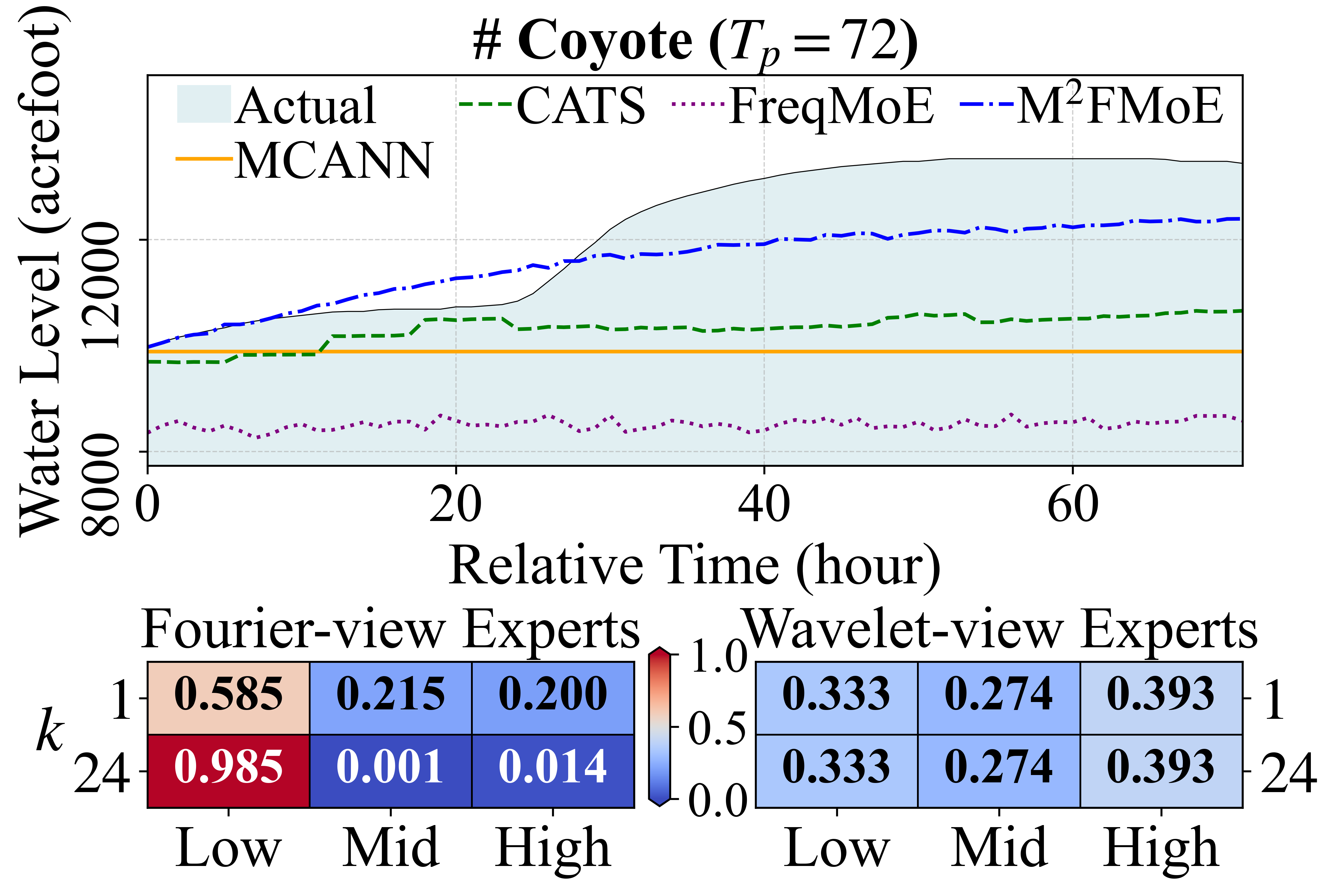}
    \caption{Prediction results and expert weights of M$^2$FMoE.}
    \label{fig:pre_comparison}
\end{figure}

\begin{table*}[t]
  \centering
  \small

    \resizebox{1.0\linewidth}{!}{	
    \begin{tabular}{lcccccccccc}
    \toprule
    \multirow{2}[4]{*}{\textbf{Model}} & \multicolumn{2}{c}{\textbf{Almaden}} & \multicolumn{2}{c}{\textbf{Coyote}} & \multicolumn{2}{c}{\textbf{Lexington}} & \multicolumn{2}{c}{\textbf{Stevens Creek}} & \multicolumn{2}{c}{\textbf{Vasona}} \\
\cmidrule{2-11}          & RMSE  & MAPE  & RMSE  & MAPE  & RMSE  & MAPE  & RMSE  & MAPE  & RMSE  & MAPE \\
    \midrule
    \textbf{M$^2$FMoE} & \textbf{54.120 } & \textbf{0.015 } & \underline{449.944}  & \textbf{0.012 } & \textbf{772.836 } & \textbf{0.014 } & \textbf{76.939 } & \underline{0.014}  & {19.571}  & \underline{0.021 } \\
    \midrule
    \textit{w/o-WaveletView} & 57.697  & \underline{0.016}  & 555.641  & 0.014  & 827.392  & 0.016  & 87.194  & 0.016  & 19.836  & {0.022}  \\
    \textit{w/o-FourierView} & 59.035  & 0.020  & 558.262  & 0.014  & 870.735  & 0.017  & 85.022  & 0.017  & 19.813  & 0.024  \\
    \textit{w/o-}$\mathcal{L}_\text{div}$ \& $\mathcal{L}_\text{cons}$ & \underline{54.950}  & 0.017  & \textbf{448.150 } & \underline{0.013}  & 826.408  & \underline{0.015}  & \underline{77.293}  & \textbf{0.012 } & 20.080  & \underline{0.021 } \\
    \textit{w/o-Multi-Res} & 59.479  & 0.017  & 483.223  & \underline{0.013}  & 855.236  & 0.017  & 85.004  & 0.017  & {19.508 } & {0.022}  \\
    \textit{w/o-CSS} & 55.575  & 0.017  & 541.594  & \underline{0.013}  & 916.925  & 0.016  & 85.997  & 0.019  & 20.001  & \underline{0.021 } \\
    \textit{w/o-Alignment} & 59.115  & 0.019  & 516.731  & \textbf{0.012}  & 872.033  & 0.018  & 85.799  & 0.019  & \textbf{19.280}  & 0.022  \\
    \textit{w/o-DualView} & 56.269  & 0.020  & 453.460  & \textbf{0.012}  & \underline{789.044}  & \underline{0.015}  & 79.064  & 0.015  & \underline{19.377}  & \textbf{0.020}  \\
    \bottomrule
    \end{tabular}
    }
    \caption{Ablation study results on the five reservoirs with a prediction horizon of 72 hours.}   
    \label{tab:ablation}
\end{table*}

\subsection{Main Results}
\subsubsection{Comparison with Benchmarks}
As shown in \textbf{Table~\ref{tab:main_results}}, we compare the proposed M$^2$FMoE model with nine state-of-the-art baselines on five reservoirs with prediction horizons of 8 and 72 hours. The results demonstrate that M$^2$FMoE achieves the best average rank across all datasets and prediction horizons, outperforming all baselines in most cases. The improvements in RMSE are statistically significant on all reservoirs according to the Wilcoxon signed-rank test.
Specifically, M$^2$FMoE achieves the average improvement of 22.30\% over the best baseline without extreme labels and the maximum RMSE improvement of 52.86\% on the Coyote dataset with a prediction horizon of 8 hours. Compared to the baselines with extreme labels, M$^2$FMoE also achieves competitive performance, with an average RMSE improvement of 9.19\% across all settings, and a maximum improvement of 43.8\% on the Coyote dataset with a prediction horizon of 8 hours.
These results indicate that the proposed M$^2$FMoE model effectively captures the complex temporal dynamics of reservoir water levels, demonstrating its superiority over existing methods.
\textbf{Fig.~\ref{fig:pre_comparison}} presents the prediction results and expert weights of M$^2$FMoE, using three experts under two temporal resolutions ($k{=}1$ and $k{=}24$). The results show that Fourier-view experts primarily capture low-frequency trends, while Wavelet-view experts provide complementary high-frequency details. The adaptive weighting mechanism dynamically adjusts expert contributions based on input characteristics, improving M$^2$FMoE’s performance on both regular and extreme events.

\subsubsection{Ablation Studies}
We conduct ablation studies to evaluate the effectiveness of each component in the proposed M$^2$FMoE model. The ablation experiments are performed on the five reservoirs with a prediction horizon of 72 hours. The results are summarized in \textbf{Table~\ref{tab:ablation}}.
The ablation studies include the following variants: (1) \textit{w/o-WaveletView}: removes the Wavelet-view expert branch, (2) \textit{w/o-FourierView}: removes the Fourier-view expert branch, (3) \textit{w/o-}$\mathcal{L}_\text{div}$ \& $\mathcal{L}_\text{cons}$: removes the expert diversity and consistency losses, (4) \textit{w/o-Multi-Res}: utilizes the single-resolution and removes the multi-resolution fusion module, (5) \textit{w/o-CSS}: replace the cross-view shared band splitter with a uniform band splitter, (6) \textit{w/o-Alignment}: removes the alignment mechanism in the cross-view shared band splitter, and (7) \textit{w/o-DualView}: employs only a simple MLP for temporal modeling without the dual spectral views.
We observe that the full model achieves the optimal performance across almost all datasets, demonstrating the effectiveness of the proposed multi-view and multi-resolution fusion strategy.

\subsubsection{Impact of the Recent Segment Length}

\begin{figure}
\centering
\includegraphics[width=0.49\columnwidth]{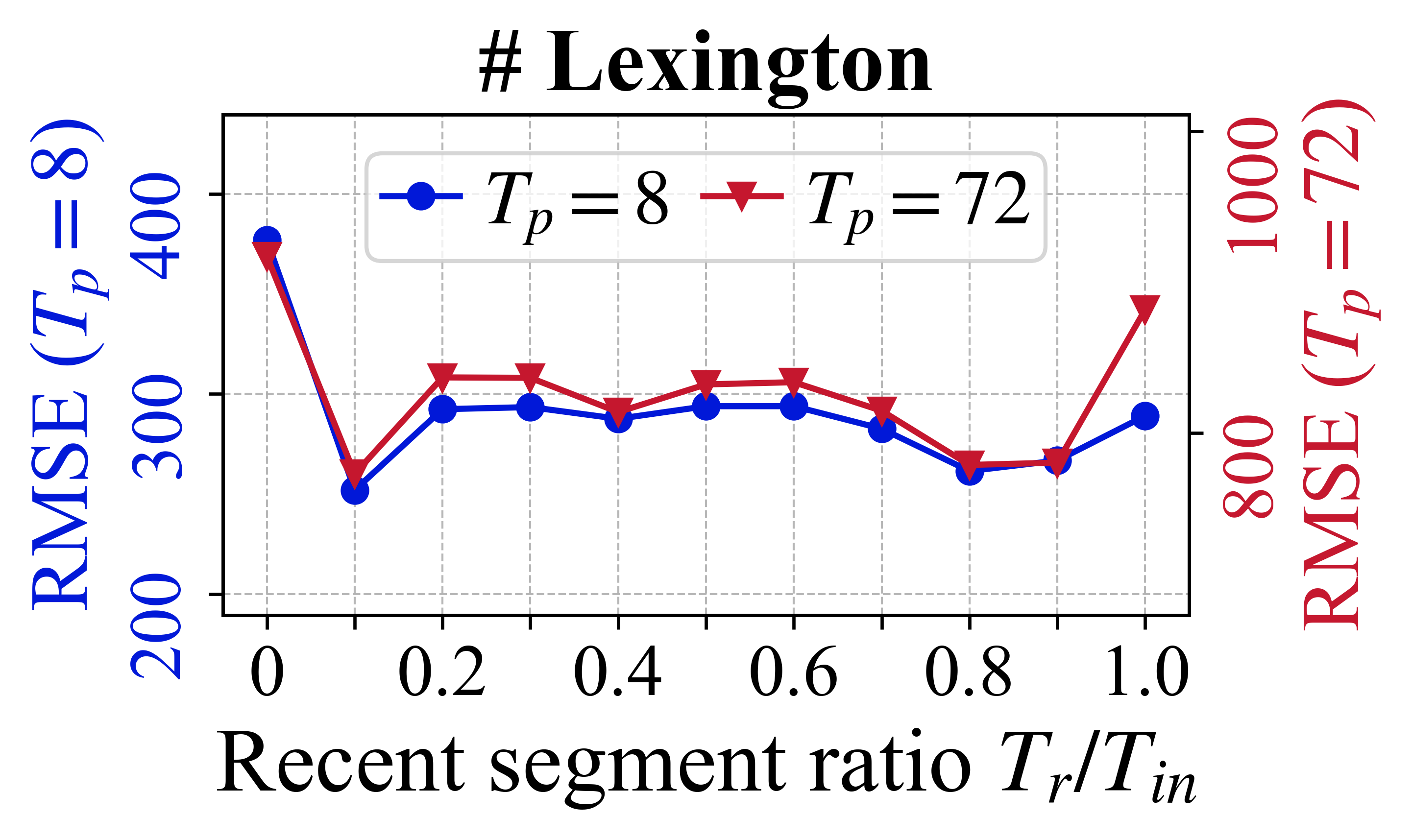}
\includegraphics[width=0.49\columnwidth]{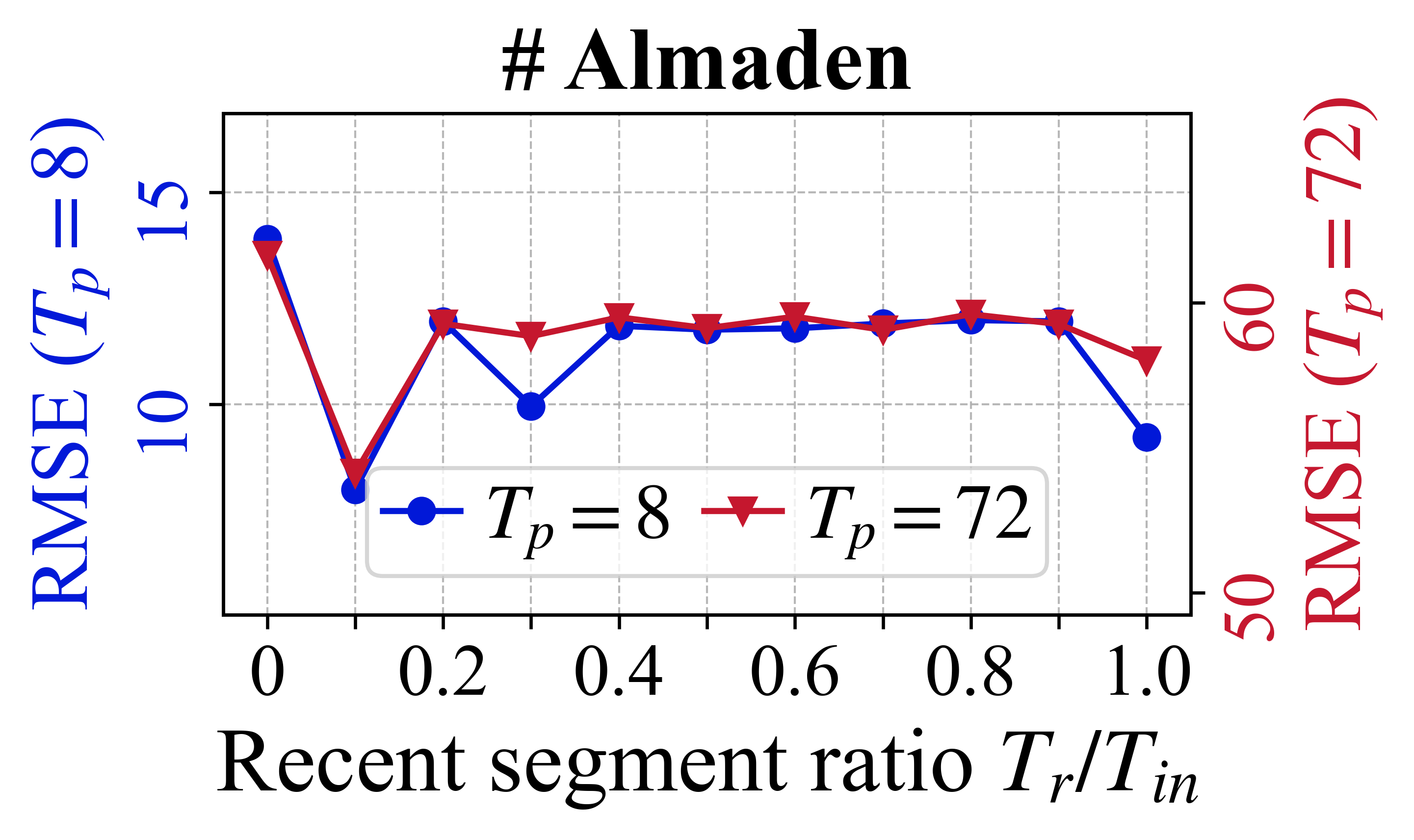}
\caption{Impact of the length of recent segment $T_r/T_{in}$.}
\label{fig:ts_rmse}
\end{figure}

The length of the recent segment critically affects the model's ability to capture extreme events. As shown in \textbf{Fig.~\ref{fig:ts_rmse}}, reducing its length appropriately improves prediction accuracy by emphasizing relevant information. However, removing it entirely causes a significant performance drop, while overly long segments introduce noise and weaken the model’s focus on extremes.

\subsubsection{Impact of the Number of Experts}

\begin{figure}
\centering
\includegraphics[width=0.49\columnwidth]{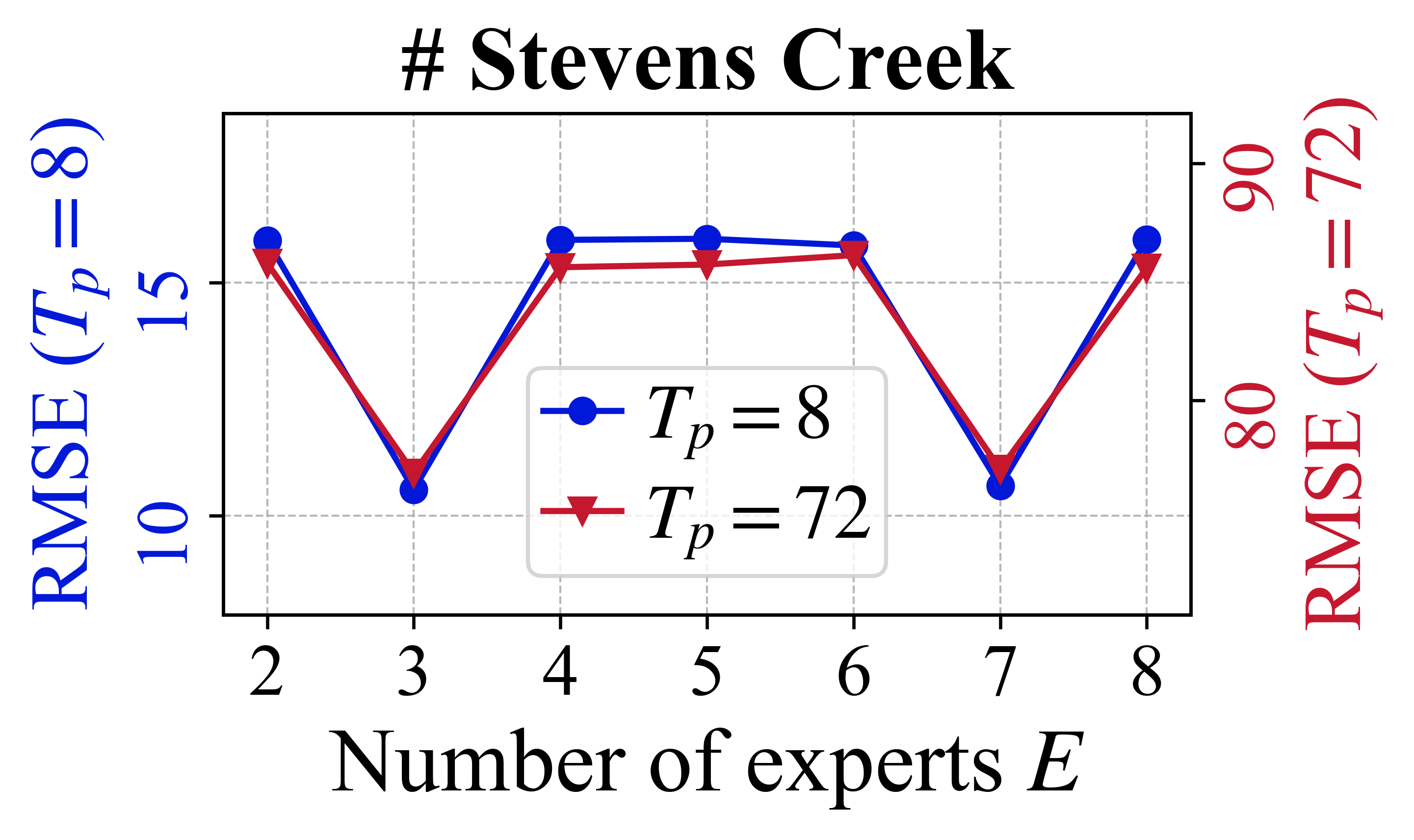}
\includegraphics[width=0.49\columnwidth]{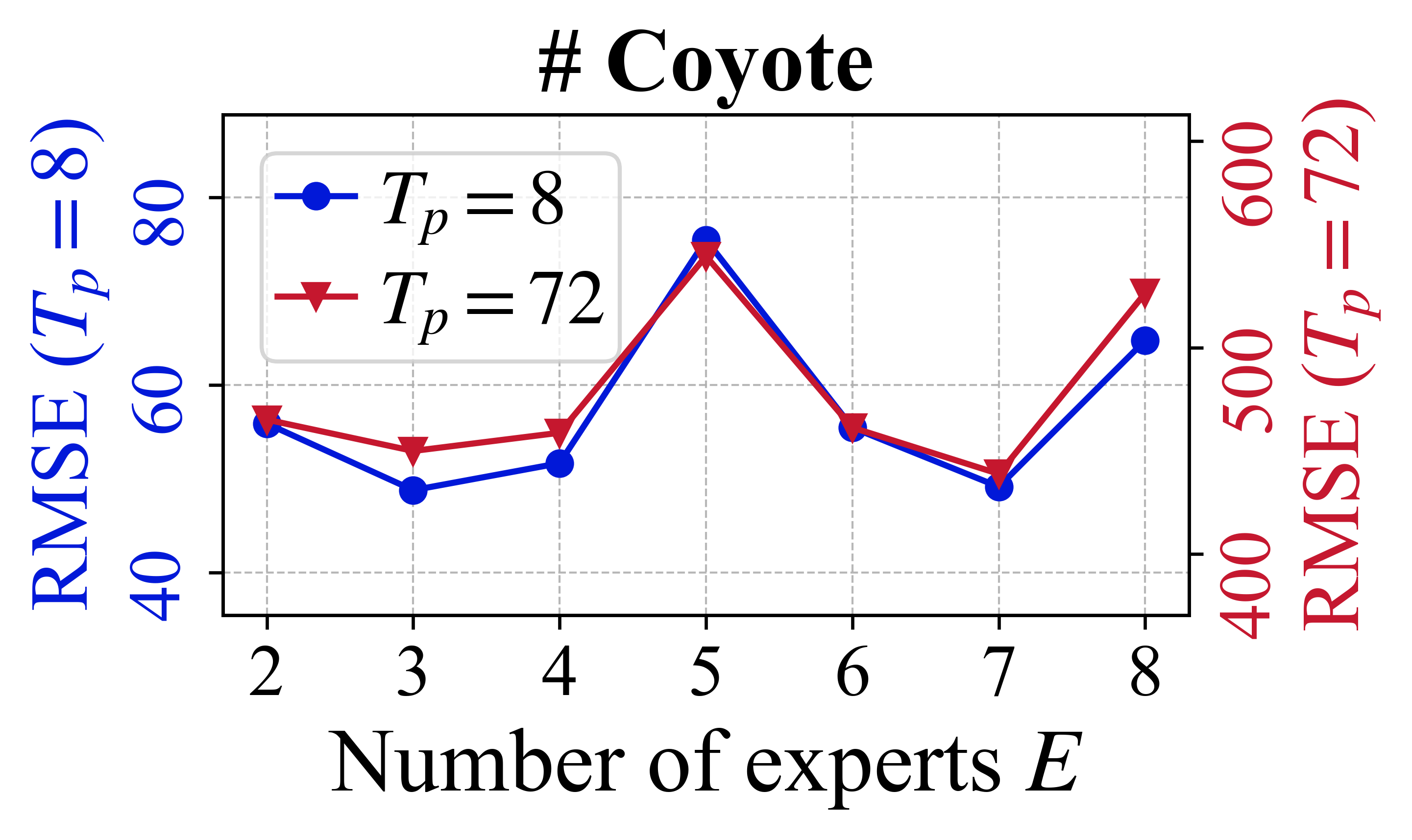}
\caption{Impact of the number of experts $E$.}
\label{fig:expert_rmse}
\end{figure}

We further examined the impact of expert count in M$^2$FMoE. As shown in \textbf{Fig.~\ref{fig:expert_rmse}}, increasing the number of experts may improve pattern diversity but can also introduce noise and overfitting, leading to performance instability. Results suggest that using a moderate number (e.g., 3 or 4) yields the best predictive accuracy.

\begin{figure}[t]
\centering

\includegraphics[width=0.49\columnwidth]{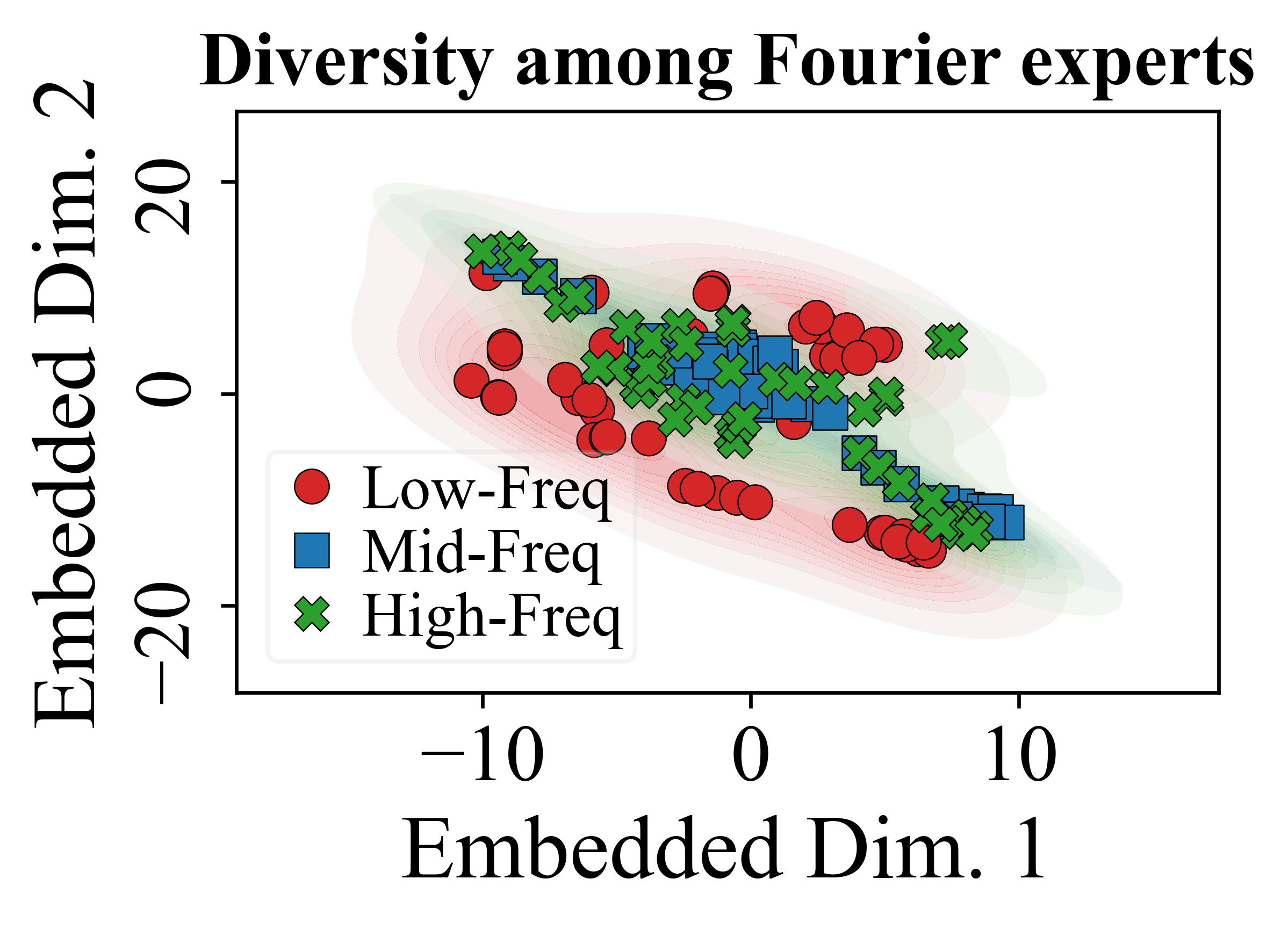}
\includegraphics[width=0.49\columnwidth]{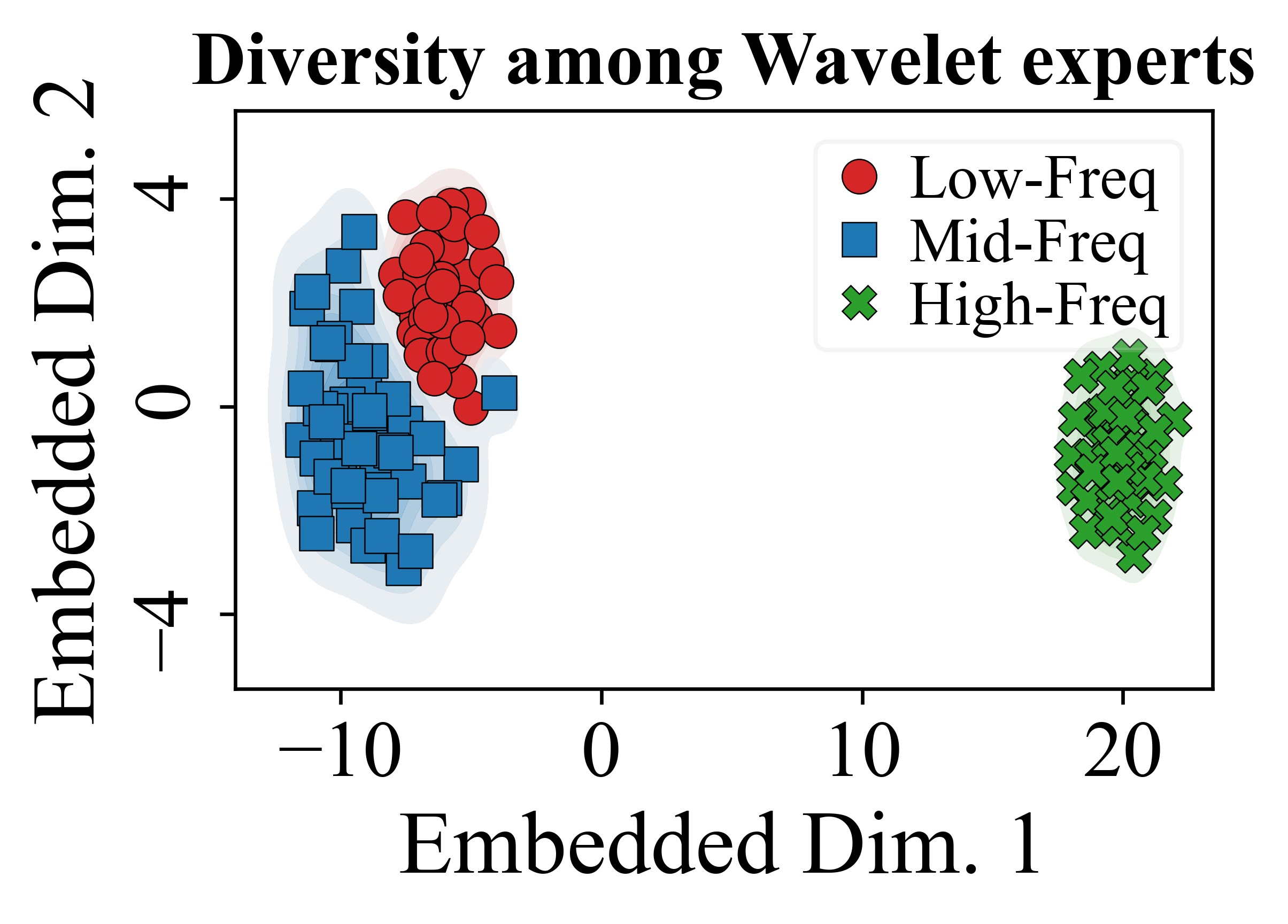}
\includegraphics[width=0.49\columnwidth]{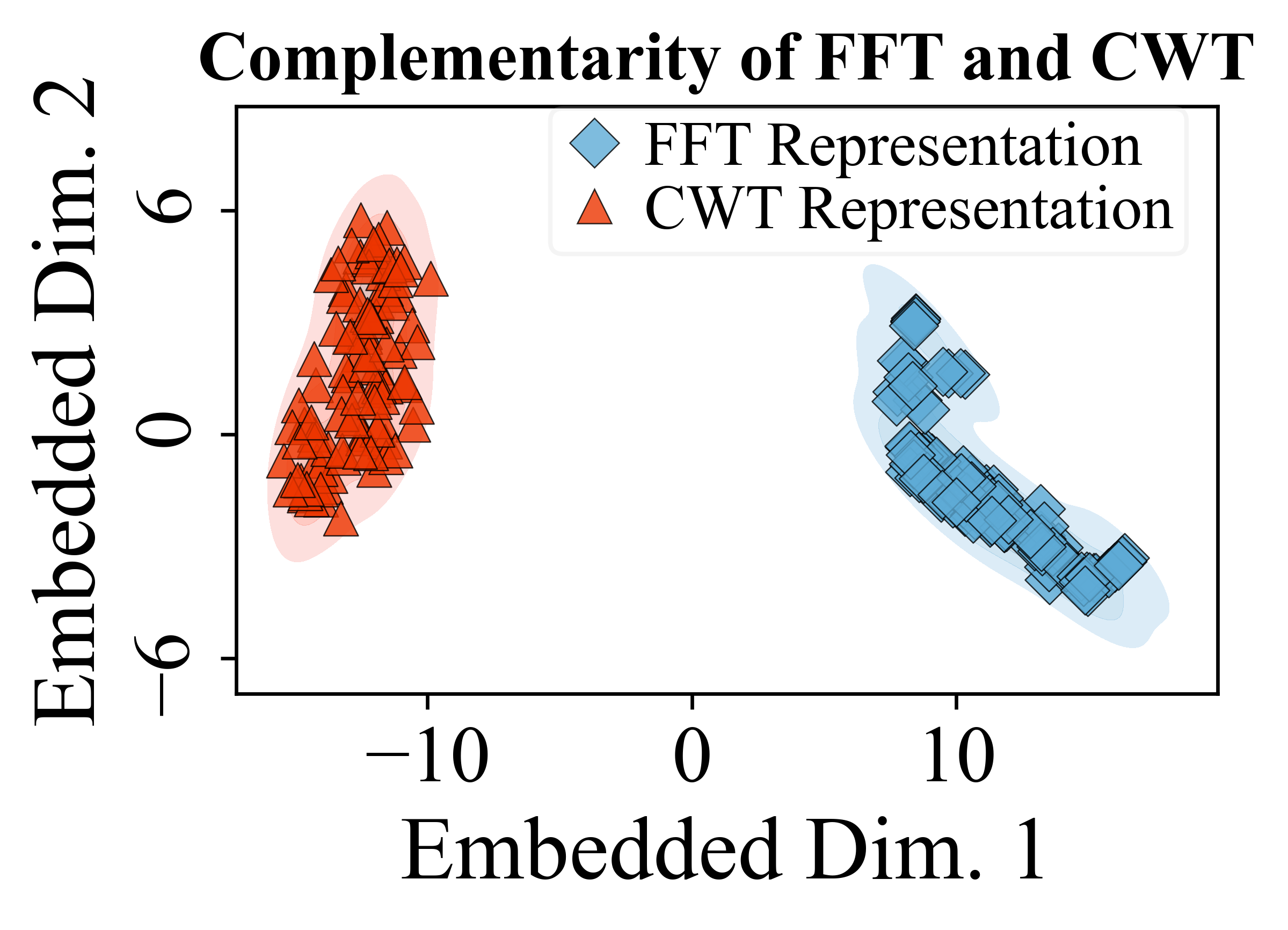}
\includegraphics[width=0.49\columnwidth]{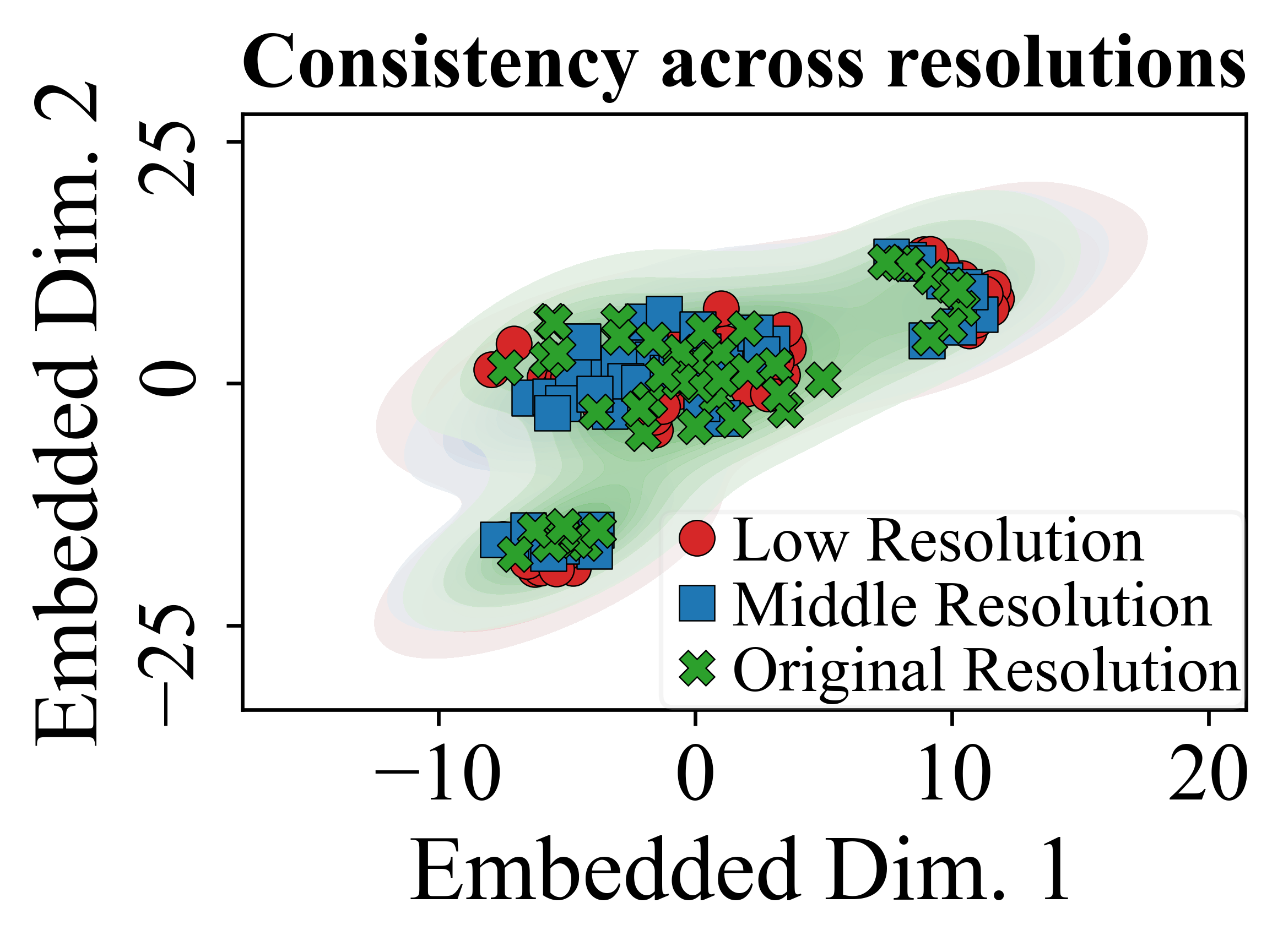}

\caption{The t-SNE visualization of feature representations.}
\label{fig:tsne}
\end{figure}

\subsubsection{Visualization Analysis}

To better interpret the feature representations learned by M$^2$FMoE, the t-SNE \cite{maaten2008visualizing} is employed to visualize expert embeddings trained on the Almaden dataset using three spectral experts corresponding to high-, mid-, and low-frequency bands, as shown in \textbf{Fig.~\ref{fig:tsne}}. The visualizations reveal the following insights:
(1) In the Fourier view, low-frequency features form a well-separated cluster from mid- and high-frequency features, indicating its effectiveness in capturing global trends.
(2) The Wavelet view exhibits clearer separation between mid- and high-frequency features, suggesting superior sensitivity to localized, sparse patterns. (3) The cross-view distribution highlights the complementary nature of the two spectral views, with distinct clustering structures in each domain. (4) The cross-resolution view demonstrates that the multi-resolution fusion module maintains consistency while also capturing local variations. These results validate the proposed multi-view and multi-resolution strategy for effectively modeling diverse temporal patterns.

\section{Conclusion}
\label{sec:conclusion}

This study proposes M$^2$FMoE, an extreme-adaptive time series forecasting model that leverages multi-view frequency learning and multi-resolution fusion to capture both global trends and local extreme variations.
Specifically, M$^2$FMoE employs specialized Fourier- and Wavelet-based experts to extract multi-frequency representations, while a multi-resolution fusion module progressively integrates temporal dependencies across resolutions. The model is optimized using forecasting, diversity, and consistency losses to promote adaptive and complementary expert behavior.
Experiments on five reservoir datasets show that M$^2$FMoE outperforms state-of-the-art methods without using extreme event labels.

\section{Acknowledgments}
This work was supported in part by the National Natural Science Foundation of China under Grant 62376289, in part by the Natural Science Foundation of Hunan Province, China under Grant 2024JJ4069, and in part supported by the Fundamental Research Funds for the Central Universities of Central South University.

\bibliography{aaai2026ref}

@inproceedings{conference/ijcai2024/wang5135,
  author       = {Yiyang Wang and
                  Yuchen Han and
                  Yuhan Guo},
  title        = {Self-adaptive Extreme Penalized Loss for Imbalanced Time Series Prediction},
  booktitle    = {Proceedings of the 33rd International Joint Conference on Artificial Intelligence},
  //booktitle    = {IJCAI},
  pages        = {5135--5143},
  //publisher    = {ijcai.org},
  year         = {2024},
  month        = {Aug},
  address      = {Jeju, South Korea},
  //url          = {https://www.ijcai.org/proceedings/2024/568},
}

@inproceedings{conference/icml2025/jin,
  title={{MOH}: Multi-head attention as mixture-of-head attention},
  author={Jin, Peng and Zhu, Bo and Yuan, Li and Yan, Shuicheng},
  booktitle={Proceedings of the 42nd annual conference of the International Conference on Machine Learning},
  //booktitle={ICML},
  //address = {Vancouver, Canada},
  year={2025},
}

@inproceedings{conference/aaai2024/li27768,
  title={Learning from polar representation: An extreme-adaptive model for long-term time series forecasting},
  author={Li, Yanhong and Xu, Jack and Anastasiu, David},
  booktitle={Proceedings of the 38th AAAI Conference on Artificial Intelligence},
  //booktitle={AAAI},
  //volume={38},
  number={1},
  pages={171--179},
  year={2024},
  month = {Feb},
  address = {Vancouver, Canada},
  url = {https://doi.org/10.1609/aaai.v38i1.27768},
  doi = {10.1609/AAAI.V38I1.27768},
}

@inproceedings{conference/aaai2023/li26045,
  author    =   {Yanhong Li and Jack Xu and David C. Anastasiu},
  title     =   {An Extreme-Adaptive Time Series Prediction Model Based on Probability-Enhanced {LSTM} Neural Networks},
  booktitle =   {Proceedings of the 37th {AAAI} Conference on Artificial Intelligence},
  //booktitle =   {AAAI},
  pages     =   {8684--8691},
  //publisher =   {{AAAI} Press},
  year      =   {2023},
  //volume    =   {37},
  number    =   {7},
  address   =   {Washington, DC, USA},
  month     =   {Feb},
  url       =   {https://doi.org/10.1609/aaai.v37i7.26045},
  doi       =   {10.1609/AAAI.V37I7.26045},
}

@ARTICLE{journal/tpami2025/li6888,
  author={Li, Yanhong and Anastasiu, David C.},
  journal={IEEE Transactions on Pattern Analysis and Machine Intelligence}, 
  title={MC-ANN: A Mixture Clustering-Based Attention Neural Network for Time Series Forecasting}, 
  year={2025},
  volume={47},
  number={8},
  pages={6888-6899},
  doi={10.1109/TPAMI.2025.3565224}
}

@article{journal/neurocom2023/zhang,
  title={Time series forecasting using a hybrid ARIMA and neural network model},
  author={Zhang, G Peter},
  journal={Neurocomputing},
  volume={50},
  pages={159--175},
  year={2003},
  //publisher={Elsevier}
}

@inproceedings{conference/sigkdd2019/Ding1114,
  author    =   {Ding, Daizong and Zhang, Mi and Pan, Xudong and Yang, Min and He, Xiangnan},
  title     =   {Modeling Extreme Events in Time Series Prediction},
  booktitle =   {Proceedings of the 25th {ACM} {SIGKDD} International Conference on Knowledge Discovery and Data Mining},
  //booktitle =   {KDD},
  publisher =   {Association for Computing Machinery},
  year      =   {2019},
  volume    =   {},
  number    =   {},
  pages     =   {1114–1122},
  address   =   {Anchorage, AK, USA},
  month     =   {Jul},
  url       =   {https://doi.org/10.1145/3292500.3330896},
  doi       =   {10.1145/3292500.3330896},
}

@inproceedings{conference/icassp2023/Liu1,
  author    =   {Liu, Hengbo and Ma, Ziqing and Yang, Linxiao and Zhou, Tian and Xia, Rui and Wang, Yi and Wen, Qingsong and Sun, Liang},
  title     =   {SADI: A Self-Adaptive Decomposed Interpretable Framework for Electric Load Forecasting Under Extreme Events},
  booktitle =   {Proceedings of the {ICASSP} 2023 - 2023 {IEEE} International Conference on Acoustics, Speech and Signal Processing},
  //booktitle =   {ICASSP},
  year      =   {2023},
  volume    =   {},
  number    =   {},
  pages     =   {1-5},
  address   =   {Rhodes Island, Greece},
  month     =   {Jun},
  url       =   {https://doi.org/10.1109/ICASSP49357.2023.10096002},
  doi       =   {10.1109/ICASSP49357.2023.10096002},
}

@inproceedings{conference/bigdata/Li728,
  author    =   {Li, Yanhong and Xu, Jack and Anastasiu, David C.},
  title     =   {SEED: An Effective Model for Highly-Skewed Streamflow Time Series Data Forecasting},
  booktitle =   {Proceedings of the 2023 {IEEE} International Conference on Big Data},
  //booktitle =   {BigData},
  publisher =   {{IEEE}},
  year      =   {2023},
  pages     =   {728-737},
  address   =   {Sorrento, Italy},
  month     =   {Aug},
  url       =   {https://doi.org/10.1109/BigData59044.2023.10386959},
  doi       =   {10.1109/BigData59044.2023.10386959},
}

@inproceedings{journal/aaai2025/Fei11645,
  author    = {Fei, Jingru and Yi, Kun and Fan, Wei and Zhang, Qi and Niu, Zhendong},
  booktitle={Proceedings of the 39th AAAI Conference on Artificial Intelligence}, 
  //booktitle={AAAI},
  title     = {Amplifier: Bringing Attention to Neglected Low-Energy Components in Time Series Forecasting}, 
  year      = {2025},
  volume    = {39},
  number    = {11},
  month     = {Apr},
  pages     = {11645-11653},
  doi       = {10.1609/aaai.v39i11.33267},
}

@article{journal/tkde2021/Zhang2021,
  title={Enhancing time series predictors with generalized extreme value loss},
  author={Zhang, Mi and Ding, Daizong and Pan, Xudong and Yang, Min},
  journal={IEEE Transactions on Knowledge and Data Engineering},
  volume={35},
  number={2},
  pages={1473--1487},
  year={2021},
  //publisher={IEEE},
}

@inproceedings{conference/aaai2021/Xiu10469,
  title={Variational disentanglement for rare event modeling},
  author={Xiu, Zidi and Tao, Chenyang and Gao, Michael and Davis, Connor and Goldstein, Benjamin A and Henao, Ricardo},
  booktitle={Proceedings of the 35th AAAI Conference on Artificial Intelligence},
  //booktitle={AAAI},
  //volume={35},
  number={12},
  pages={10469--10477},
  year={2021},
}

@InProceedings{conference/aistats2025/liu25i,
  title = 	 {FreqMoE: Enhancing Time Series Forecasting through Frequency Decomposition Mixture of Experts},
  author =       {Liu, Ziqi},
  booktitle = 	 {Proceedings of the 28th International Conference on Artificial Intelligence and Statistics},
  //booktitle = 	 {AISTATS},
  pages = 	 {3430--3438},
  year = 	 {2025},
  editor = 	 {Li, Yingzhen and Mandt, Stephan and Agrawal, Shipra and Khan, Emtiyaz},
  volume = 	 {258},
  //series = 	 {Proceedings of Machine Learning Research},
  month = 	 {03--05 May},
  publisher =    {PMLR},
  pdf = 	 {https://raw.githubusercontent.com/mlresearch/v258/main/assets/liu25i/liu25i.pdf},
  url = 	 {https://proceedings.mlr.press/v258/liu25i.html},
}

@inproceedings{conference/nips2024/106315lin,
  author       = {Shengsheng Lin and Weiwei Lin and Xinyi Hu and Wentai Wu and Ruichao Mo and Haocheng Zhong},
  title        = {CycleNet: Enhancing Time Series Forecasting through Modeling Periodic Patterns},
  booktitle    = {Proceedings of the 38th Annual Conference on Neural Information Processing Systems},
  //booktitle    = {NeurIPS},
  year         = {2024},
  //volume = {37},
  pages = {106315--106345},
  publisher = {Curran Associates, Inc.},
  month        = {Dec},
  address      = {Vancouver, BC, Canada},
  //url          = {http://papers.nips.cc/paper\_files/paper/2024/hash/bfe7998398779dde03cad7a73b1f81b6-Abstract-Conference.html},
}

@article{wang2024deep,
  title={Deep time series models: A comprehensive survey and benchmark},
  author={Wang, Yuxuan and Wu, Haixu and Dong, Jiaxiang and Liu, Yong and Long, Mingsheng and Wang, Jianmin},
  journal={arXiv preprint arXiv:2407.13278},
  year={2024}
}

@ARTICLE{journal/tpami2025/jin,
  author={Jin, Ming and Shi, Guangsi and Li, Yuan-Fang and Xiong, Bo and Zhou, Tian and Salim, Flora D. and Zhao, Liang and Wu, Lingfei and Wen, Qingsong and Pan, Shirui},
  journal={IEEE Transactions on Pattern Analysis and Machine Intelligence}, 
  title={Towards Expressive Spectral-Temporal Graph Neural Networks for Time Series Forecasting}, 
  year={2025},
  volume={47},
  number={6},
  pages={4926-4939},
  doi={10.1109/TPAMI.2025.3545671}
}

@inproceedings{conference/iclr2023/wu,
  title={TimesNet: Temporal 2D-Variation Modeling for General Time Series Analysis},
  author={Haixu Wu and Tengge Hu and Yong Liu and Hang Zhou and Jianmin Wang and Mingsheng Long},
  booktitle={Proceedings of the 11th International Conference on Learning Representations},
  //booktitle={ICLR},
  address = {Kigali, Rwanda},
  year={2023},
}

@inproceedings{conference/iclr2025/liu,
  title={Timer-XL: Long-Context Transformers for Unified Time Series Forecasting},
  author={Liu, Yong and Qin, Guo and Huang, Xiangdong and Wang, Jianmin and Long, Mingsheng},
  booktitle={Proceedings of the 13th International Conference on Learning Representations},
  //booktitle={ICLR},
  year={2025},
  address = {Singapore},
}

@inproceedings{conference/iclr2025/liu2,
  title={Kan: Kolmogorov-arnold networks},
  author={Liu, Ziming and Wang, Yixuan and Vaidya, Sachin and Ruehle, Fabian and Halverson, James and Solja{\v{c}}i{\'c}, Marin and Hou, Thomas Y and Tegmark, Max},
  booktitle={Proceedings of the 13th International Conference on Learning Representations},
  //booktitle={ICLR},
  year={2025},
  address = {Singapore},
}

@inproceedings{conference/iclr2024/liu,
  title={itransformer: Inverted transformers are effective for time series forecasting},
  author={Liu, Yong and Hu, Tengge and Zhang, Haoran and Wu, Haixu and Wang, Shiyu and Ma, Lintao and Long, Mingsheng},
  booktitle={Proceedings of the 12th International Conference on Learning Representations},
  //booktitle={ICLR},
  year={2024},
  address = {Vienna, Austria},
}

@inproceedings{conference/icml2025/lin,
title={Temporal Query Network for Efficient Multivariate Time Series Forecasting},
author={Shengsheng Lin and Haojun Chen and Haijie Wu and Chunyun Qiu and Weiwei Lin},
booktitle={Proceedings of the 42nd International Conference on Machine Learning},
//booktitle={ICML},
year={2025},
//url={https://openreview.net/forum?id=e24CueVty2}
}

@inproceedings{conference/aaai2024/ma,
  title={U-mixer: An unet-mixer architecture with stationarity correction for time series forecasting},
  author={Ma, Xiang and Li, Xuemei and Fang, Lexin and Zhao, Tianlong and Zhang, Caiming},
  booktitle={Proceedings of the 38th AAAI conference on Artificial Intelligence},
  //booktitle={AAAI},
  //volume={38},
  number={13},
  pages={14255--14262},
  year={2024}
}

@inproceedings{conference/neurips2024/kim,
  title={Are self-attentions effective for time series forecasting?},
  author={Kim, Dongbin and Park, Jinseong and Lee, Jaewook and Kim, Hoki},
  booktitle={Proceedings of the 38th Annual Conference on Neural Information Processing Systems},
  //booktitle={NeurIPS},
  //volume={37},
  pages={114180--114209},
  year={2024},
  url = {http://papers.nips.cc/paper\_files/paper/2024/hash/cf66f995883298c4db2f0dcba28fb211-Abstract-Conference.html},
}

@inproceedings{conference/ijcai/WenZZCMY023,
  author       = {Qingsong Wen and
                  Tian Zhou and
                  Chaoli Zhang and
                  Weiqi Chen and
                  Ziqing Ma and
                  Junchi Yan and
                  Liang Sun},
  title        = {Transformers in Time Series: {A} Survey},
  booktitle    = {Proceedings of the 32nd International Joint Conference on Artificial Intelligence},
  //booktitle={IJCAI},
  pages        = {6778--6786},
  publisher    = {ijcai.org},
  year         = {2023},
  //url          = {https://doi.org/10.24963/ijcai.2023/759},
  //doi          = {10.24963/IJCAI.2023/759},
}

@ARTICLE{journal/tpami2024/jin2,
  author={Jin, Ming and Koh, Huan Yee and Wen, Qingsong and Zambon, Daniele and Alippi, Cesare and Webb, Geoffrey I. and King, Irwin and Pan, Shirui},
  journal={IEEE Transactions on Pattern Analysis and Machine Intelligence}, 
  title={A Survey on Graph Neural Networks for Time Series: Forecasting, Classification, Imputation, and Anomaly Detection}, 
  year={2024},
  volume={46},
  number={12},
  pages={10466-10485},
  doi={10.1109/TPAMI.2024.3443141}
}

@inproceedings{conference/iclr2025/Shi,
      title={Time-MoE: Billion-Scale Time Series Foundation Models with Mixture of Experts}, 
      author={Xiaoming Shi and Shiyu Wang and Yuqi Nie and Dianqi Li and Zhou Ye and Qingsong Wen and Ming Jin},
      booktitle={Proceedings of the 13th International Conference on Learning Representations},
      //booktitle={ICLR},
      year={2025},
      address = {Singapore},
}

@inproceedings{conference/iclr2025/Wang,
  title={Timemixer++: A general time series pattern machine for universal predictive analysis},
  author={Wang, Shiyu and Li, Jiawei and Shi, Xiaoming and Ye, Zhou and Mo, Baichuan and Lin, Wenze and Ju, Shengtong and Chu, Zhixuan and Jin, Ming},
  booktitle={Proceedings of the 13th International Conference on Learning Representations},
  //booktitle={ICLR},
  year={2025},
  address = {Singapore},
}

@inproceedings{conference/iclr2024/wangtmixer,
  title={TimeMixer: Decomposable Multiscale Mixing for Time Series Forecasting},
  author={Wang, Shiyu and Wu, Haixu and Shi, Xiaoming and Hu, Tengge and Luo, Huakun and Ma, Lintao and Zhang, James Y and ZHOU, JUN},
  booktitle={Proceedings of the 12th International Conference on Learning Representations},
  //booktitle={ICLR},
  year={2024},
  address = {Vienna, Austria},
}

@inproceedings{conference/iclr2023/niepatchtst,
  title={A Time Series is Worth 64 Words:  Long-term Forecasting with Transformers},
  author={Yuqi Nie and Nam H Nguyen and Phanwadee Sinthong and Jayant Kalagnanam},
  booktitle={Proceedings of the 11th International Conference on Learning Representations},
  //booktitle={ICLR},
  year={2023},
  address = {Kigali, Rwanda},
}

@inproceedings{conference/nips2024/jia,
  author       = {Yuxin Jia and
                  Youfang Lin and
                  Jing Yu and
                  Shuo Wang and
                  Tianhao Liu and
                  Huaiyu Wan},
  title        = {{PGN:} The RNN's New Successor is Effective for Long-Range Time Series Forecasting},
  booktitle    = {Proceedings of the 38th Annual Conference on Neural Information Processing Systems},
  //booktitle    = {NeurIPS},
  year         = {2024},
  month        = {Dec},
  address    = {Vancouver, BC, Canada},
  //url          = {http://papers.nips.cc/paper\_files/paper/2024/hash/990641d09f71bcee0060a8f1704ab8e2-Abstract-Conference.html},
}

@inproceedings{conference/aaai2025/kong,
  author       = {Yaxuan Kong and
                  Zepu Wang and
                  Yuqi Nie and
                  Tian Zhou and
                  Stefan Zohren and
                  Yuxuan Liang and
                  Peng Sun and
                  Qingsong Wen},
  title        = {Unlocking the Power of {LSTM} for Long Term Time Series Forecasting},
  booktitle={Proceedings of the 39th AAAI Conference on Artificial Intelligence},
  //booktitle= {AAAI},
  pages        = {11968--11976},
  //publisher    = {{AAAI} Press},
  year         = {2025},
  month        = {Feb},
  address      = {Philadelphia, PA, USA},
  //url          = {https://doi.org/10.1609/aaai.v39i11.33303},
  //doi          = {10.1609/AAAI.V39I11.33303},
}

@inproceedings{conference/ecmlpkdd2023/chen,
  author       = {Yuzhou Chen and
                  Tian Jiang and
                  Yulia R. Gel},
  title        = {H\({}^{\mbox{2}}\)-Nets: Hyper-hodge Convolutional Neural Networks for Time-Series Forecasting},
  booktitle    = {Proceedings of the Joint European Conference on Machine Learning and Knowledge Discovery in Databases},
  //booktitle    = {ECML-PKDD},
  volume       = {14173},
  pages        = {271--289},
  publisher    = {Springer},
  year         = {2023},
  month        = {Sep},
  address      = {Turin, Italy},
  //url          = {https://doi.org/10.1007/978-3-031-43424-2\_17},
  //doi          = {10.1007/978-3-031-43424-2\_17},
}

@article{lavers2014extending,
  title={Extending medium-range predictability of extreme hydrological events in Europe},
  author={Lavers, David A and Pappenberger, Florian and Zsoter, Ervin},
  journal={Nature Communications},
  volume={5},
  number={1},
  pages={5382},
  year={2014},
  //publisher={Nature Publishing Group UK London}
}

@article{camps2025artificial,
  title={Artificial intelligence for modeling and understanding extreme weather and climate events},
  author={Camps-Valls, Gustau and Fern{\'a}ndez-Torres, Miguel-{\'A}ngel and Cohrs, Kai-Hendrik and H{\"o}hl, Adrian and Castelletti, Andrea and Pacal, Aytac and Robin, Claire and Martinuzzi, Francesco and Papoutsis, Ioannis and Prapas, Ioannis and others},
  journal={Nature Communications},
  volume={16},
  number={1},
  pages={1919},
  year={2025},
  //publisher={Nature Publishing Group UK London}
}

@inproceedings{conference/iclr2015/Kingma,
  author       = {Diederik P. Kingma and
                  Jimmy Ba},
  title        = {Adam: {A} Method for Stochastic Optimization},
  booktitle    = {Proceedings of the 3rd International Conference on Learning Representations},
  address = {San Diego, CA, USA},
  //booktitle    = {ICLR},
  year         = {2015},
}

@article{maaten2008visualizing,
  title={Visualizing data using t-SNE},
  author={Maaten, Laurens van der and Hinton, Geoffrey},
  journal={Journal of Machine Learning Research},
  volume={9},
  //number={Nov},
  pages={2579--2605},
  year={2008}
}

@article{mallat2002theory,
  title={A theory for multiresolution signal decomposition: the wavelet representation},
  author={Mallat, Stephane G},
  journal={IEEE Transactions on Pattern Analysis and Machine Intelligence},
  volume={11},
  number={7},
  pages={674--693},
  year={2002},
  publisher={Ieee}
}

@book{mallat1999wavelet,
  title={A wavelet tour of signal processing},
  author={Mallat, St{\'e}phane},
  year={1999},
  publisher={Elsevier}
}

@book{daubechies1992ten,
  title={Ten lectures on wavelets},
  author={Daubechies, Ingrid},
  year={1992},
  publisher={SIAM}
}

@book{oppenheim1999discrete,
  title={Discrete-time signal processing},
  author={Oppenheim, Alan V},
  year={1999},
  publisher={Pearson Education India}
}

@article{lee2019pywavelets,
  title={PyWavelets: A Python package for wavelet analysis},
  author={Lee, Gregory and Gommers, Ralf and Waselewski, Filip and Wohlfahrt, Kai and O'Leary, Aaron},
  journal={Journal of Open Source Software},
  volume={4},
  number={36},
  pages={1237},
  year={2019},
  publisher={The Open Journal}
}

@misc{SCVWDalert2025,
  title        = {ALERT System Real­Time Data – Santa Clara Valley Water District},
  author       = {{Santa Clara Valley Water District}},
  //howpublished = {\url{https://valleywateralert.org/scvwd/}},
  year         = {2025},
  note         = {Accessed: 2025‑08‑05}
}





\appendix

\section{Appendix}

\subsection{Appendix A. Proof of Theorem 1}
\label{app:proof_fixed_ratio}

This proof is organized in three parts. Part~1 establishes the mapping between frequency and scale boundaries. Part~2 demonstrates energy conservation under this mapping. Part~3 bounds the discretization error in the wavelet transform.

\subsubsection{Part 1: Boundary Correspondence}
Let $f$ denote the normalized frequency. Let $a$ be the scale parameter of the continuous wavelet transform. Define $\gamma = f_0 / f_{\mathrm{nyq}}$. The mapping $a = \gamma / f$ is a bijection between frequency and scale domains \cite{mallat2002theory}. At boundary frequencies $f_{\min}$ and $f_{\max}$, the corresponding scales satisfy:
\begin{equation}
a_{\min} = \gamma / f_{\max}, \quad a_{\max} = \gamma / f_{\min}.
\label{eq:amin_amax}
\end{equation}

This mapping preserves the ratio of scales to frequencies, i.e., the ratio of maximum to minimum scales is equal to the ratio of maximum to minimum frequencies:
\begin{equation}
\frac{a_{\max}}{a_{\min}} = \frac{f_{\max}}{f_{\min}}.
\label{eq:ratio_preserve}
\end{equation}

Then, taking logarithms of both sides gives:
\begin{equation}
\log\left(\frac{a_{\max}}{a_{\min}}\right) = \log\left(\frac{f_{\max}}{f_{\min}}\right).
\label{eq:log_equiv}
\end{equation}

\subsubsection{Part 2: Energy Conservation}
Let $s(t)$ be a continuous-time signal sampled at interval $\Delta t$. Denote by $\mathcal{F}[n]$ its discrete Fourier transform coefficients. The energy of the Fourier coefficients in the frequency band $[f_{\min}, f_{\max}]$ is:
\begin{equation}
\mathcal{E}_{\mathrm{DFT}} = \sum_{f_n \in [f_{\min}, f_{\max}]} \left|\mathcal{F}[n]\right|^2 .
\label{eq:edft}
\end{equation}

Parseval’s theorem \cite{oppenheim1999discrete} for the continuous wavelet transform states that:
\begin{equation}
\int_{-\infty}^{\infty} |s(t)|^2 \, dt = \frac{1}{C_\psi} \int_0^\infty \int_{-\infty}^{\infty} \bigl|\mathcal{W}(a,b)\bigr|^2 \, db \, \frac{da}{a^2} ,
\label{eq:parseval}
\end{equation}
where $C_\psi$ is the admissibility constant of the mother wavelet \cite{mallat1999wavelet}.
Apply the change of variables $f = \gamma / a$. The Jacobian for this transformation satisfies:
\begin{equation}
\frac{da}{a^2} = \frac{1}{\gamma} \, df .
\label{eq:jacobian}
\end{equation}

The energy of the continuous wavelet transform within scales $[a_{\min}, a_{\max}]$ can be expressed as:
\begin{equation}
\mathcal{E}_{\mathrm{CWT}} = \frac{1}{C_\psi} \int_{a_{\min}}^{a_{\max}} \int_{-\infty}^{\infty} \bigl|\mathcal{W}(a,b)\bigr|^2 \, db \, \frac{da}{a^2} .
\label{eq:ecwt_def}
\end{equation}

Use the substitution $f = \gamma / a$ to rewrite \eqref{eq:ecwt_def} as:
\begin{equation}
\mathcal{E}_{\mathrm{CWT}} = \frac{1}{C_\psi \gamma} \int_{f_{\min}}^{f_{\max}} \int_{-\infty}^{\infty} \left|\mathcal{W}\!\left(\frac{\gamma}{f}, b\right)\right|^2 db \, df .
\label{eq:ecwt_sub}
\end{equation}

For a discrete signal of length $T_{in}$, the wavelet coefficient $\mathcal{W}(a,b)$ concentrates energy near frequency $f = \gamma / a$ \cite{daubechies1992ten}. The finite sampling interval induces a discretization error term $\mathcal{E}$ that is bounded by a term of order $1/T_{in}$. Then, the energy of the wavelet transform can be approximated as:
\begin{equation}
\int_{-\infty}^{\infty} \left|\mathcal{W}\!\left(\frac{\gamma}{f}, b\right)\right|^2 db = \frac{C_\psi}{\gamma} \sum_{f_n \approx f} \left|\mathcal{F}[n]\right|^2 + \mathcal{E} .
\label{eq:local_energy}
\end{equation}

Substituting \eqref{eq:local_energy} into \eqref{eq:ecwt_sub} yields the energy of the continuous wavelet transform:
\begin{equation}
\mathcal{E}_{\mathrm{CWT}} = \frac{1}{\gamma^2} \int_{f_{\min}}^{f_{\max}} \sum_{f_n \approx f} \left|\mathcal{F}[n]\right|^2 df + O\!\left(\frac{1}{T_{in}}\right),
\label{eq:ecwt_integral}
\end{equation}
where $O(\cdot)$ denotes the order of the discretization error.
Upon integrating over the band $[f_{\min}, f_{\max}]$, we obtain:
\begin{equation}
\mathcal{E}_{\mathrm{CWT}} = \frac{\gamma}{C_\psi} \Delta f \cdot \mathcal{E}_{\mathrm{DFT}} + \mathcal{E},
\label{eq:energy_proportional}
\end{equation}
where $\Delta f = f_{\max} - f_{\min}$. $\mathcal{F}[n]$ denotes the discrete Fourier transform coefficient at the $n$th frequency bin, $\mathcal{W}(a,b)$ is the continuous wavelet transform coefficient at scale $a$ and time shift $b$, $C_\psi$ is the wavelet admissibility constant, $\gamma = f_0 / f_{\mathrm{nyq}}$ is the constant linking frequency to scale, and $\mathcal{E}$ is the discretization error arising from finite signal length.

\subsubsection{Part 3: Discretization Error Bound}
If the mother wavelet decays exponentially, the error term $\mathcal{E}$ satisfies $|\mathcal{E}| \leq \frac{Q}{T_{in}}$, for some constant $Q$ that depends on the time-frequency localization of the wavelet. For signals with length $T_{in}$ much greater than the support of the wavelet, this error becomes negligible.

\subsection{Appendix B. Data Description}

\begin{figure*}[t!]
    \centering
    \includegraphics[width=0.33\linewidth]{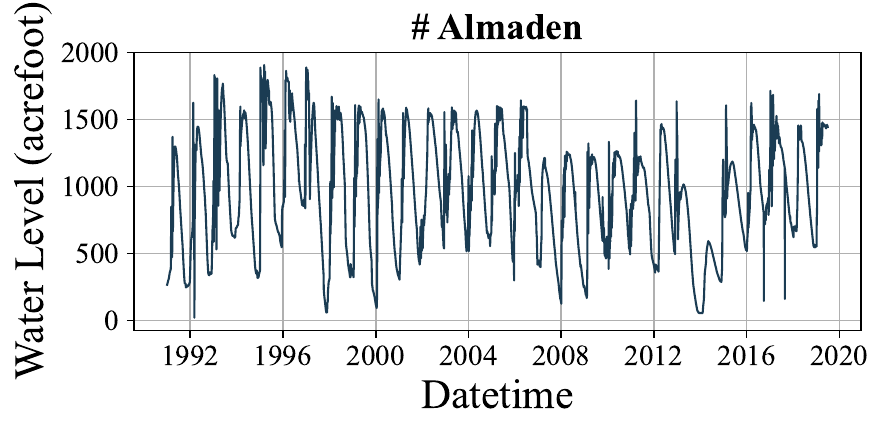}
    \includegraphics[width=0.33\linewidth]{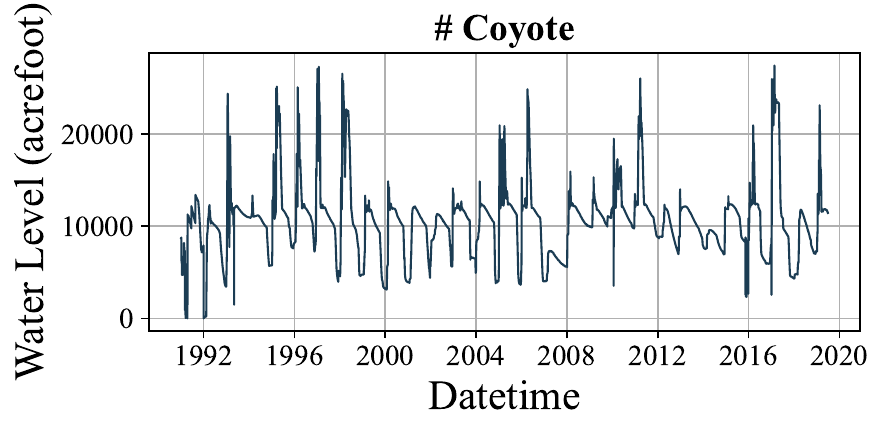}
    \includegraphics[width=0.33\linewidth]{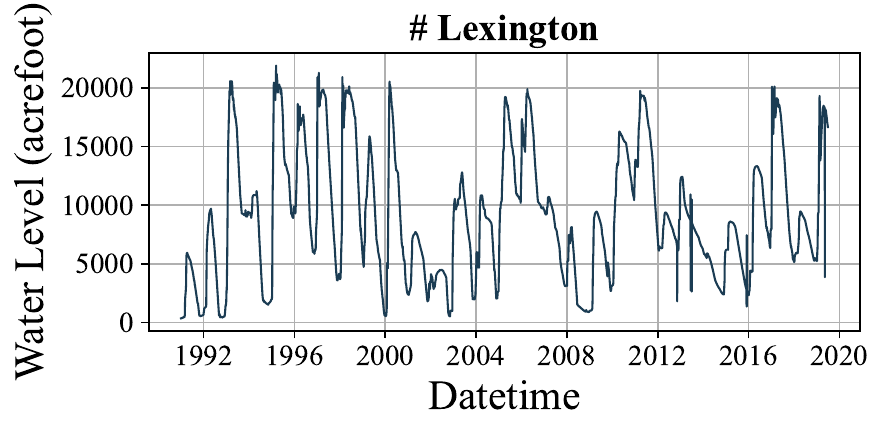}
    \includegraphics[width=0.33\linewidth]{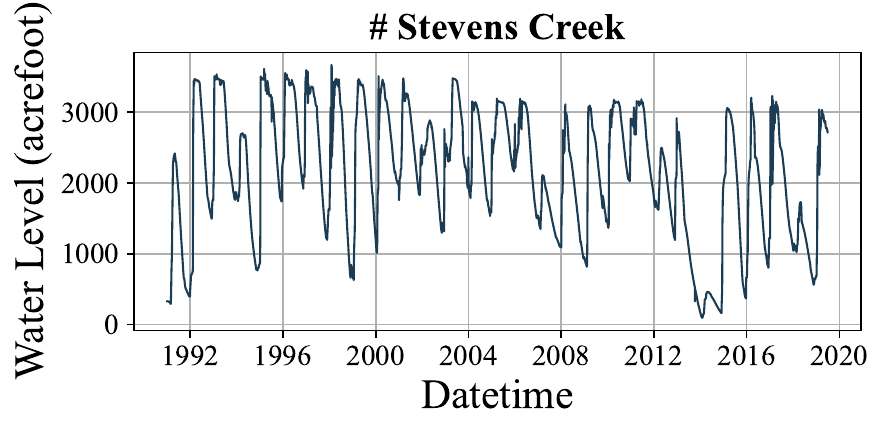}
    \includegraphics[width=0.33\linewidth]{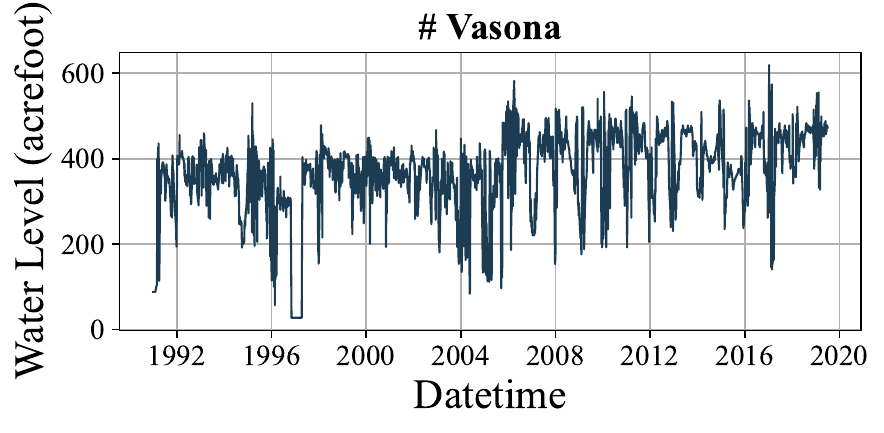}

    \caption{Water levels of five reservoirs in our study.}
    \label{fig:reservoirs}
\end{figure*}

This study employs a comprehensive dataset comprising hourly water level measurements from five reservoirs in Santa Clara County, California: Almaden, Coyote, Lexington, Stevens Creek, and Vasona. The dataset spans the period from July 1, 1991 to June 30, 2019, totaling 245,426 data points. The original data, sourced from the Santa Clara Valley Water District \cite{SCVWDalert2025}, is publicly available and has been preprocessed in prior work \cite{journal/tpami2025/li6888}, which we directly adopt.
The data is divided into training, validation, and test sets. The training and validation sets cover the period from July 1, 1991 to June 30, 2018, while the test set spans from July 1, 2018 to June 30, 2019. To ensure comparability with \cite{journal/tpami2025/li6888}, we maintain identical splits across all experiments. The training set contains over 30,000 samples for Almaden and more than 40,000 samples for each of the other reservoirs. The validation set comprises 60 randomly selected samples excluded from training.
\textbf{Figure~\ref{fig:reservoirs}} illustrates the water level trends of the five reservoirs. Descriptive statistics, including skewness and kurtosis, are summarized in \textbf{Table~\ref{tab:reservoir_stats}}. Positive skewness values suggest right-tailed distributions, while high kurtosis values indicate the presence of heavy tails.

\begin{table}[t]
  \centering
  \small
  \setlength{\tabcolsep}{1mm}
    \begin{tabular}{cccccc}
    \toprule
    \textbf{Data} & \textbf{Min} & \textbf{Max} & \textbf{Variance} & \textbf{Skew.} & \textbf{Kurt.} \\
    \midrule
    Almaden & 20.24  & 1906.11  & 177282.68  & -0.13  & -0.88  \\
    Coyote & 4.26  & 27421.26  & 15161787.15  & 1.01  & 2.66  \\
    Lexington & 406.17  & 21895.50  & 29690104.22  & 0.39  & -0.80  \\
    Stevens Creek & 93.92  & 3667.37  & 786364.99  & -0.45  & -0.74  \\
    Vasona & 27.44  & 619.49  & 7690.87  & -1.29  & 2.75  \\
    \bottomrule
    \end{tabular}%
  \caption{Statistical properties of water level data for each reservoir. Processed data is publicly available from \cite{journal/tpami2025/li6888}. The table includes the minimum, maximum, variance, skewness (Skew.), and kurtosis (Kurt.) of the water levels.}
  \label{tab:reservoir_stats}%
\end{table}%

We adopt the same cluster-based oversampling strategy proposed in \cite{journal/tpami2025/li6888}. Specifically, a Gaussian Mixture Model (GMM) is fitted to the training set to partition the data into $M = 3$ clusters. From the cluster means, upper and lower thresholds are derived to identify right-extreme and left-extreme events. For each identified extreme point, additional training samples are generated by sliding a window (step size = 4, number of windows = 18). The amount of oversampled data is constrained to 20\% of the original training set for Stevens Creek and 40\% for the other reservoirs.
The GMM assumes that the data distribution is modeled as a weighted sum of Gaussian components. The probability density function of a data point $X$ is given by:
\begin{align}
p(X|\Lambda) = \sum_{k=1}^{M} w_k \, g(X|\mu_k, \Sigma_k) \quad \text{s.t.} \sum_{k=1}^{M} w_k = 1,
\label{1}
\end{align}
where $M$ is the number of Gaussian components (clusters), $w_k$ is the mixture weight of the $k$-th component, and $\Lambda = \{w_k, \mu_k, \Sigma_k\}_{k=1}^M$ represents the set of all model parameters. Each component density $g(X|\mu_k, \Sigma_k)$ is a Gaussian distribution with mean $\mu_k$ and covariance $\Sigma_k$:
\begin{align}
g(X|\mu_k, \Sigma_k) 
&= \frac{1}{(2\pi)^{D/2}|\Sigma_k|^{1/2}} \notag\\
&\quad \times \exp\left(-\frac{1}{2}(X-\mu_k)^T\Sigma_k^{-1}(X-\mu_k)\right),
\label{2}
\end{align}
where $D$ is the dimensionality of the data point $X$. The parameters $\Lambda$ are estimated from the training data using the Expectation-Maximization (EM) algorithm.

The resulting clusters and their associated Gaussian distributions are further utilized to define data-driven thresholds for the oversampling process. Specifically, the lower threshold is computed based on the mean values of the two clusters with the lowest averages, while the upper threshold is derived from the means of the two clusters with the highest averages. These thresholds are then employed to identify left- and right-extreme events, as described in the oversampling strategy in the main text.
Notably, the same GMM-based clustering procedure is also leveraged to construct input features for the MCANN and DAN models in the form of \textbf{extreme event labels}, providing auxiliary supervision during training.

\subsection{Appendix C. Implementation Details}

We follow the standard training and evaluation protocols outlined in \cite{journal/tpami2025/li6888}. All models are trained using the Adam optimizer \cite{conference/iclr2015/Kingma} with a learning rate of either 0.001 or 0.0005, and a batch size of 48. Training proceeds for up to 100 epochs with early stopping based on validation loss. All implementations are based on PyTorch (v1.10.1) with Python (v3.9.20), and experiments are conducted on a single NVIDIA RTX 3090 GPU (24GB).
Input data is standardized using z-score normalization for all models, except for MCANN \cite{journal/tpami2025/li6888} and DAN \cite{conference/aaai2024/li27768}, which adopt model-specific normalization schemes as described in their original works. The input sequence length is fixed at 360 hours (15 days), and the forecasting horizon is set to 72 hours (3 days). Following \cite{journal/tpami2025/li6888}, we adopt a rolling evaluation strategy to assess both 72-hour-ahead and 8-hour-ahead forecasting performance.
The loss function is Mean Squared Error (MSE) for all models, except DAN, which employs a customized loss incorporating extreme event labels. All reported results are averaged over five independent runs with different random seeds to ensure robustness. Following the official evaluation protocol, we report two metrics: Root Mean Squared Error (RMSE) and Mean Absolute Percentage Error (MAPE), defined as follows:
\begin{align}
\mathrm{RMSE} &= \sqrt{\frac{1}{N} \sum_{i=1}^{N} (X_i - \hat{X}_i)^2}, \\
\mathrm{MAPE} &= \frac{1}{N} \sum_{i=1}^{N} \left| \frac{X_i - \hat{X}_i}{X_i + \epsilon} \right|,
\end{align}
where $X_i$ and $\hat{X}_i$ denote the ground truth and predicted values, respectively, and $N$ is the number of evaluation samples. To avoid division by zero in MAPE calculation, a small constant $\epsilon$ is added to the denominator. Following the protocol in \cite{journal/tpami2025/li6888}, we set $\epsilon = 1$ in all experiments.

The model hyper-parameters are selected via grid search based on validation performance. The search ranges for each model are summarized in \textbf{Table~\ref{table:parameter_settings}}. For all models, the dropout rate is selected from the set \{0, 0.1, 0.2\}. For DAN and MCANN, we adopt the optimal hyper-parameter configurations reported in their respective original papers.
For the continuous wavelet transform in M$^2$FMoE, we use “\texttt{cgau7}” as the mother wavelet, a commonly adopted choice in time series analysis \cite{mallat1999wavelet}. The wavelet transform is implemented using the \texttt{PyWavelets} library \cite{lee2019pywavelets} with a scale resolution of 16.
The detailed hyper-parameter settings specific to M$^2$FMoE are provided in \textbf{Table~\ref{tab:hyper_our}}.

\begin{table}[t!]
    \small
	\centering
		\begin{tabular}{c c c}
			\toprule
			\textbf{Model} & \textbf{Parameter} & \textbf{Option range} \\ 
			
			\midrule
			
			\multirow{5}{*}{{CATS}} &  Patch size & \{6, 12, 24\} \\
            & Stride size &  \{6, 8, 12, 24\} \\
			 &  Hidden dimension & \{$2^5, 2^6, 2^7, 2^8$ \} \\
             &  The number of heads & \{$2^3, 2^4, 2^5, 2^6, 2^7$ \} \\
             &  The number of layers & 1-3 (1 per step) \\

			\midrule
            \multirow{2}{*}{{CycleNet}} &  Hidden dimension & \{$2^5, 2^6, 2^7, 2^8$ \} \\
            & cycle length & \{12, 24, 48\} \\
            \midrule
            \multirow{2}{*}{{FreqMoE}} &  The number of blocks & 1-3 (1 per step) \\
            &  The number of experts & 1-9 (2 per step) \\
            \midrule
            \multirow{3}{*}{{iTransformer}} &  The number of layers & 1-3 (1 per step) \\
            &  Hidden dimension & \{$2^5, 2^6, 2^7, 2^8$ \} \\
            &  The number of heads & \{$2^3, 2^4, 2^5, 2^6, 2^7$ \} \\
            \midrule
            \multirow{3}{*}{{KAN}} & Grid size & \{5, 10, 15\} \\
            & Spline degree & \{1, 2, 3\} \\
            & Scale Noise & \{0.1, 0.2, 0.3\} \\
            \midrule
            \multirow{2}{*}{TQNet} &  Cycle length & \{12, 24, 48\} \\
            &  Hidden dimension & \{$2^5, 2^6, 2^7, 2^8$ \} \\
            \midrule
            \multirow{4}{*}{{Umixer}} &  The number of layers & 1-3 (1 per step) \\
            &  Hidden dimension & \{$2^5, 2^6, 2^7, 2^8$ \} \\
            &  Patch size & \{6, 12, 24\} \\
            &  Stride size & \{6, 8, 12, 24\} \\
            \midrule
            \multirow{5}{*}{M$^2$FMoE} &  The number of experts & 1-9 (2 per step) \\
            &  Hidden dimension & \{$2^5, 2^6, 2^7, 2^8$ \} \\
            & Resolution size & \{6, 12, 24\} \\
            & Diversity loss weight & \{0.05, 0.1, 0.5\} \\
            & Consistency loss weight & \{0.05, 0.1, 0.5\} \\
			\bottomrule	
		\end{tabular}
        \caption{Hyper-parameter settings.} 
        \label{table:parameter_settings}
\end{table}

\begin{table}[t!]
  \centering
    \small
    \setlength{\tabcolsep}{1mm}
    \resizebox{1\linewidth}{!}{
    \begin{tabular}{cccccc}
    \toprule
    \textbf{Parameter} & \textbf{Almaden} & \textbf{Coyote} & \textbf{Lexington} & \textbf{\shortstack{Stevens \\ Creek}} & \textbf{Vasona} \\
    \midrule
    Number of experts  & 3     & 3     & 3     & 3     & 3 \\
    Hidden dimension & 64    & 64    & 64    & 256   & 64 \\
    Resolution size & [12, 24] & [12, 24] & [12, 24] & [12, 24] & [12, 24] \\
    Diversity loss weight & 0.05  & 0.1   & 0.1   & 0.05  & 0.05 \\
    Consistency loss weight & 0.05  & 0.1   & 0.5  & 0.05  & 0.05 \\
    \bottomrule
    \end{tabular}%
    }
    \caption{Hyper-parameter settings of M$^2$FMoE for each dataset.}
  \label{tab:hyper_our}%
\end{table}%

\begin{table*}[t!]
  \centering
  \small
\resizebox{1.0\linewidth}{!}{	
    \begin{tabular}{lcccccccccc}
    \toprule
    \multirow{2}[4]{*}{\textbf{Model}} & \multicolumn{2}{c}{\textbf{Almaden}} & \multicolumn{2}{c}{\textbf{Coyote}} & \multicolumn{2}{c}{\textbf{Lexington}} & \multicolumn{2}{c}{\textbf{Stevens Creek}} & \multicolumn{2}{c}{\textbf{Vasona}} \\
\cmidrule{2-11}          & RMSE  & MAPE  & RMSE  & MAPE  & RMSE  & MAPE  & RMSE  & MAPE  & RMSE  & MAPE \\
\midrule
    \textbf{M$^2$FMoE} & \textbf{7.990 } & \textbf{0.002 } & \textbf{48.797 } & \textbf{0.002 } & \textbf{251.957 } & 0.004  & \textbf{10.559 } & \textbf{0.002 } & \textbf{5.129 } & \textbf{0.004 } \\
\midrule
    \textit{w/o-WaveletView} & 10.362  & 0.003  & 89.775  & 0.003  & 289.561  & 0.004  & 16.101  & 0.003  & 5.178  & \textbf{0.004 } \\
    \textit{w/o-FourierView} & 12.274  & 0.003  & 90.717  & 0.003  & 294.209  & 0.004  & 16.520  & \textbf{0.002 } & 6.753  & 0.007  \\
    \textit{w/o-}$\mathcal{L}_\text{div}$ \& $\mathcal{L}_\text{cons}$  & 8.385  & 0.003  & 50.499  & \textbf{0.002 } & 273.837  & 0.004  & 10.843  & \textbf{0.002 } & 5.449  & 0.005  \\
    \textit{w/o-Multi-Res} & 12.255  & 0.003  & 80.057  & \textbf{0.002 } & 293.086  & 0.004  & 16.496  & 0.003  & 6.626  & 0.006  \\
    \textit{w/o-CSS} & 9.154  & 0.003  & 71.246  & \textbf{0.002 } & 291.258  & \textbf{0.003 } & 15.802  & 0.003  & 5.514  & 0.005  \\
    \textit{w/o-Alignment} & 11.800  & 0.004  & 63.845  & \textbf{0.002 } & 296.010  & 0.004  & 15.870  & 0.003  & 6.453  & 0.007  \\
    \textit{w/o-DualView} & 8.962  & 0.004  & 63.440  & 0.003  & 268.500  & 0.004  & 10.669  & \textbf{0.002 } & 5.313  & 0.005  \\

    \bottomrule
    \end{tabular}
}
    \caption{Ablation study results on the five reservoirs with a prediction horizon of 8 hours. Best results are highlighted in \textbf{bold}. } 
    \label{tab:ablation}
\end{table*}

\begin{figure*}[t!]
    \centering
    \includegraphics[width=0.245\linewidth]{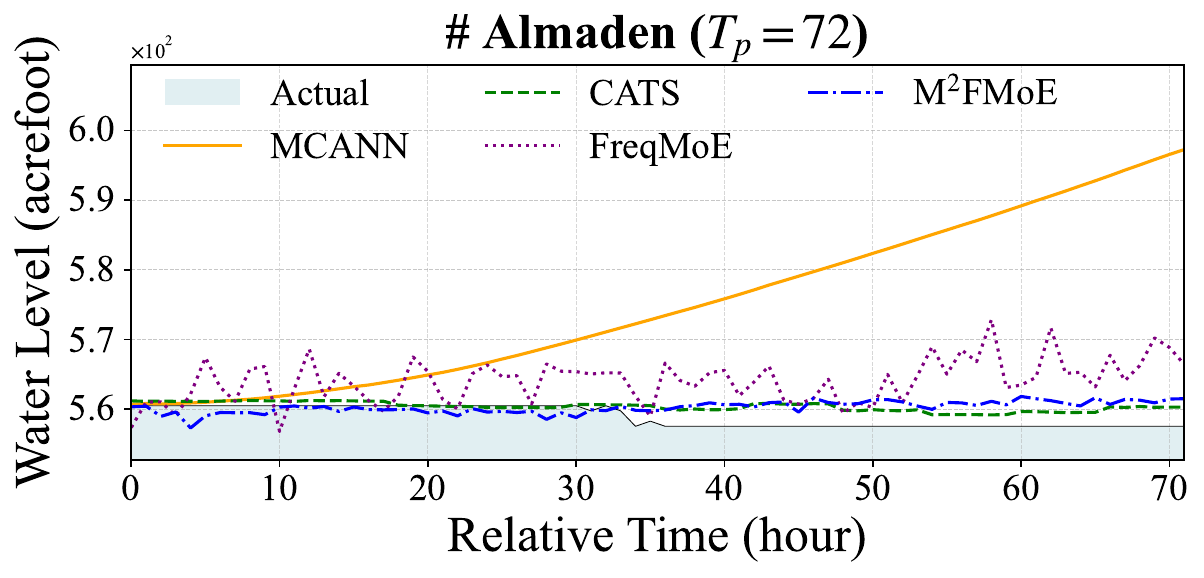}
    \includegraphics[width=0.245\linewidth]{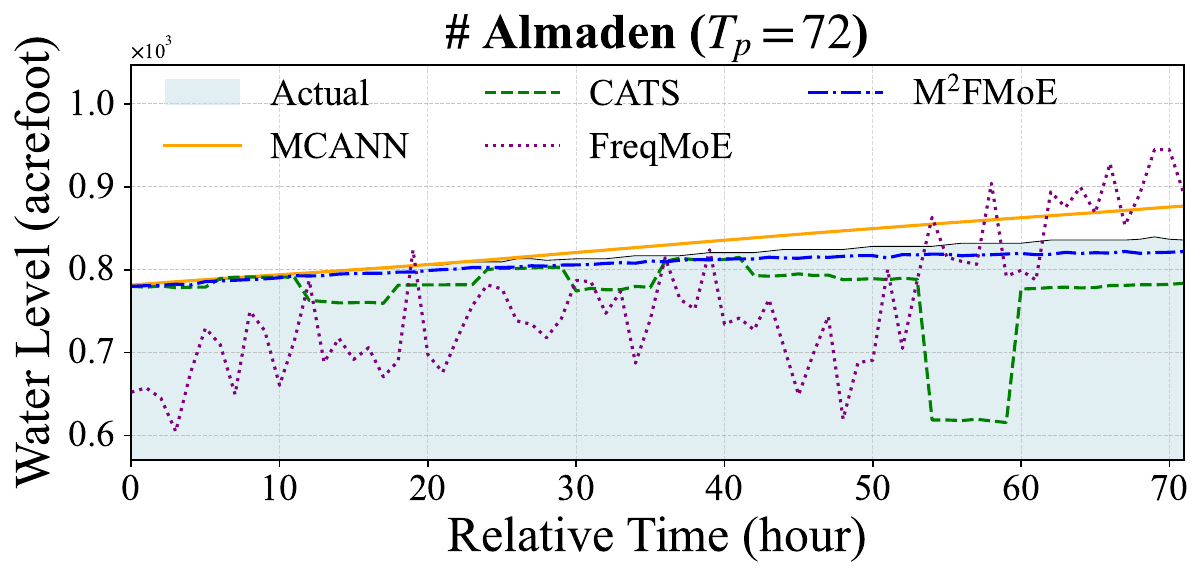}
    \includegraphics[width=0.245\linewidth]{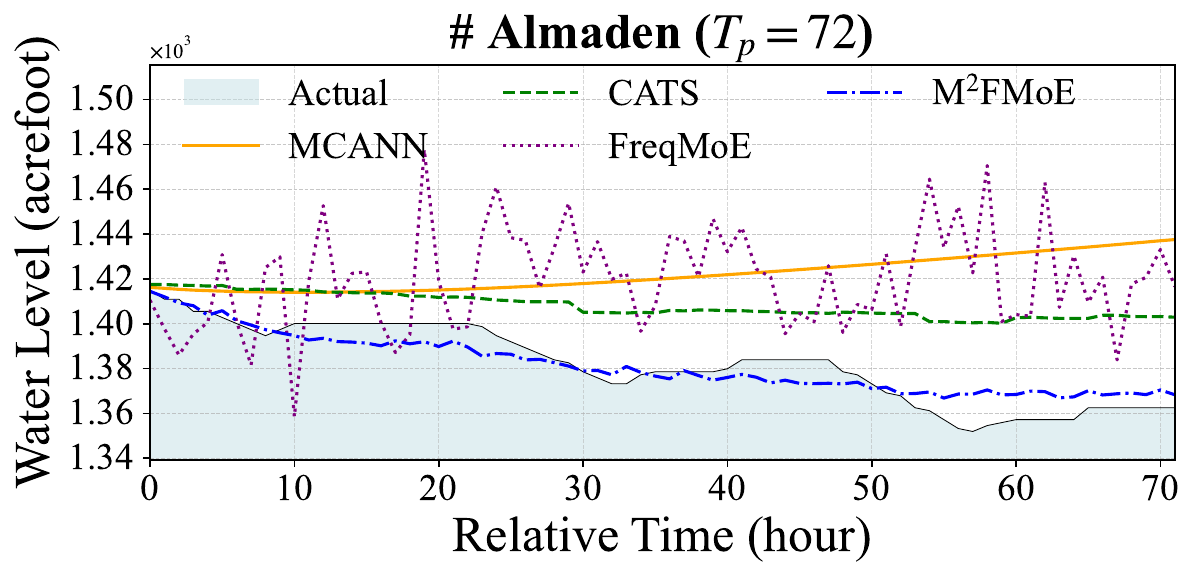}
    \includegraphics[width=0.245\linewidth]{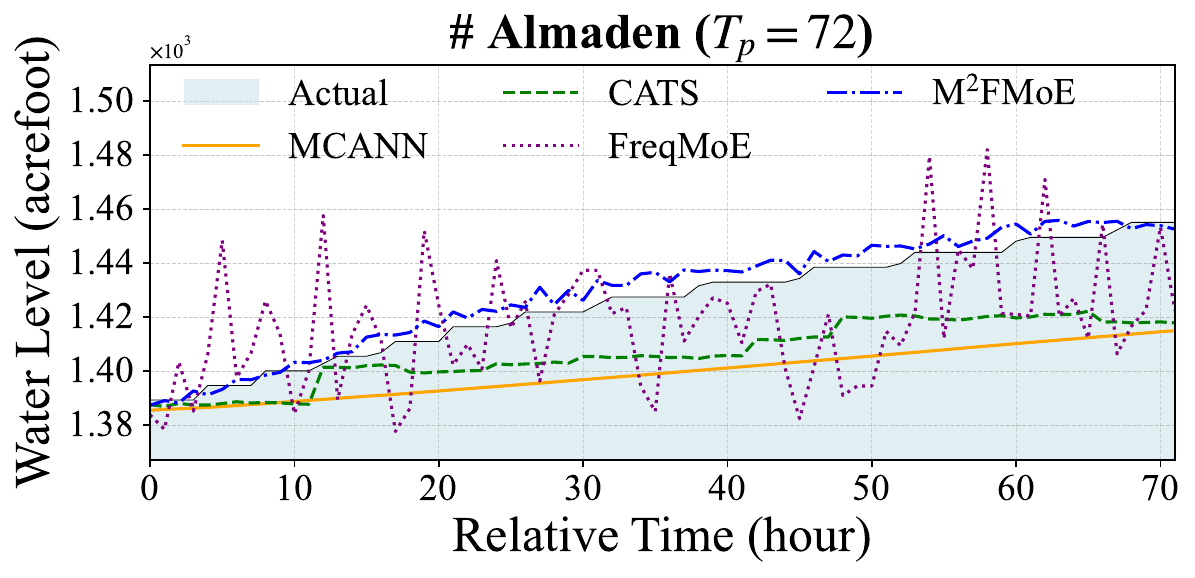}

    \includegraphics[width=0.245\linewidth]{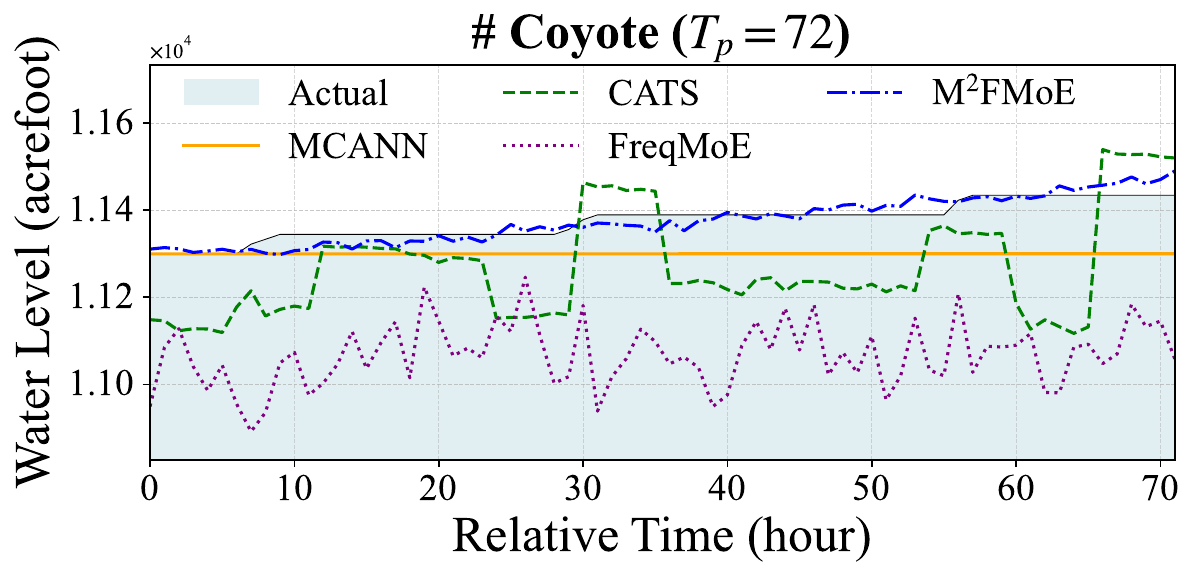}
    \includegraphics[width=0.245\linewidth]{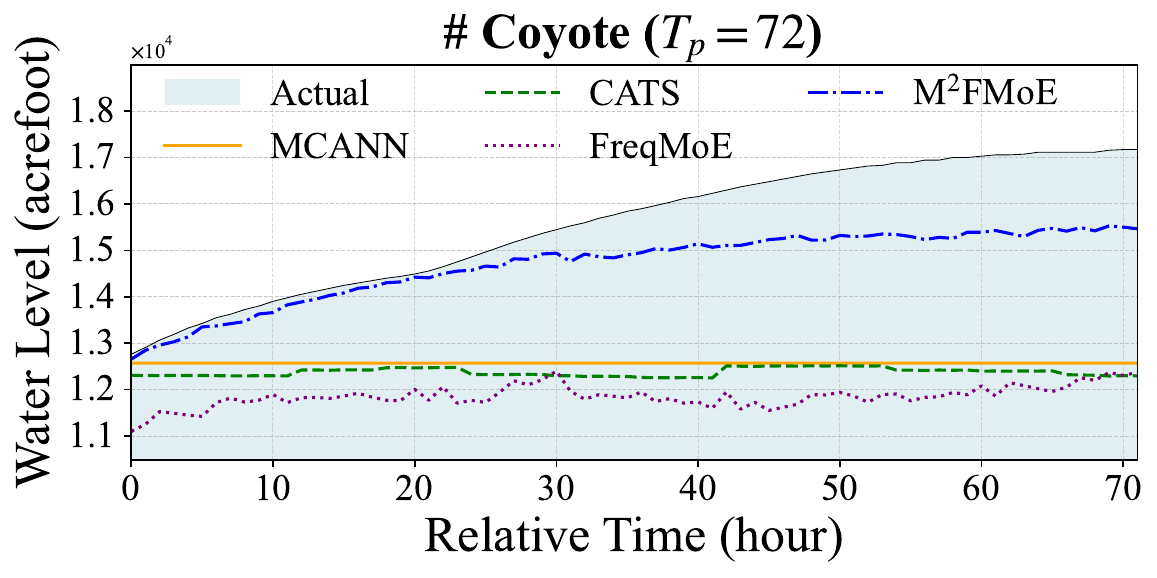}
    \includegraphics[width=0.245\linewidth]{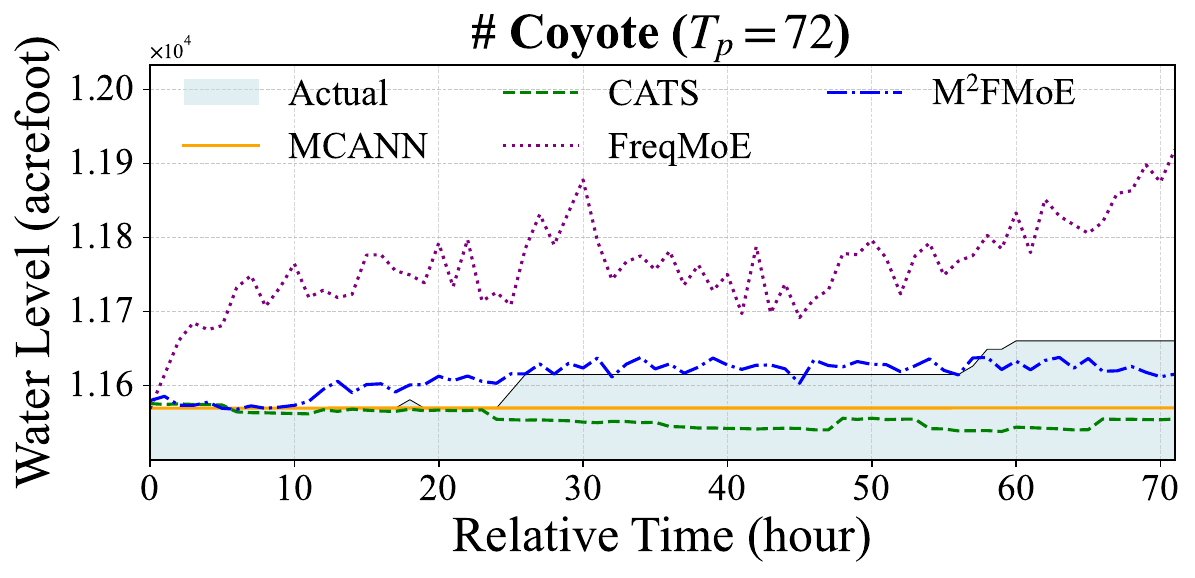}
    \includegraphics[width=0.245\linewidth]{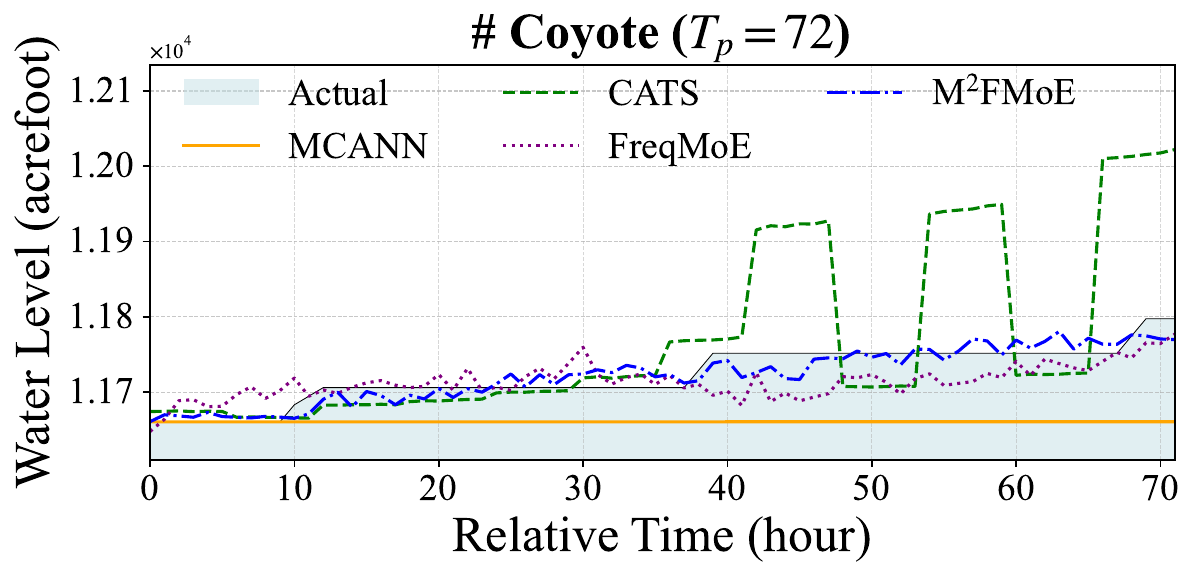}

    \includegraphics[width=0.245\linewidth]{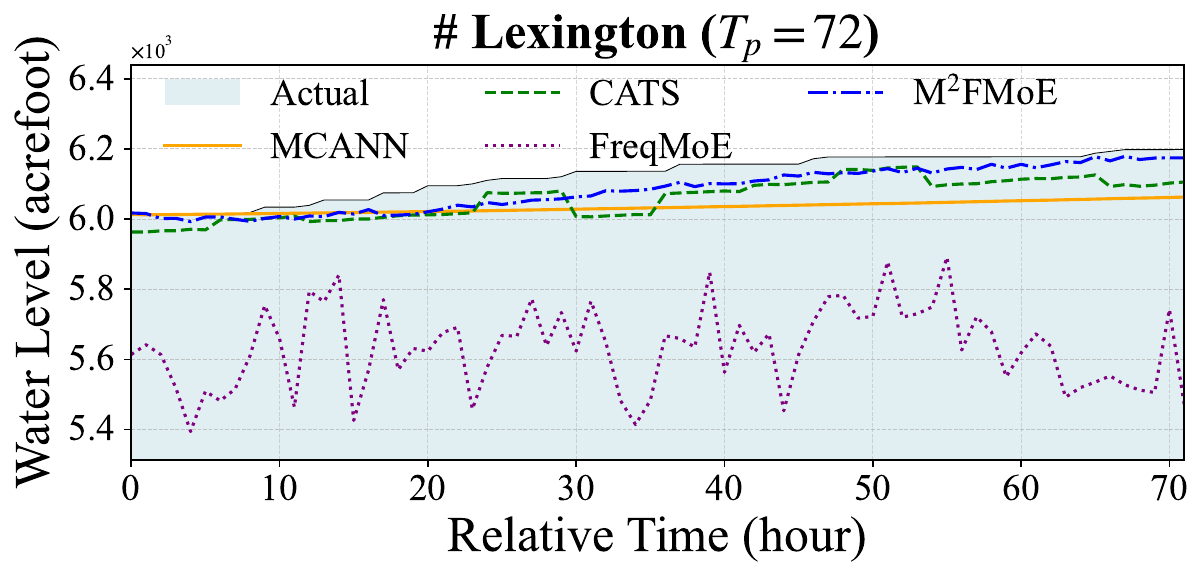}
    \includegraphics[width=0.245\linewidth]{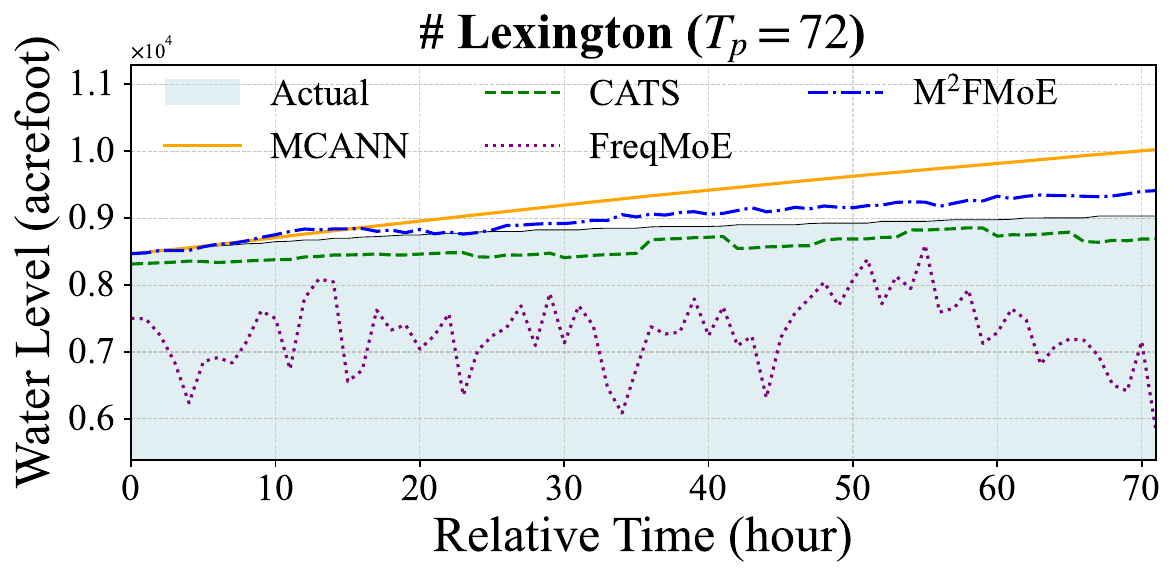}
    \includegraphics[width=0.245\linewidth]{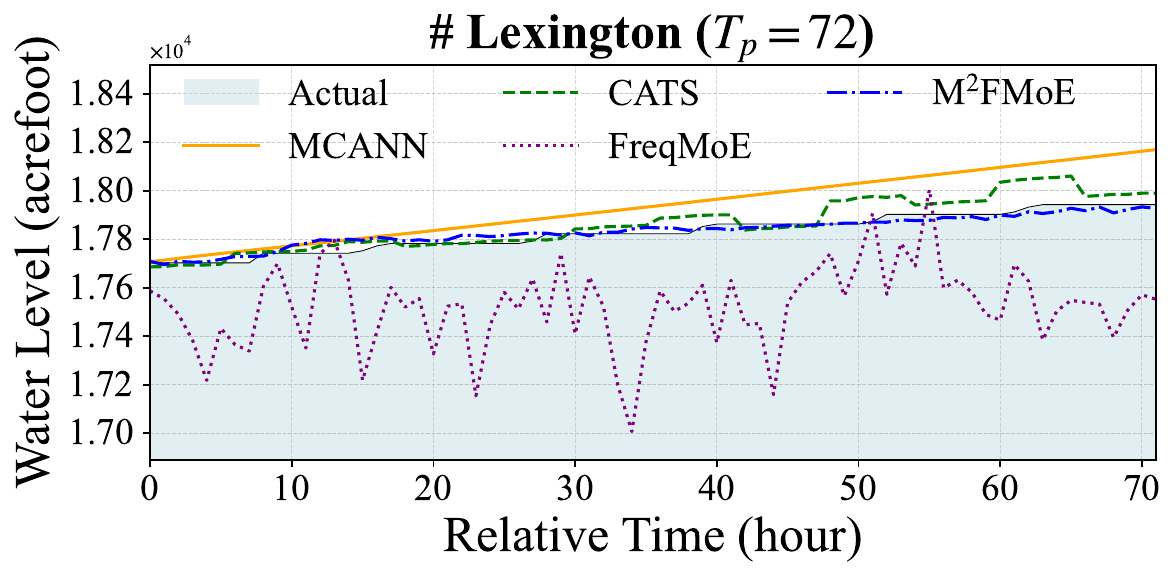}
    \includegraphics[width=0.245\linewidth]{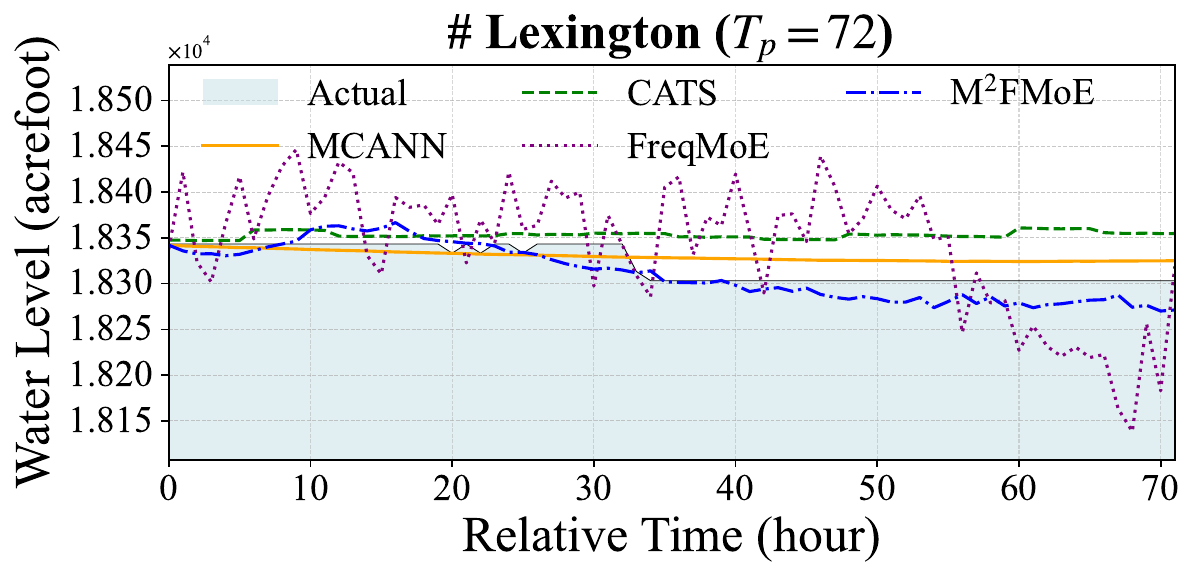}

    \includegraphics[width=0.245\linewidth]{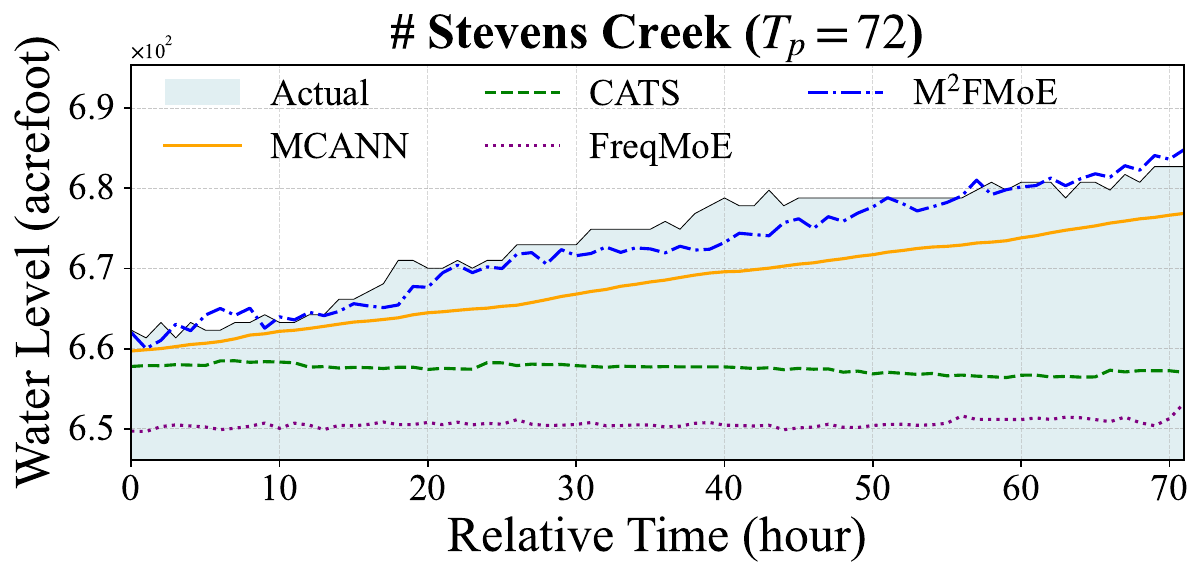}
    \includegraphics[width=0.245\linewidth]{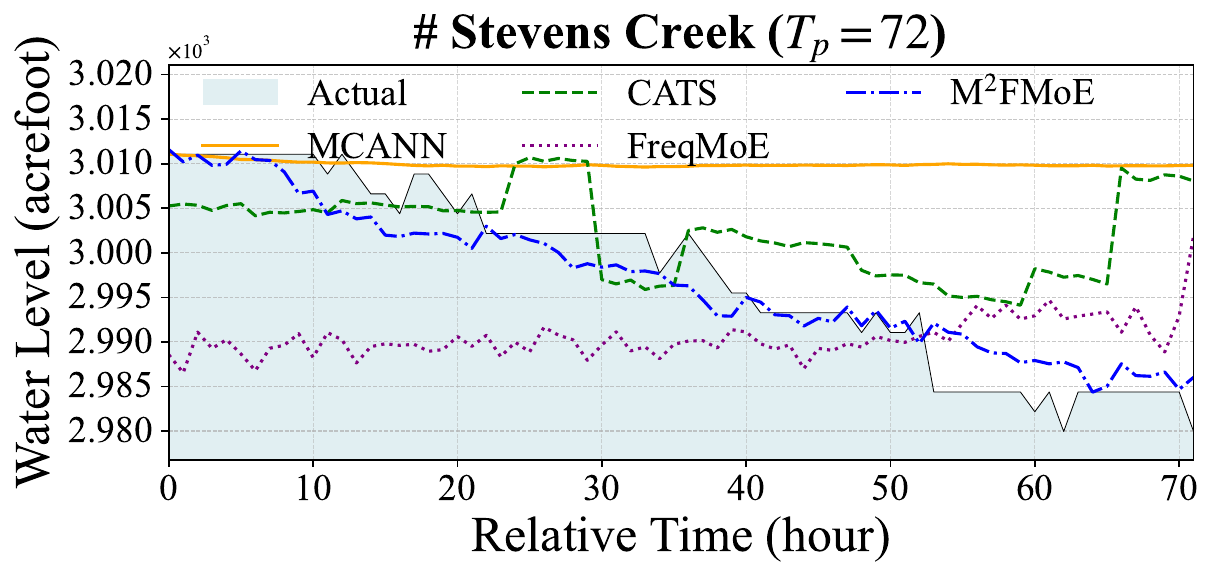}
    \includegraphics[width=0.245\linewidth]{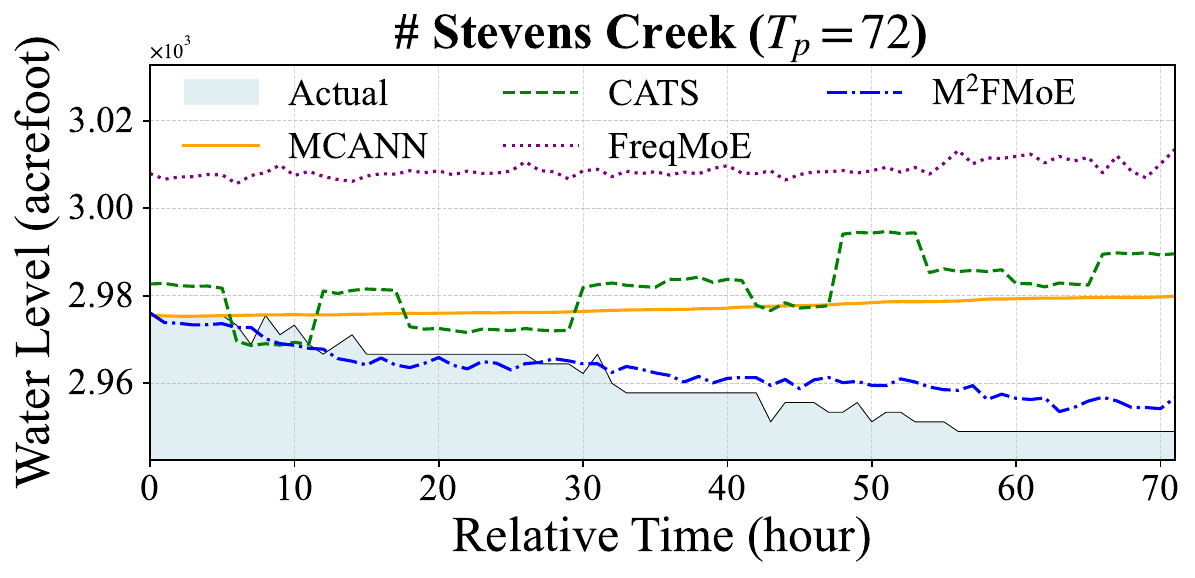}
    \includegraphics[width=0.245\linewidth]{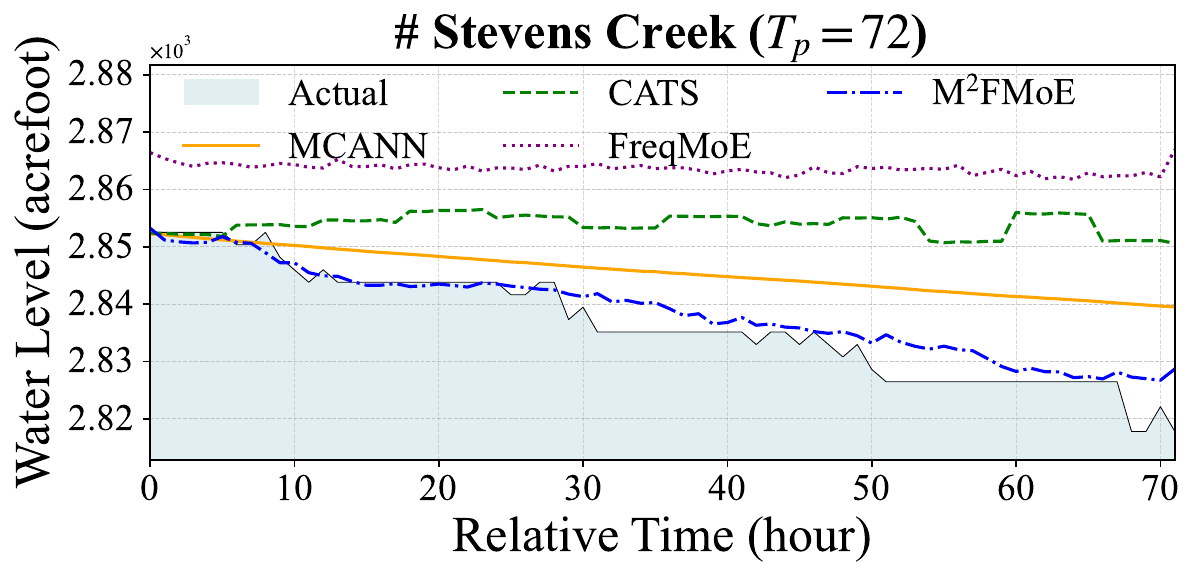}

    \includegraphics[width=0.245\linewidth]{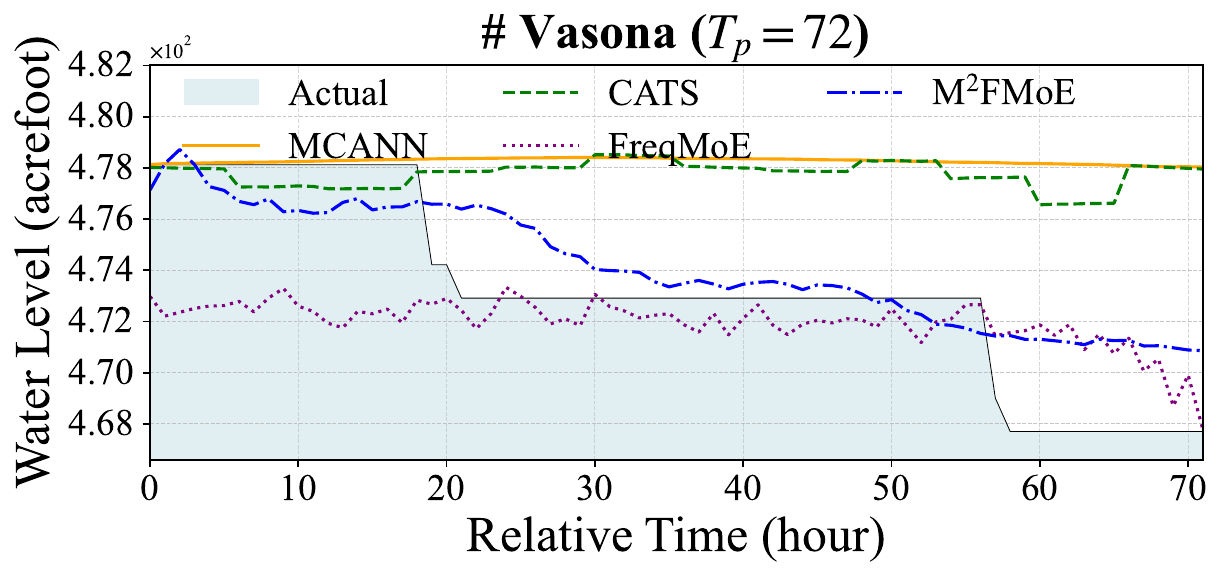}
    \includegraphics[width=0.245\linewidth]{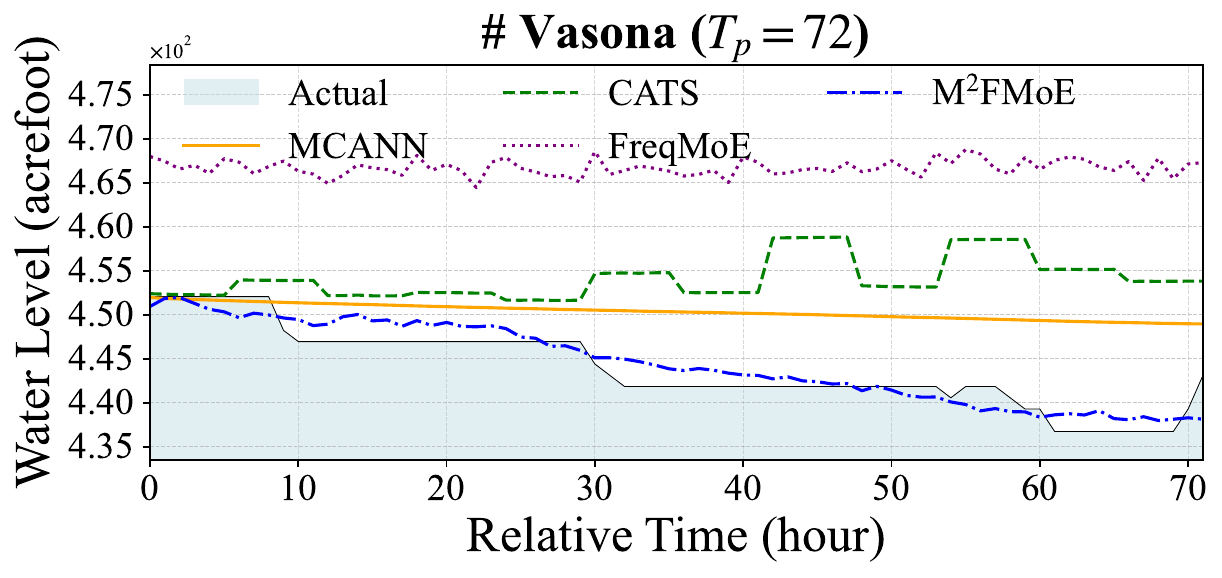}
    \includegraphics[width=0.245\linewidth]{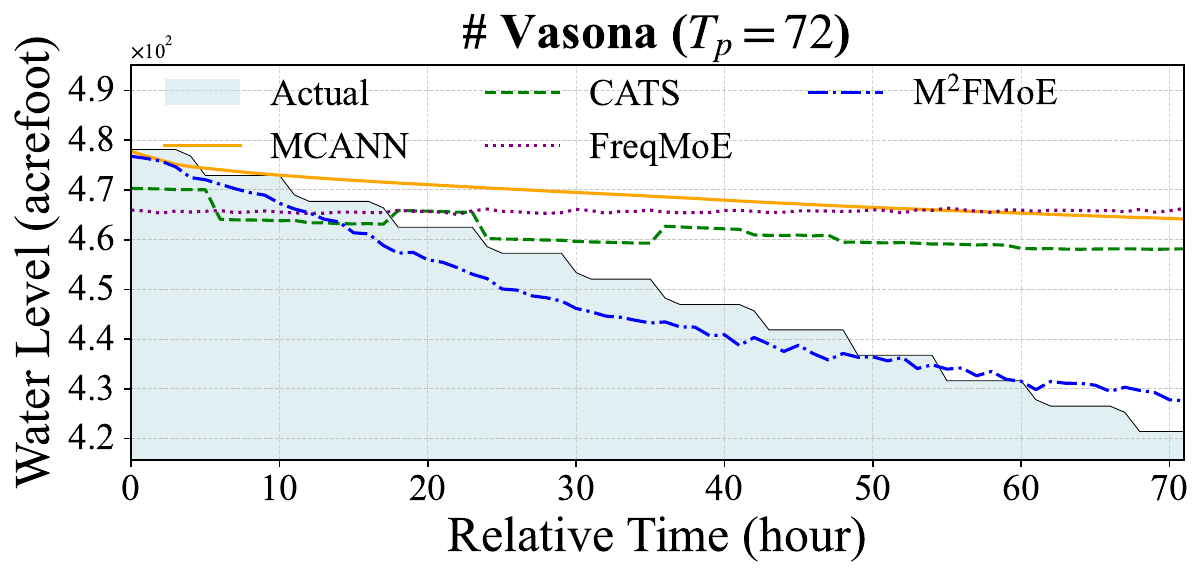}
    \includegraphics[width=0.245\linewidth]{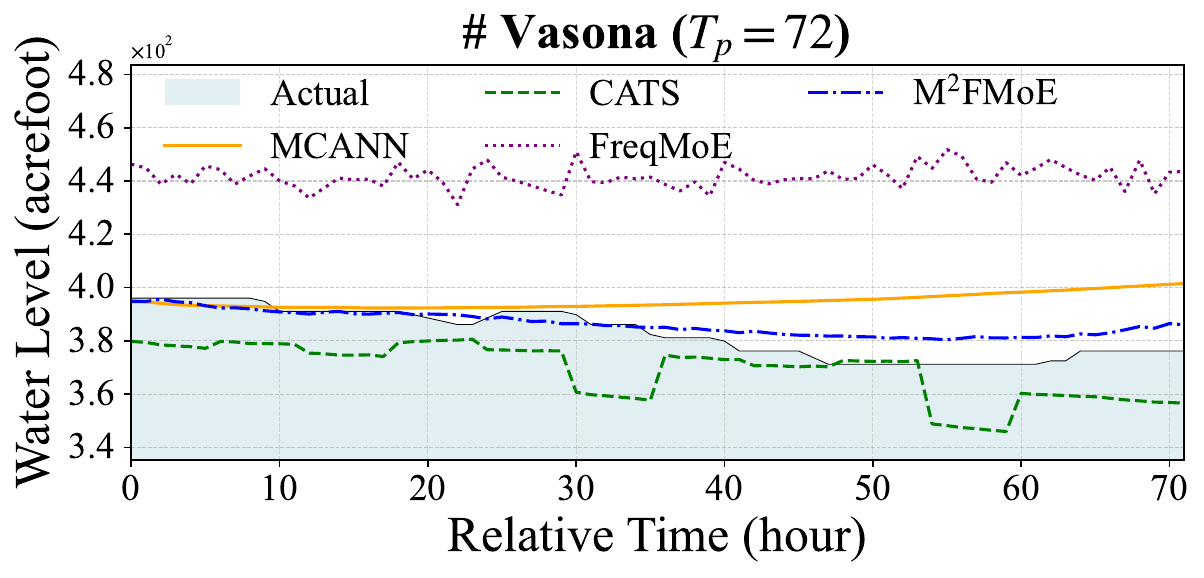}
    \caption{Additional prediction cases of M$^2$FMoE and three baselines (MCANN, FreqMoE, and CATS) on the five reservoirs.}
    \label{fig:additional_cases}
\end{figure*}

\subsection{Appendix D. Additional Results of Ablation Studies}

In this section, we provide further ablation results to validate the effectiveness of the proposed components in M$^2$FMoE. The ablation studies are conducted on the five reservoirs with a prediction horizon of 8 hours, and the results are summarized in \textbf{Table~\ref{tab:ablation}}. The results further validate that each component plays a critical role in enhancing the overall performance of M$^2$FMoE.

\subsection{Appendix E. Additional Cases of Predictions}

In this section, we provide additional cases of predictions made by the M$^2$FMoE and three baselines (MCANN, FreqMoE, and CATS) on the five reservoirs. The results are shown in \textbf{Fig. \ref{fig:additional_cases}}. Each row corresponds to a different reservoir, and each column represents a specific case of prediction. These results further demonstrate the superior performance of M$^2$FMoE in accurately capturing the water level dynamics across different reservoirs and prediction scenarios.

\subsection{Appendix F. Benchmarks}
The detailed descriptions of the selected state-of-the-art baselines are as follows:

\begin{itemize}
   \item \textbf{CATS} \cite{conference/neurips2024/kim} [NeurIPS'24]:
A cross-attention-based framework that utilizes future horizons as queries, enabling parameter sharing across forecasting steps. Demonstrates reduced computational complexity while maintaining high forecasting accuracy across varying horizons.

   \item \textbf{TQNet} \cite{conference/icml2025/lin} [ICML'25]:
A temporal query-based architecture that captures global inter-variable dependencies via single-layer attention. Combines lightweight MLP structures with temporal priors to achieve a favorable trade-off between accuracy and efficiency, particularly in high-dimensional settings.

   \item \textbf{iTransformer} \cite{conference/iclr2024/liu} [ICLR'24]:
An inverted Transformer design that embeds multivariate time series as independent variate tokens. Effectively models multivariate correlations and addresses scalability issues, achieving robust performance on long-sequence, high-dimensional tasks.

   \item \textbf{FreqMoE} \cite{conference/aistats2025/liu25i} [AISTATS'25]:
A frequency-aware Mixture-of-Experts architecture that decomposes time series into frequency components. Assigns specialized expert networks to distinct frequency bands, enabling fine-grained modeling of periodic and oscillatory patterns.

   \item \textbf{Umixer} \cite{conference/aaai2024/ma} [AAAI'24]:
A hybrid Unet-Mixer framework tailored for non-stationary time series. Integrates stationarity correction mechanisms while preserving local temporal dependencies, resulting in enhanced modeling of short-term dynamics under distribution shifts.

   \item \textbf{KAN} \cite{conference/iclr2025/liu2} [ICLR'25]:
A neural architecture replacing traditional MLP layers with learnable spline-based univariate activation functions on edges. Offers improved accuracy and interpretability through flexible, non-linear function approximation.

   \item \textbf{CycleNet} \cite{conference/nips2024/106315lin} [NeurIPS'24]:
A cycle-aware forecasting model that explicitly decouples periodic structure via Residual Cycle Forecasting. Augments existing architectures with cycle learning components, leading to substantial gains in long-term predictive performance.

   \item \textbf{PatchTST} \cite{conference/iclr2023/niepatchtst} [ICLR'23]:
A patch-based, channel-independent Transformer that embeds local subseries as tokens and shares weights across univariate channels. Patching reduces attention cost, extends receptive field, and preserves local semantics. Delivers strong long-term forecasting and state-of-the-art self-supervised transfer performance.

\item \textbf{TimesNet} \cite{conference/iclr2023/wu} [ICLR’23]: A task general backbone that models time series in a two dimensional space by transforming one dimensional sequences into multiple two dimensional representations based on adaptive periodicity. It captures intra period and inter period variations through rows and columns, enabling efficient extraction of complex temporal patterns.

\item \textbf{TimeMixer} \cite{conference/iclr2024/wangtmixer} [ICLR’24]: A fully MLP based architecture that models time series through multiscale mixing. It separates short term and long term patterns using past and future mixing modules, enabling effective forecasting across different horizons.

   \item \textbf{DAN} \cite{conference/aaai2024/li27768} [AAAI'24]:
An event-conditioned model that integrates domain-specific extreme event labels into the forecasting process. Enhances predictive precision during anomalous intervals by leveraging supervision on high-impact events.

   \item \textbf{MCANN} \cite{journal/tpami2025/li6888} [TPAMI'25]:
A multi-component framework combining autoencoder-based forecasting with clustering-informed attention. Adapts to skewed and imbalanced time series distributions, offering strong performance in long-horizon forecasting under extreme-value scenarios.

\end{itemize}

\subsection{Appendix G. Sensitivity Analysis of Resolution Size}

\begin{figure*}[t!]
    \centering
    \includegraphics[width=0.48\linewidth]{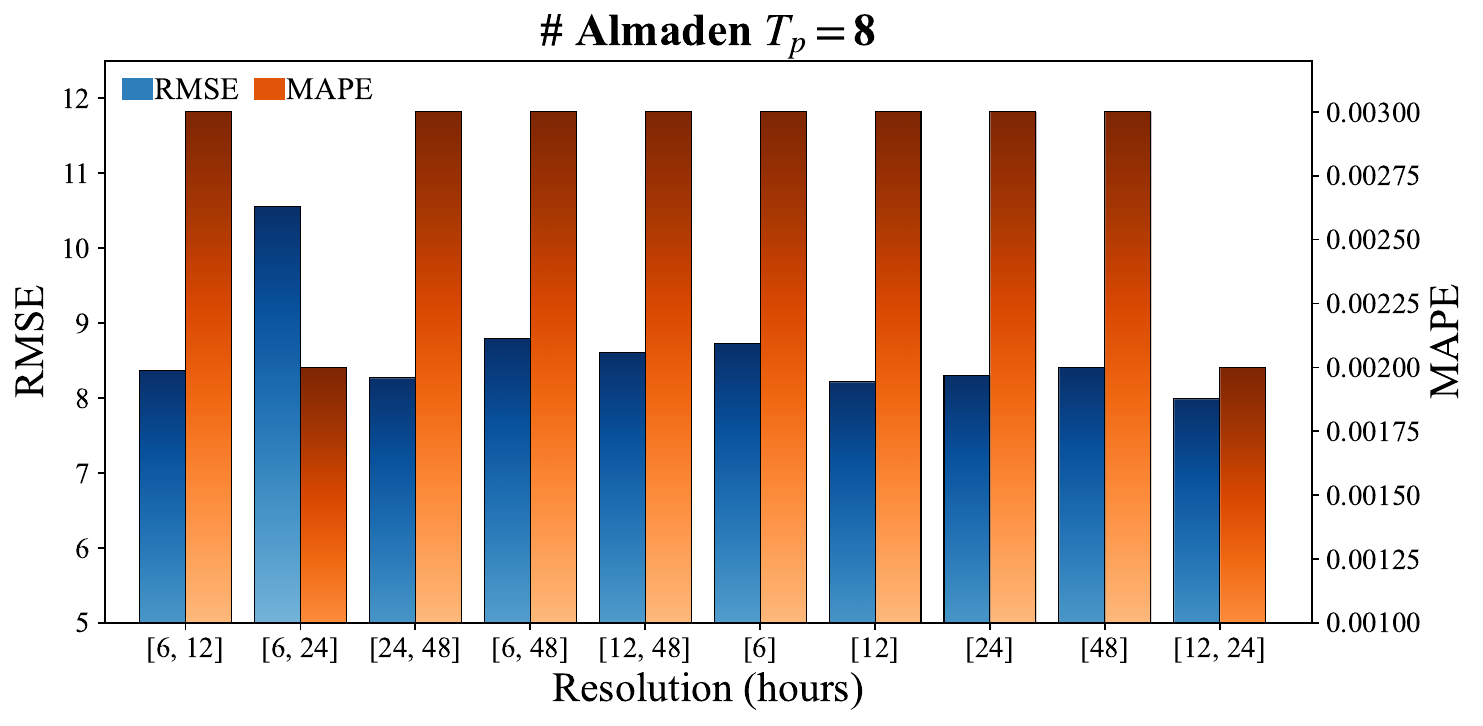}
    \includegraphics[width=0.48\linewidth]{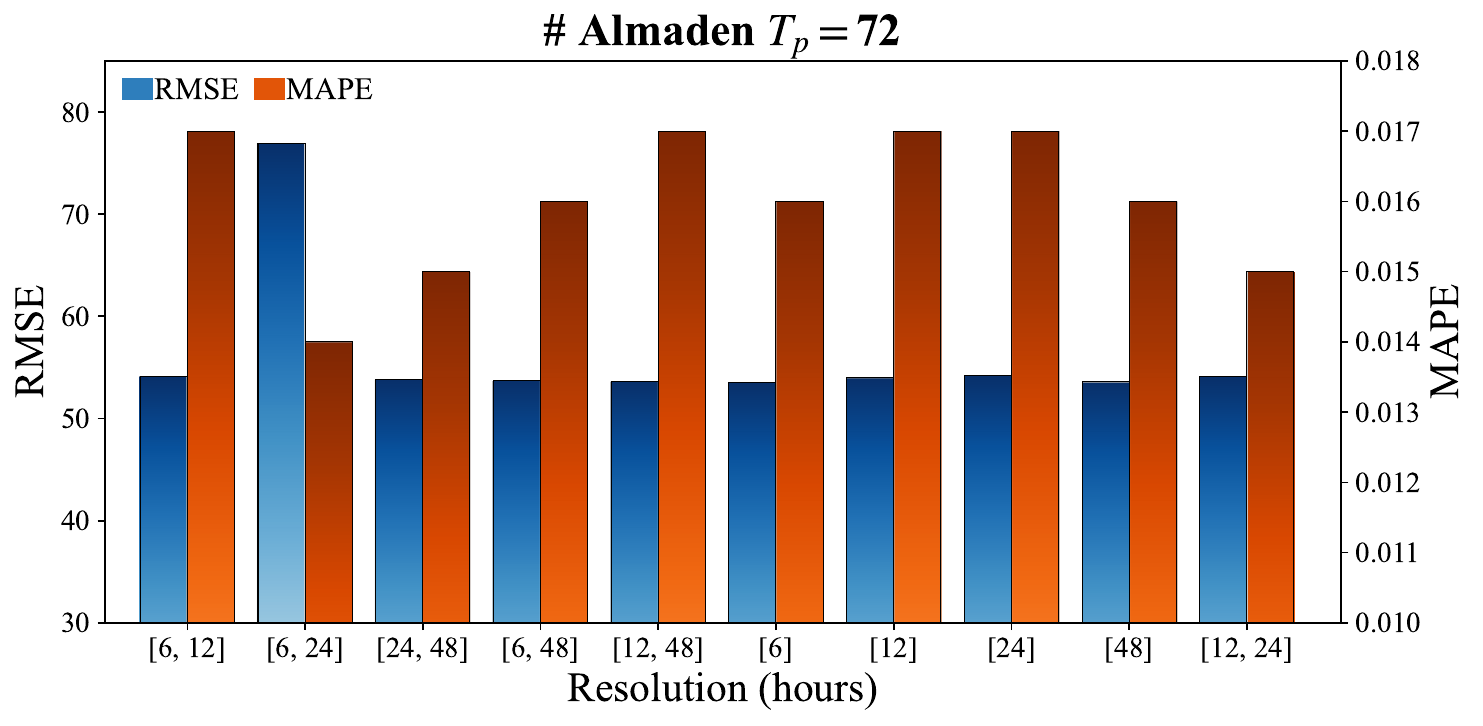}
    \caption{Sensitivity analysis of resolution size on Almaden reservoir.}
    \label{fig:resolution_sensitivity}
\end{figure*}

\textbf{Fig.~\ref{fig:resolution_sensitivity}} presents a sensitivity analysis of the resolution size hyper-parameter on the Almaden reservoir, evaluated at prediction horizons of 8 hours and 72 hours. The resolution size determines the granularity of the wavelet transform, influencing the model's ability to capture temporal features at varying scales.

\subsection{Appendix H. Computational Cost}

\begin{figure}[t!]
    \centering
    \includegraphics[width=1.0\linewidth]{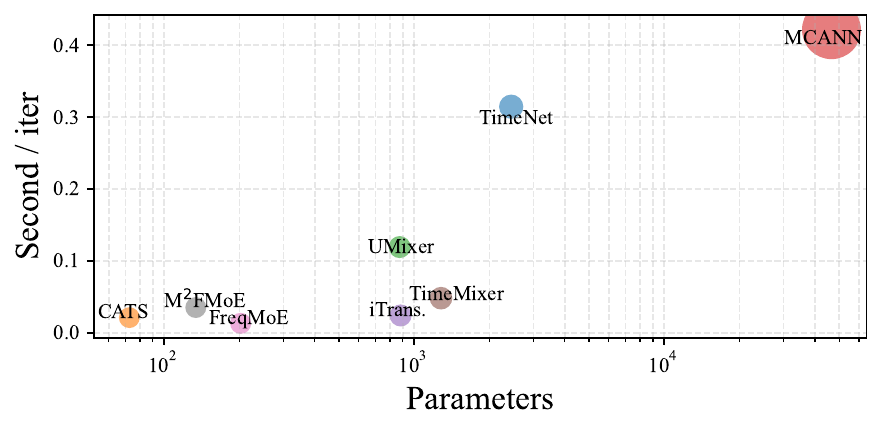}

    \caption{Computational cost comparison of different models.}

    \label{fig:computational_cost}
\end{figure}

\textbf{Fig.~\ref{fig:computational_cost}} presents a comparative analysis of the computational costs associated with various models. The evaluation metrics include training time per iteration (s/iter) and model size (number of parameters in thousands, K).
The integration of the wavelet perspective for extreme-event modeling introduces only a minor overhead of 0.03 s/iter on the Almaden dataset. Overall, M$^2$FMoE remains highly efficient: it is 0.32× slower than iTrans, yet still 0.37× faster than TimeMixer and 11.1× faster than MCANN.
In terms of model size, M$^2$FMoE employs an MLP-based MoE with 134K parameters, which is slightly larger than FreqMoE and CATS, but 4–8× smaller than TimeNet, TimeMixer, and iTrans, and over 20× smaller than MCANN.

\begin{figure*}[t!]
    \centering
    \subfigure[Extreme Events]{
        \includegraphics[width=0.48\linewidth]{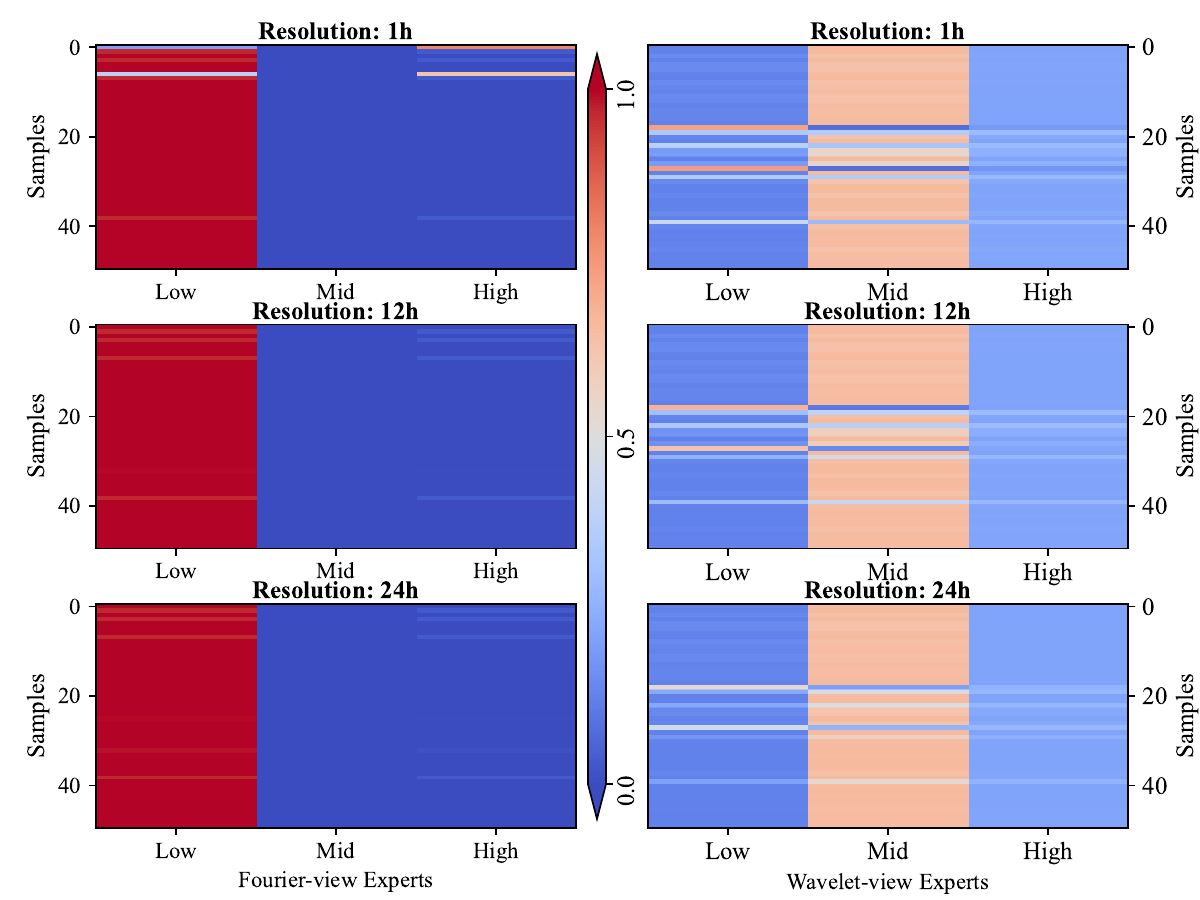}
        \label{fig:expert_behavior_extreme}
    }
    \subfigure[Regular Events]{
        \includegraphics[width=0.48\linewidth]{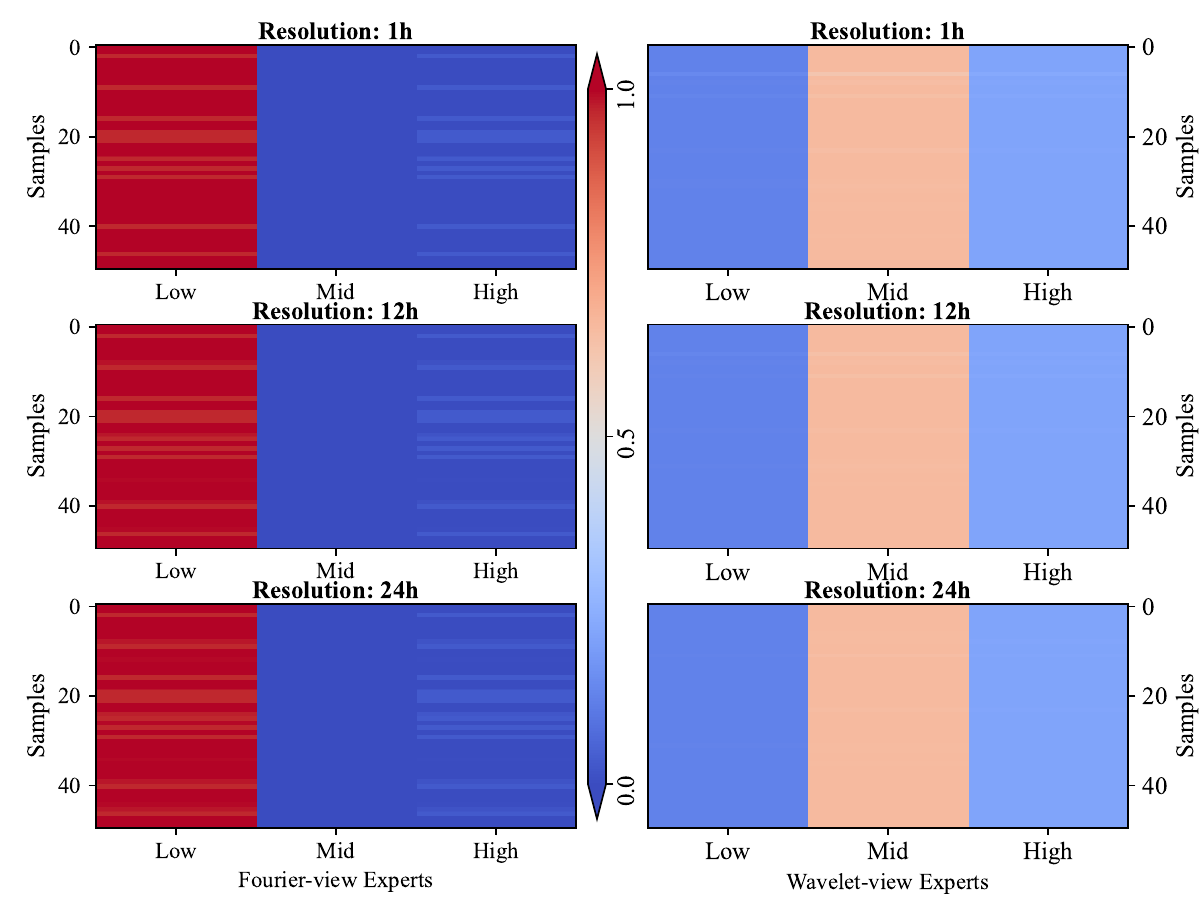}
        \label{fig:expert_behavior_regular}
    }
    \caption{Visualization of expert behavior under extreme and regular events.}
    \label{fig:expert_behavior}
\end{figure*}

\subsection{Appendix I. Visualization of Expert Behavior under Regular and Extreme Events}

\textbf{Fig.~\ref{fig:expert_behavior}} visualizes the expert activations of M$^2$FMoE under both extreme and regular events.
\textbf{Fig.~\ref{fig:expert_behavior_extreme}} illustrates expert activations during extreme events, while \textbf{Fig.~\ref{fig:expert_behavior_regular}} depicts activations during regular events. In each panel, the left plot shows the Fourier view expert activations, and the right plot displays the Wavelet view expert activations.

The results indicate that during extreme events, the Wavelet view experts exhibit heightened activation levels compared to the Fourier view experts. This suggests that the Wavelet view is more responsive to high-impact fluctuations characteristic of extreme events. Conversely, during regular events, the Fourier view experts demonstrate stronger activations, reflecting their proficiency in capturing stable, periodic patterns prevalent in normal conditions.

\subsection{Appendix J. Additional Benchmark Results} 

\begin{table*}[t!]
\centering
\small
\setlength{\tabcolsep}{1mm}
\resizebox{1.0\linewidth}{!}{	
    \begin{tabular}{ccc|c|cccccccccc|cc}
    \toprule
    \multirow{2}[4]{*}{\textbf{Data}} & \multirow{2}[4]{*}{\textbf{Metrics}} & \multirow{2}[4]{*}{\textbf{Steps}} & \multicolumn{11}{c|}{\textit{ without extreme lables}}                                & \multicolumn{2}{c}{\textit{ with extreme lables}} \\
\cmidrule{4-16}          &       &       & \textbf{M2FMoE} & \textbf{CATS} & \textbf{CycleNet} & \textbf{FreqMoE} & \textbf{iTrans} & \textbf{KAN} & \textbf{TQNet} & \textbf{Umixer} & \textbf{PatchTST} & \textbf{TimesNet} & \textbf{TimeMixer} & \textbf{DANet} & \textbf{MCANN} \\
    \midrule
    \multirow{4}[3]{*}{\begin{sideways}\textbf{Almaden}\end{sideways}} & RMSE  & \multirow{2}[2]{*}{8} & \textbf{7.990 } & 16.087  & 17.754  & 14.729  & 32.127  & 18.934  & 18.023  & 18.658  & 13.324  & 15.485  & 9.056  & 37.857  & \underline{8.447}  \\
          & MAPE  &       & \textbf{0.002 } & 0.006  & 0.007  & 0.005  & 0.017  & 0.009  & 0.010  & 0.007  & 0.005  & \underline{0.004}  & \underline{0.004}  & 0.021  & \textbf{0.002 } \\
\cmidrule{2-16}          & RMSE  & \multirow{2}[1]{*}{72} & \textbf{54.120 } & 57.916  & 61.379  & 63.038  & 65.325  & 70.181  & 59.427  & 64.816  & 60.830  & 62.965  & 56.996  & 66.597  & \underline{56.840}  \\
          & MAPE  &       & \textbf{0.015 } & \textbf{0.015 } & 0.019  & 0.017  & 0.025  & 0.033  & 0.018  & 0.018  & 0.017  & \underline{0.016}  & \underline{0.016}  & 0.025  & \textbf{0.015 } \\
    \midrule
    \multirow{4}[3]{*}{\begin{sideways}\textbf{Coyote}\end{sideways}} & RMSE  & \multirow{2}[1]{*}{8} & \textbf{48.797 } & 110.849  & 113.706  & 593.141  & 372.523  & 116.398  & 103.521  & 174.892  & 95.355  & 107.731  & \underline{71.803}  & 505.941  & 86.829  \\
          & MAPE  &       & \textbf{0.002 } & 0.004  & \underline{0.003}  & 0.018  & 0.022  & 0.005  & \underline{0.003}  & 0.005  & \underline{0.003}  & \underline{0.003}  & 0.004  & 0.025  & \textbf{0.002 } \\
\cmidrule{2-16}          & RMSE  & \multirow{2}[2]{*}{72} & \textbf{449.944 } & 509.077  & 528.962  & 855.096  & 673.853  & 587.132  & 504.606  & 566.429  & 509.353  & 528.424  & \underline{499.276}  & 829.623  & 559.747  \\
          & MAPE  &       & \textbf{0.012 } & \textbf{0.012 } & \textbf{0.012 } & 0.025  & 0.029  & 0.021  & \textbf{0.012 } & \underline{0.013}  & \textbf{0.012 } & \textbf{0.012 } & 0.014  & 0.042  & \textbf{0.012 } \\
    \midrule
    \multirow{4}[4]{*}{\begin{sideways}\textbf{Lexington}\end{sideways}} & RMSE  & \multirow{2}[2]{*}{8} & \textbf{251.957 } & 618.991  & 463.293  & 386.995  & 690.426  & 429.054  & 400.991  & 466.669  & 434.376  & 380.766  & 354.418  & 476.936  & \underline{252.965}  \\
          & MAPE  &       & \underline{0.004}  & 0.011  & 0.011  & 0.006  & 0.041  & 0.008  & 0.013  & 0.008  & 0.008  & 0.006  & 0.008  & 0.015  & \textbf{0.003 } \\
\cmidrule{2-16}          & RMSE  & \multirow{2}[2]{*}{72} & \textbf{772.836 } & 906.531  & 865.092  & 1003.818  & 960.652  & 956.134  & 860.456  & 829.541  & 805.202  & 856.938  & 897.549  & 908.308  & \underline{778.023}  \\
          & MAPE  &       & \textbf{0.014 } & 0.020  & 0.021  & 0.018  & 0.048  & 0.020  & 0.025  & 0.018  & 0.017  & 0.017  & 0.020  & 0.024  & \underline{0.015}  \\
    \midrule
    \multicolumn{1}{c}{\multirow{4}[4]{*}{\begin{sideways}\textbf{Ste. Creek}\end{sideways}}} & RMSE  & \multirow{2}[2]{*}{8} & \textbf{10.559 } & 18.500  & 28.400  & 80.937  & 48.876  & 25.672  & 24.475  & 37.654  & 22.349  & 26.923  & 16.593  & 24.319  & \underline{12.130}  \\
          & MAPE  &       & \textbf{0.002 } & 0.004  & 0.005  & 0.017  & 0.010  & 0.005  & 0.006  & 0.007  & 0.005  & 0.004  & 0.004  & 0.011  & \textbf{0.002 } \\
\cmidrule{2-16}          & RMSE  & \multirow{2}[2]{*}{72} & \textbf{76.939 } & 82.739  & 94.578  & 117.282  & 106.606  & 94.034  & 89.265  & 141.505  & 82.244  & 96.069  & 88.339  & 82.794  & \underline{81.084}  \\
          & MAPE  &       & 0.014  & \textbf{0.011 } & 0.014  & 0.025  & 0.017  & 0.015  & \underline{0.012}  & 0.017  & 0.013  & 0.013  & 0.017  & 0.020  & \textbf{0.011 } \\
    \midrule
    \multirow{4}[4]{*}{\begin{sideways}\textbf{Vasona}\end{sideways}} & RMSE  & \multirow{2}[2]{*}{8} & \textbf{5.129 } & 6.913  & 7.903  & 14.318  & 12.179  & 11.308  & 7.741  & 9.299  & 7.574  & 7.492  & 6.241  & 9.562  & \underline{5.353}  \\
          & MAPE  &       & \textbf{0.004 } & 0.007  & 0.007  & 0.020  & 0.013  & 0.019  & 0.007  & 0.009  & 0.008  & \underline{0.006}  & \underline{0.006}  & 0.012  & \textbf{0.004 } \\
\cmidrule{2-16}          & RMSE  & \multirow{2}[2]{*}{72} & 19.571  & 20.381  & 20.713  & 20.740  & 21.534  & 21.605  & 20.173  & 23.718  & \underline{19.122}  & 21.618  & 21.247  & 20.542  & \textbf{18.634 } \\
          & MAPE  &       & 0.021  & 0.021  & 0.021  & 0.027  & 0.023  & 0.027  & \underline{0.020}  & 0.023  & \underline{0.020}  & 0.021  & 0.022  & 0.023  & \textbf{0.019 } \\
    \midrule
    \multicolumn{3}{c|}{Average Rank ~/~ Significance} & \textbf{1.6 } & 5.3~/~$\star$   & 7.1~/~$\ast$   & 9.8~/~$\ast$   & 11.4~/~$\ast$  & 9.6~/~$\ast$   & 6.2~/~$\ast$   & 9.0~/~$\ast$   & 4.5~/~$\ast$   & 5.1~/~$\ast$   & 5.0~/~$\ast$   & 10.7~/~$\ast$  & 1.9~/~$\star$  \\
    \bottomrule
    \end{tabular}%
}
  \caption{Performance comparison on five reservoirs with predicted length as \{8, 72\} hours. 
  $\ast$: both metrics are statistically significant ($p<0.05$, Wilcoxon signed-rank test); $\star$: indicates significance in RMSE. Best results are \textbf{bold}, second-best \underline{underlined}.}
  \label{tab:additional_main_results}
\end{table*}

\textbf{Table~\ref{tab:additional_main_results}} presents additional benchmark results with prediction lengths of 8 and 72 hours, incorporating three advanced baselines (PatchTST~\cite{conference/iclr2023/niepatchtst}, TimesNet~\cite{conference/iclr2023/wu}, and TimeMixer~\cite{conference/iclr2024/wangtmixer}) recommended by anonymous reviewers. The results further confirm the superior performance of M$^2$FMoE compared to these state-of-the-art methods across various metrics and prediction horizons.

\begin{table*}[t!]
  \centering
  \resizebox{1.0\linewidth}{!}{	
    \begin{tabular}{c|cc|cc|cc|cc|cc|cc}
    \toprule
    Datasets & \multicolumn{4}{c|}{Wellington, NZ} & \multicolumn{4}{c|}{Long Beach, CA, US} & \multicolumn{2}{c|}{ETTh1} & \multicolumn{2}{c}{ETTh2} \\
    \midrule
    Settings & \multicolumn{2}{c|}{72h$\rightarrow$3h} & \multicolumn{2}{c|}{72h$\rightarrow$6h} & \multicolumn{2}{c|}{72h$\rightarrow$3h} & \multicolumn{2}{c|}{72h$\rightarrow$6h} & \multicolumn{2}{c|}{96h$\rightarrow$96h} & \multicolumn{2}{c}{96h$\rightarrow$96h} \\
    \midrule
    Metrics & RMSE  & MAPE  & RMSE  & MAPE  & RMSE  & MAPE   & RMSE  & MAPE  & MAE   & MSE   & MAE   & MSE \\
    \midrule
    KAN   & 1.636  & 0.150  & 2.348  & 0.222  & 0.851  & 0.208  & 1.120  & 0.268  & 0.476  & 0.492  & -     & - \\
    PatchTST & 1.496  & 0.135  & 2.184  & 0.202  & 0.798  & 0.187  & \underline{1.095} & 0.244  & 0.419  & \underline{0.414} & \textbf{0.348} & \textbf{0.302} \\
    TimesNet & \underline{1.294} & \underline{0.114} & \underline{2.113} & \underline{0.188} & \underline{0.793} & \underline{0.180} & 1.113  & \underline{0.242} & \textbf{0.402} & \textbf{0.384} & \underline{0.374} & \underline{0.340} \\
    M$^2$FMoE & \textbf{1.248} & \textbf{0.106} & \textbf{2.041} & \textbf{0.180} & \textbf{0.777} & \textbf{0.177} & \textbf{1.072} & \textbf{0.238} & \underline{0.414} & 0.422  & 0.399  & 0.370  \\
    \bottomrule
    \end{tabular}%
  }

    \caption{Performance comparison on two wind speed datasets and two ETT datasets. Best results are \textbf{bold}, second-best \underline{underlined}. `–' indicates that the model failed to converge and therefore no valid results were obtained.}
    \label{tab:wind_speed}%
\end{table*}%

\textbf{Table \ref{tab:wind_speed}} presents a performance comparison between M$^2$FMoE and three strong baseline models, namely KAN, PatchTST, and TimesNet, evaluated on two wind speed datasets, Wellington and Long Beach, as well as two ETT benchmark datasets, ETTh1 and ETTh2.
The wind speed datasets are collected from the NASA POWER project\footnote{https://power.larc.nasa.gov/data-access-viewer/} and consist of hourly wind speed measurements at a height of 50 meters from two locations, Wellington in New Zealand and Long Beach in California, United States, covering the period from January 1, 2025 to November 30, 2025. The data are split into training, validation, and testing sets with ratios of 70\%, 10\%, and 20\%, respectively.
The ETT datasets are widely used benchmarks for multivariate time series forecasting.
The experimental results indicate that M$^2$FMoE consistently outperforms the baseline models across various prediction horizons and evaluation metrics on the short term wind speed datasets. M$^2$FMoE also shows competitive performance on the ETTh1 and ETTh2 benchmark datasets. Overall, its performance is stronger for short term forecasting, while for the 96 hour ahead forecasts, M$^2$FMoE obtains slightly higher MAE and MSE values than PatchTST and TimesNet.

\end{document}